\documentclass{article}

\PassOptionsToPackage{numbers, compress}{natbib}

\usepackage[final]{neurips_2019}

\usepackage[utf8]{inputenc} 
\usepackage[T1]{fontenc}    
\usepackage{hyperref}       
\usepackage{url}            
\usepackage{booktabs}       
\usepackage{nicefrac}       
\usepackage{microtype}      
\usepackage[normalem]{ulem}      

\usepackage[pdftex]{graphicx}
\usepackage{amsmath,amssymb,amsthm, dsfont}
\usepackage[caption = false]{subfig}
\usepackage[usenames,dvipsnames]{xcolor}
\usepackage{graphicx}
\usepackage[noline,ruled]{algorithm2e}
\usepackage{tikz}
\usepackage{cancel}

\newtheorem{proposition}{Proposition}

\definecolor{cffffff}{RGB}{255,255,255}

\makeatletter
\newsavebox\myboxA
\newsavebox\myboxB
\newlength\mylenA
\newcommand*\xoverline[2][0.75]{%
    \sbox{\myboxA}{$\m@th#2$}%
    \setbox\myboxB\null
    \ht\myboxB=\ht\myboxA%
    \dp\myboxB=\dp\myboxA%
    \wd\myboxB=#1\wd\myboxA
    \sbox\myboxB{$\m@th\overline{\copy\myboxB}$}
    \setlength\mylenA{\the\wd\myboxA}
    \addtolength\mylenA{-\the\wd\myboxB}%
    \ifdim\wd\myboxB<\wd\myboxA%
       \rlap{\hskip 0.5\mylenA\usebox\myboxB}{\usebox\myboxA}%
    \else
        \hskip -0.5\mylenA\rlap{\usebox\myboxA}{\hskip 0.5\mylenA\usebox\myboxB}%
    \fi}
\makeatother

\title{Expressive power of tensor-network\\ factorizations for probabilistic modeling\\[.2cm] \large{\textit{with applications from hidden Markov models to quantum machine learning}}}

\author{Ivan Glasser\textsuperscript{1,2}\thanks{Corresponding author, \texttt{ivan.glasser@mpq.mpg.de}}\,\,, Ryan Sweke\textsuperscript{3}, Nicola Pancotti\textsuperscript{1,2}, Jens Eisert\textsuperscript{3,4}, J. Ignacio Cirac\textsuperscript{1,2}\\
\textsuperscript{1}Max-Planck-Institut f\"ur Quantenoptik, D-85748 Garching\\
\textsuperscript{2}Munich Center for Quantum Science and Technology (MCQST), D-80799 M\"unchen\\
\textsuperscript{3}Dahlem Center for Complex Quantum Systems, Freie Universit\"at Berlin, D-14195 Berlin\\
\textsuperscript{4}Department of Mathematics and Computer Science, Freie Universit\"at Berlin, D-14195 Berlin
}

\begin{document}

\maketitle

\begin{abstract}
Tensor-network techniques have enjoyed outstanding success in physics, and have recently attracted attention in machine learning, both as a tool for the formulation of new learning algorithms and for enhancing the mathematical understanding of existing methods.
Inspired by these developments, and the natural correspondence between tensor networks and probabilistic graphical models, we provide a rigorous analysis of the expressive power of various tensor-network factorizations of discrete multivariate probability distributions. 
These factorizations include non-negative tensor-trains/MPS, which are in correspondence with hidden Markov models, and Born machines, which are naturally related to the probabilistic interpretation of local quantum circuits. 
When used to model probability distributions, they exhibit tractable likelihoods and admit efficient learning algorithms.
Interestingly, we prove that there exist probability distributions for which there are unbounded separations between the resource requirements of some of these tensor-network factorizations. Particularly surprising is the fact that using complex instead of real tensors can lead to an arbitrarily large reduction in the number of parameters of the network.
Additionally, we introduce locally purified states (LPS), a new factorization inspired by techniques for the simulation of quantum systems, with provably better expressive power than all other representations considered. The ramifications of this result are explored through numerical experiments. Our findings imply that LPS should be considered over hidden Markov models, and furthermore provide guidelines for the design of local quantum circuits for probabilistic modeling.
\end{abstract}

\section{Introduction}
\label{sec:intro}

Many problems in diverse areas of computer science and physics involve constructing efficient representations of high-dimensional functions. Neural networks are a particular example of such representations that have enjoyed great empirical success, and much effort has been dedicated to understanding their expressive power - i.e. the set of functions that they can efficiently represent.
Analogously, tensor networks are a class of powerful representations of high-dimensional arrays (tensors), for which a variety of algorithms and methods have been developed. Examples of such tensor networks are tensor trains/matrix product states (MPS) \cite{PhysRevLett.75.3537,Oseledets2011} or the hierarchical Tucker decomposition \cite{Hackbusch2009ell,DBLP:journals/siammax/Grasedyck10}, which have found application in data compression \cite{MAL-059,MAL-067, Novikov2015}, the simulation of physical systems \cite{SchollwockReview, Orus2014, VerstraeteReview} and the design of machine learning algorithms \cite{DBLP:conf/aistats/SedghiJA16, DBLP:journals/jcss/HsuKZ12,627946317, Stoudenmire2016,Novikov2016,627946325}. In addition to their use in numerical algorithms, tensor networks enjoy a rich analytical understanding which has facilitated their use as a tool for obtaining rigorous results on the expressive power of deep learning models  \cite{627946320,Cohen2016,627946390,Cohen2017,Levine2017,DBLP:conf/iclr/KhrulkovNO18}, and fundamental insights into the structure of quantum mechanical systems \cite{eisert2013entanglement}.

In the context of probabilistic modeling, tensor networks have been shown to be in natural correspondence with probabilistic graphical models \cite{Morton2014,Kliesch2014,Chen2017,Glasser2017,Robeva2017,Glasser2018}, as well as with Sum-Product Networks and Arithmetic Circuits \cite{627946320,DBLP:conf/nips/JainiPY18,DBLP:journals/corr/SharirTCS16}. Motivated by this correspondence, and with the goal of enhancing the toolbox for deriving analytical results on the properties of machine learning algorithms, we study the expressive power of various tensor-network models of discrete multivariate probability distributions. The models we consider, defined in Section~\ref{sec:ansatz}, fall into two main categories: 
\begin{itemize}
    \item {\bf Non-negative tensor networks}, which decompose a probability mass function as a network of non-negative tensors \cite{Shashua:2005:NTF:1102351.1102451}, as in a probabilistic graphical model \cite{Frey2003}.
    \item {\bf Born machines (BM)}, which model a probability mass function as the absolute value squared of a real or complex function, which is itself represented as a network of real or complex tensors. While Born machines have been previously employed for probabilistic modeling \cite{Han2017, Cheng2017, PhysRevLett.121.260602, DBLP:journals/corr/abs-1711-01416, Stoudenmire2018, DBLP:journals/corr/abs-1902-06888,PhysRevB.99.155131}, they have additional potential applications in the context of quantum machine learning \cite{Liu2018b, Benedetti2018,Grant2018, Huggins2018}, since they arise naturally from the probabilistic interpretation of quantum mechanics.
\end{itemize}

These models are considered precisely because they represent non-negative tensors by construction. In this work we focus on tensor networks which are based on tensor-trains/MPS and generalizations thereof, motivated by the fact that these have tractable likelihood, and thus efficient learning algorithms, while lending themselves to a rigorous theoretical analysis. In this setting non-negative tensor networks encompass hidden Markov models (HMM), while Born machines include models that arise from local quantum circuits of fixed depth. Our results also apply to tensor networks with a tree structure, and as such can be seen as a more general comparison of the difference between non-negative tensor networks and Born machines.

The main result of this work is a characterization of the expressive power of these tensor networks. Interestingly, we prove that there exist families of probability distributions for which there are unbounded separations between the resource requirements of some of these tensor-network factorizations. This allows us to show that neither HMM nor Born machines should be preferred to each other in general. Moreover, we prove that using complex instead of real tensors can sometimes lead to an arbitrarily large reduction in the number of parameters of the network.

Furthermore, we introduce a new tensor-network model of discrete multivariate probability distributions with provably better expressive power than the previously introduced models. This tensor network, which retains an efficient learning algorithm, is referred to as a locally purified state (LPS) due to its origin in the classical simulation of quantum systems \cite{PhysRevLett.93.207204, Cuevas_2013, Barthel_2013, werner2016positive}. We demonstrate through numerical experiments on both random probability distributions as well as realistic data sets that our theoretical findings are relevant in practice - i.e. that LPS should be considered over HMM and Born machines for probabilistic modeling.

This paper is structured as follows: The models we consider are introduced in Section~\ref{sec:ansatz}. Their relation with HMM and quantum circuits is made explicit in Section~\ref{sec:hmmquantum}.
The main results on expressive power are presented in Section~\ref{sec:expressivity2}. Section~\ref{sec:algo} then introduces learning algorithms for these tensor networks, and the results of numerical experiments are provided in Section~\ref{sec:numerical}.

\section{Tensor-network models of probability distributions}
\label{sec:ansatz}

Consider a multivariate probability mass function $P(X_1,\ldots,X_N)$ over $N$ discrete random variables $\{X_i\}$ taking values in $\{1,\ldots,d\}$. This probability mass function is naturally represented as a multi-dimensional array, or tensor, with $N$ indices, each of which can take $d$ values. As such, we use the notation $P$ to refer simultaneously to both the probability mass function and the equivalent tensor representation. More specifically, for each configuration $X_1,\ldots,X_N$ the tensor element $P_{X_1,\ldots,X_N}$ stores the probability $P(X_1,\ldots,X_N)$.  
Note that as $P$ is a representation of a probability mass function, it is a tensor with non-negative entries summing to one.

Here we are interested in the case where $N$ is large. Since the number of elements of this tensor scales exponentially with $N$, it is quickly impossible to store. In cases where there is some structure to the variables, one may use a compact representation of $P$ which exploits this structure, such as Bayesian networks or Markov random fields defined on a graph. In the following we consider models, known as tensor networks, in which a tensor $T$ is factorized into the contraction of multiple smaller tensors. As long as $T$ is non-negative, one can model $P$ as $P=T/Z_T$, where $Z_T=\sum_{X_1,\ldots , X_N} T_{X_1,\ldots, X_N}$
is a normalization factor. For all tensor networks considered in this work, this normalization factor can be evaluated efficiently, as explained in Section~\ref{sec:algo}.

In particular, we define the following tensor networks, in both algebraic and graphical notation. In the diagrams each box represents a tensor and lines emanating from these boxes represent tensor indices. Connecting two lines implies a contraction, which is a summation over the connected index.

\begin{enumerate}
    \item \textbf{Tensor-train/matrix product state (MPS$_\mathbb{F}$)}: A tensor $T$, with $N$ $d$-dimensional indices, admits an MPS$_\mathbb{F}$ representation of TT-rank$_\mathbb{F}$ $r$ when the entries of $T$ can be written as
    \begin{align}
    T_{X_1,\ldots, X_N}\qquad \ \ \ &=\ \ \sum_{\{\alpha_i=1\}}^{r} A^{\alpha_1}_{1,X_1}A^{\alpha_1, \alpha_2}_{2,X_2}\cdots A^{\alpha_{N-2},\alpha_{N-1}}_{N-1,X_{N-1}}A^{\alpha_{N-1}}_{N,X_N}\ ,\\[0.2cm]
    \begin{tikzpicture}[y=0.80pt, x=0.80pt, yscale=-1.200000, xscale=1.200000, inner sep=0pt, outer sep=0pt, baseline={([yshift=-12pt]current bounding box.center)}]
\path[draw=black,line join=miter,line cap=butt,miter limit=4.00,line
  width=0.800pt] (288.1262,456.0122) -- (288.1262,444.4994);
\path[draw=black,line join=miter,line cap=butt,miter limit=4.00,line
  width=0.800pt] (309.1569,456.0122) -- (309.1569,444.4994);
\path[draw=black,line join=miter,line cap=butt,miter limit=4.00,line
  width=0.800pt] (267.0955,456.0122) -- (267.0955,444.4994);
\path[draw=black,line join=miter,line cap=butt,miter limit=4.00,line
  width=0.800pt] (330.1876,456.0122) -- (330.1876,444.4994);
\path[scale=-1.000,draw=black,fill=cffffff,line join=miter,line cap=butt,miter
  limit=4.00,even odd rule,line width=0.800pt,rounded corners=0.0000cm]
  (-337.7193,-471.1178) rectangle (-260.4949,-456.1295);
\node[align=center] at (268,440) {$\scriptstyle X_1$};
\node[align=center] at (330.5,440) {$\scriptstyle X_N$};
\node[align=center] at (300,464) {$T$};
\end{tikzpicture} \quad \ \  &= \quad \ \  \begin{tikzpicture}[y=0.80pt, x=0.80pt, yscale=-1.200000, xscale=1.200000, inner sep=0pt, outer sep=0pt, baseline={([yshift=-12pt]current bounding box.center)}]
\path[draw=black,line join=miter,line cap=butt,miter limit=4.00,line
  width=0.800pt] (451.5867,398.9683) -- (515.3178,398.9683);
\path[draw=black,line join=miter,line cap=butt,miter limit=4.00,line
  width=0.800pt] (377.6601,398.9550) -- (437.9984,398.9550);
\path[cm={{0.04091,0.0,0.0,-0.04091,(437.31587,407.80667)}},fill=black,miter
  limit=4.00,line width=19.553pt]
  (137.3807,216.2084)arc(-0.000:180.000:24.496)arc(-180.000:0.000:24.496) --
  cycle;
\path[cm={{0.04091,0.0,0.0,-0.04091,(442.97543,407.80667)}},fill=black,miter
  limit=4.00,line width=19.553pt]
  (137.3807,216.2084)arc(-0.000:180.000:24.496)arc(-180.000:0.000:24.496) --
  cycle;
\path[draw=black,line join=miter,line cap=butt,miter limit=4.00,line
  width=0.800pt] (417.4428,392.4548) -- (417.4428,380.9420);
\path[xscale=1.000,yscale=-1.000,draw=black,fill=cffffff,line join=miter,line
  cap=butt,miter limit=4.00,even odd rule,line width=0.800pt,rounded
  corners=0.0000cm] (409.6792,-406.6681) rectangle (424.7018,-392.0370);
\path[draw=black,line join=miter,line cap=butt,miter limit=4.00,line
  width=0.800pt] (470.9259,392.4548) -- (470.9259,380.9420);
\path[xscale=1.000,yscale=-1.000,draw=black,fill=cffffff,line join=miter,line
  cap=butt,miter limit=4.00,even odd rule,line width=0.800pt,rounded
  corners=0.0000cm] (463.4089,-406.6681) rectangle (478.4315,-392.0370);
\path[draw=black,line join=miter,line cap=butt,miter limit=4.00,line
  width=0.800pt] (375.9597,392.4548) -- (375.9597,380.9420);
\path[xscale=1.000,yscale=-1.000,draw=black,fill=cffffff,line join=miter,line
  cap=butt,miter limit=4.00,even odd rule,line width=0.800pt,rounded
  corners=0.0000cm] (368.8107,-406.6681) rectangle (383.8333,-392.0370);
\path[draw=black,line join=miter,line cap=butt,miter limit=4.00,line
  width=0.800pt] (514.4090,392.4548) -- (514.4090,380.9420);
\path[xscale=1.000,yscale=-1.000,draw=black,fill=cffffff,line join=miter,line
  cap=butt,miter limit=4.00,even odd rule,line width=0.800pt,rounded
  corners=0.0000cm] (507.4000,-406.6681) rectangle (522.4226,-392.0370);
\node[align=center] at (376,399) {$A_1$};
\node[align=center] at (515,399) {$A_N$};
\node[align=center] at (376,376) {$\scriptstyle X_1$};
\node[align=center] at (515,376) {$\scriptstyle X_N$};
\node[align=center] at (396,393) {$\scriptstyle \alpha_1$};
\node[align=center] at (435,393) {$\scriptstyle \alpha_2$};
\node[align=center] at (492,393) {$\scriptstyle \alpha_{N-1}$};
\end{tikzpicture}\ \ \ ,
    \intertext{where $A_1$ and $A_N$ are $d\times r$ matrices, and $A_i$ are order-$3$ tensors of dimension $d\times r\times r$, with elements in $\mathbb{F}\in\{\mathbb{R}_{\geq 0},\mathbb{R},\mathbb{C}\}$. The indices $\alpha_i$ of these constituent tensors run from $1$ to $r$ and are contracted (summed over) to construct $T$.
    \item \textbf{Born machine (BM$_\mathbb{F}$)}: A tensor $T$, with $N$ $d$-dimensional indices, admits a BM$_\mathbb{F}$ representation of Born-rank$_\mathbb{F}$ $r$ when the entries of $T$ can be written as}
    T_{X_1,\ldots, X_N}\qquad \ \ \ &= \ \ \  \left|\sum_{\{\alpha_i=1\}}^{r} A^{\alpha_1}_{1,X_1}A^{\alpha_1 ,\alpha_2}_{2,X_2}\cdots A^{\alpha_{N-2},\alpha_{N-1}}_{N-1,X_{N-1}}A^{\alpha_{N-1}}_{N,X_N}\right|^2,\\[0.2cm]
    \begin{tikzpicture}[y=0.80pt, x=0.80pt, yscale=-1.200000, xscale=1.200000, inner sep=0pt, outer sep=0pt, baseline={([yshift=-12pt]current bounding box.center)}]
\path[draw=black,line join=miter,line cap=butt,miter limit=4.00,line
  width=0.800pt] (288.1262,456.0122) -- (288.1262,444.4994);
\path[draw=black,line join=miter,line cap=butt,miter limit=4.00,line
  width=0.800pt] (309.1569,456.0122) -- (309.1569,444.4994);
\path[draw=black,line join=miter,line cap=butt,miter limit=4.00,line
  width=0.800pt] (267.0955,456.0122) -- (267.0955,444.4994);
\path[draw=black,line join=miter,line cap=butt,miter limit=4.00,line
  width=0.800pt] (330.1876,456.0122) -- (330.1876,444.4994);
\path[scale=-1.000,draw=black,fill=cffffff,line join=miter,line cap=butt,miter
  limit=4.00,even odd rule,line width=0.800pt,rounded corners=0.0000cm]
  (-337.7193,-471.1178) rectangle (-260.4949,-456.1295);
\node[align=center] at (268,440) {$\scriptstyle X_1$};
\node[align=center] at (330.5,440) {$\scriptstyle X_N$};
\node[align=center] at (300,464) {$T$};
\end{tikzpicture} \quad \ \  &= \quad \ \ \begin{tikzpicture}[y=0.80pt, x=0.80pt, yscale=-1.200000, xscale=1.200000, inner sep=0pt, outer sep=0pt, baseline=(current bounding box.center)]
\path[draw=black,line join=miter,line cap=butt,miter limit=4.00,line
  width=0.800pt] (375.1601,420.0639) -- (438.8912,420.0639);
\path[cm={{0.04091,0.0,0.0,-0.04091,(437.31587,428.91559)}},fill=black,miter
  limit=4.00,line width=19.553pt]
  (137.3807,216.2084)arc(-0.000:180.000:24.496)arc(-180.000:0.000:24.496) --
  cycle;
\path[cm={{0.04091,0.0,0.0,-0.04091,(442.97543,428.91559)}},fill=black,miter
  limit=4.00,line width=19.553pt]
  (137.3807,216.2084)arc(-0.000:180.000:24.496)arc(-180.000:0.000:24.496) --
  cycle;
\path[draw=black,line join=miter,line cap=butt,miter limit=4.00,line
  width=0.800pt] (451.5867,420.2182) -- (515.3178,420.2182);
\path[draw=black,line join=miter,line cap=butt,miter limit=4.00,line
  width=0.800pt] (451.5867,398.9683) -- (515.3178,398.9683);
\path[draw=black,line join=miter,line cap=butt,miter limit=4.00,line
  width=0.800pt] (377.6601,398.9550) -- (437.9984,398.9550);
\path[cm={{0.04091,0.0,0.0,-0.04091,(437.31587,407.80667)}},fill=black,miter
  limit=4.00,line width=19.553pt]
  (137.3807,216.2084)arc(-0.000:180.000:24.496)arc(-180.000:0.000:24.496) --
  cycle;
\path[cm={{0.04091,0.0,0.0,-0.04091,(442.97543,407.80667)}},fill=black,miter
  limit=4.00,line width=19.553pt]
  (137.3807,216.2084)arc(-0.000:180.000:24.496)arc(-180.000:0.000:24.496) --
  cycle;
\path[draw=black,fill=cffffff,line join=miter,line cap=butt,miter
  limit=4.00,even odd rule,line width=0.800pt,rounded corners=0.0000cm]
  (409.8745,412.7626) rectangle (424.8970,427.3937);
\path[draw=black,line join=miter,line cap=butt,miter limit=4.00,line
  width=0.800pt] (417.4428,392.4548) -- (417.4428,380.9420);
\path[xscale=1.000,yscale=-1.000,draw=black,fill=cffffff,line join=miter,line
  cap=butt,miter limit=4.00,even odd rule,line width=0.800pt,rounded
  corners=0.0000cm] (409.6792,-406.6681) rectangle (424.7018,-392.0370);
\path[draw=black,fill=cffffff,line join=miter,line cap=butt,miter
  limit=4.00,even odd rule,line width=0.800pt,rounded corners=0.0000cm]
  (463.6042,412.7626) rectangle (478.6267,427.3937);
\path[draw=black,line join=miter,line cap=butt,miter limit=4.00,line
  width=0.800pt] (470.9259,392.4548) -- (470.9259,380.9420);
\path[xscale=1.000,yscale=-1.000,draw=black,fill=cffffff,line join=miter,line
  cap=butt,miter limit=4.00,even odd rule,line width=0.800pt,rounded
  corners=0.0000cm] (463.4089,-406.6681) rectangle (478.4315,-392.0370);
\path[draw=black,fill=cffffff,line join=miter,line cap=butt,miter
  limit=4.00,even odd rule,line width=0.800pt,rounded corners=0.0000cm]
  (369.0060,412.7626) rectangle (384.0286,427.3937);
\path[draw=black,line join=miter,line cap=butt,miter limit=4.00,line
  width=0.800pt] (375.9597,392.4548) -- (375.9597,380.9420);
\path[xscale=1.000,yscale=-1.000,draw=black,fill=cffffff,line join=miter,line
  cap=butt,miter limit=4.00,even odd rule,line width=0.800pt,rounded
  corners=0.0000cm] (368.8107,-406.6681) rectangle (383.8333,-392.0370);
\path[draw=black,fill=cffffff,line join=miter,line cap=butt,miter
  limit=4.00,even odd rule,line width=0.800pt,rounded corners=0.0000cm]
  (507.5953,412.7626) rectangle (522.6179,427.3937);
\path[draw=black,line join=miter,line cap=butt,miter limit=4.00,line
  width=0.800pt] (514.4090,392.4548) -- (514.4090,380.9420);
\path[xscale=1.000,yscale=-1.000,draw=black,fill=cffffff,line join=miter,line
  cap=butt,miter limit=4.00,even odd rule,line width=0.800pt,rounded
  corners=0.0000cm] (507.4000,-406.6681) rectangle (522.4226,-392.0370);
\path[draw=black,line join=miter,line cap=butt,miter limit=4.00,line
  width=0.800pt] (417.5770,439.4966) -- (417.5770,427.9839);
\path[draw=black,line join=miter,line cap=butt,miter limit=4.00,line
  width=0.800pt] (471.0601,439.4966) -- (471.0601,427.9839);
\path[draw=black,line join=miter,line cap=butt,miter limit=4.00,line
  width=0.800pt] (376.0939,439.4966) -- (376.0939,427.9839);
\path[draw=black,line join=miter,line cap=butt,miter limit=4.00,line
  width=0.800pt] (514.5432,439.4967) -- (514.5432,427.9839);
\node[align=center] at (376,399) {$A_1$};
\node[align=center] at (515,399) {$A_N$};
\node[align=center] at (376,420) {$\xoverline[0.75]{A_1}$};
\node[align=center] at (515,420) {$\xoverline[0.75]{A_N}$};
\node[align=center] at (376,376) {$\scriptstyle X_1$};
\node[align=center] at (515,376) {$\scriptstyle X_N$};
\node[align=center] at (376,445) {$\scriptstyle X_1$};
\node[align=center] at (515,445) {$\scriptstyle X_N$};
\node[align=center] at (396,393) {$\scriptstyle \alpha_1$};
\node[align=center] at (435,393) {$\scriptstyle \alpha_2$};
\node[align=center] at (492,393) {$\scriptstyle \alpha_{N-1}$};
\node[align=center] at (396,426) {$\scriptstyle \alpha'_1$};
\node[align=center] at (435,426) {$\scriptstyle \alpha'_2$};
\node[align=center] at (492,426) {$\scriptstyle \alpha'_{N-1}$};
\end{tikzpicture}\ \ \ ,
    \intertext{with elements of the constituent tensors $A_i$ in $\mathbb{F}\in\{\mathbb{R},\mathbb{C}\}$, i.e., when $T$ admits a representation as the absolute-value squared (element-wise) of an MPS$_\mathbb{F}$ of TT-rank$_\mathbb{F}$ $r$.
    \item \textbf{Locally purified state (LPS$_\mathbb{F}$)}: A tensor $T$, with $N$ $d$-dimensional indices, admits an LPS$_\mathbb{F}$ representation of puri-rank$_\mathbb{F}$ $r$ and purification dimension $\mu$ when the entries of $T$ can be written as}
    \label{eq:LPS}
    T_{X_1,\ldots, X_N}=\sum_{\{\alpha_i,\alpha'_i=1\}}^{r} \sum_{\{\beta_i=1\}}^\mu &A^{\beta_1,\alpha_1}_{1,X_1}\overline{A^{\beta_1,\alpha'_1}_{1,X_1}} A^{\beta_2, \alpha_1 ,\alpha_2}_{2,X_2}\overline{A^{\beta_2, \alpha'_1 ,\alpha'_2}_{2,X_2}}\cdots A^{\beta_{N},\alpha_{N-1}}_{N,X_N}\overline{A^{\beta_{N},\alpha'_{N-1}}_{N,X_N}},\\[0.2cm]
    \begin{tikzpicture}[y=0.80pt, x=0.80pt, yscale=-1.200000, xscale=1.200000, inner sep=0pt, outer sep=0pt, baseline={([yshift=-12pt]current bounding box.center)}]
\path[draw=black,line join=miter,line cap=butt,miter limit=4.00,line
  width=0.800pt] (288.1262,456.0122) -- (288.1262,444.4994);
\path[draw=black,line join=miter,line cap=butt,miter limit=4.00,line
  width=0.800pt] (309.1569,456.0122) -- (309.1569,444.4994);
\path[draw=black,line join=miter,line cap=butt,miter limit=4.00,line
  width=0.800pt] (267.0955,456.0122) -- (267.0955,444.4994);
\path[draw=black,line join=miter,line cap=butt,miter limit=4.00,line
  width=0.800pt] (330.1876,456.0122) -- (330.1876,444.4994);
\path[scale=-1.000,draw=black,fill=cffffff,line join=miter,line cap=butt,miter
  limit=4.00,even odd rule,line width=0.800pt,rounded corners=0.0000cm]
  (-337.7193,-471.1178) rectangle (-260.4949,-456.1295);
\node[align=center] at (268,440) {$\scriptstyle X_1$};
\node[align=center] at (330.5,440) {$\scriptstyle X_N$};
\node[align=center] at (300,464) {$T$};
\end{tikzpicture} \quad &= \quad \ \  \begin{tikzpicture}[y=0.80pt, x=0.80pt, yscale=-1.200000, xscale=1.200000, inner sep=0pt, outer sep=0pt, baseline=(current bounding box.center)]
\path[draw=black,line join=miter,line cap=butt,miter limit=4.00,line
  width=0.800pt] (375.1601,430.0639) -- (438.8912,430.0639);
\path[cm={{0.04091,0.0,0.0,-0.04091,(437.31587,438.91559)}},fill=black,miter
  limit=4.00,line width=19.553pt]
  (137.3807,216.2084)arc(-0.000:180.000:24.496)arc(-180.000:0.000:24.496) --
  cycle;
\path[cm={{0.04091,0.0,0.0,-0.04091,(442.97543,438.91559)}},fill=black,miter
  limit=4.00,line width=19.553pt]
  (137.3807,216.2084)arc(-0.000:180.000:24.496)arc(-180.000:0.000:24.496) --
  cycle;
\path[draw=black,line join=miter,line cap=butt,miter limit=4.00,line
  width=0.800pt] (451.5867,430.2182) -- (515.3178,430.2182);
\path[draw=black,line join=miter,line cap=butt,miter limit=4.00,line
  width=0.800pt] (451.5867,398.9683) -- (515.3178,398.9683);
\path[draw=black,line join=miter,line cap=butt,miter limit=4.00,line
  width=0.800pt] (377.6601,398.9550) -- (437.9984,398.9550);
\path[cm={{0.04091,0.0,0.0,-0.04091,(437.31587,407.80667)}},fill=black,miter
  limit=4.00,line width=19.553pt]
  (137.3807,216.2084)arc(-0.000:180.000:24.496)arc(-180.000:0.000:24.496) --
  cycle;
\path[cm={{0.04091,0.0,0.0,-0.04091,(442.97543,407.80667)}},fill=black,miter
  limit=4.00,line width=19.553pt]
  (137.3807,216.2084)arc(-0.000:180.000:24.496)arc(-180.000:0.000:24.496) --
  cycle;
\path[draw=black,line join=miter,line cap=butt,line width=0.800pt]
  (417.4461,400.9568) -- (417.4461,425.1532);
\path[draw=black,fill=cffffff,line join=miter,line cap=butt,miter
  limit=4.00,even odd rule,line width=0.800pt,rounded corners=0.0000cm]
  (409.8745,422.7626) rectangle (424.8970,437.3937);
\path[draw=black,line join=miter,line cap=butt,miter limit=4.00,line
  width=0.800pt] (417.4428,392.4548) -- (417.4428,380.9420);
\path[xscale=1.000,yscale=-1.000,draw=black,fill=cffffff,line join=miter,line
  cap=butt,miter limit=4.00,even odd rule,line width=0.800pt,rounded
  corners=0.0000cm] (409.6792,-406.6681) rectangle (424.7018,-392.0370);
\path[draw=black,line join=miter,line cap=butt,line width=0.800pt]
  (471.1246,401.4925) -- (471.1246,425.6889);
\path[draw=black,fill=cffffff,line join=miter,line cap=butt,miter
  limit=4.00,even odd rule,line width=0.800pt,rounded corners=0.0000cm]
  (463.6042,422.7626) rectangle (478.6267,437.3937);
\path[draw=black,line join=miter,line cap=butt,miter limit=4.00,line
  width=0.800pt] (470.9259,392.4548) -- (470.9259,380.9420);
\path[xscale=1.000,yscale=-1.000,draw=black,fill=cffffff,line join=miter,line
  cap=butt,miter limit=4.00,even odd rule,line width=0.800pt,rounded
  corners=0.0000cm] (463.4089,-406.6681) rectangle (478.4315,-392.0370);
\path[draw=black,line join=miter,line cap=butt,line width=0.800pt]
  (375.8746,401.5371) -- (375.8746,425.7336);
\path[draw=black,fill=cffffff,line join=miter,line cap=butt,miter
  limit=4.00,even odd rule,line width=0.800pt,rounded corners=0.0000cm]
  (369.0060,422.7626) rectangle (384.0286,437.3937);
\path[draw=black,line join=miter,line cap=butt,miter limit=4.00,line
  width=0.800pt] (375.9597,392.4548) -- (375.9597,380.9420);
\path[xscale=1.000,yscale=-1.000,draw=black,fill=cffffff,line join=miter,line
  cap=butt,miter limit=4.00,even odd rule,line width=0.800pt,rounded
  corners=0.0000cm] (368.8107,-406.6681) rectangle (383.8333,-392.0370);
\path[draw=black,line join=miter,line cap=butt,line width=0.800pt]
  (514.9639,401.0461) -- (514.9639,425.2425);
\path[draw=black,fill=cffffff,line join=miter,line cap=butt,miter
  limit=4.00,even odd rule,line width=0.800pt,rounded corners=0.0000cm]
  (507.5953,422.7626) rectangle (522.6179,437.3937);
\path[draw=black,line join=miter,line cap=butt,miter limit=4.00,line
  width=0.800pt] (514.4090,392.4548) -- (514.4090,380.9420);
\path[xscale=1.000,yscale=-1.000,draw=black,fill=cffffff,line join=miter,line
  cap=butt,miter limit=4.00,even odd rule,line width=0.800pt,rounded
  corners=0.0000cm] (507.4000,-406.6681) rectangle (522.4226,-392.0370);
\path[draw=black,line join=miter,line cap=butt,miter limit=4.00,line
  width=0.800pt] (417.5770,449.4966) -- (417.5770,437.9839);
\path[draw=black,line join=miter,line cap=butt,miter limit=4.00,line
  width=0.800pt] (471.0601,449.4966) -- (471.0601,437.9839);
\path[draw=black,line join=miter,line cap=butt,miter limit=4.00,line
  width=0.800pt] (376.0939,449.4966) -- (376.0939,437.9839);
\path[draw=black,line join=miter,line cap=butt,miter limit=4.00,line
  width=0.800pt] (514.5432,449.4967) -- (514.5432,437.9839);
\node[align=center] at (376,399) {$A_1$};
\node[align=center] at (515,399) {$A_N$};
\node[align=center] at (376,430) {$\xoverline[0.75]{A_1}$};
\node[align=center] at (515,430) {$\xoverline[0.75]{A_N}$};
\node[align=center] at (376,376) {$\scriptstyle X_1$};
\node[align=center] at (515,376) {$\scriptstyle X_N$};
\node[align=center] at (376,455) {$\scriptstyle X_1$};
\node[align=center] at (515,455) {$\scriptstyle X_N$};
\node[align=center] at (396,393) {$\scriptstyle \alpha_1$};
\node[align=center] at (435,393) {$\scriptstyle \alpha_2$};
\node[align=center] at (492,393) {$\scriptstyle \alpha_{N-1}$};
\node[align=center] at (396,436) {$\scriptstyle \alpha'_1$};
\node[align=center] at (435,436) {$\scriptstyle \alpha'_2$};
\node[align=center] at (492,436) {$\scriptstyle \alpha'_{N-1}$};
\node[align=center] at (384,414) {$\scriptstyle \beta_1$};
\node[align=center] at (425,414) {$\scriptstyle \beta_2$};
\node[align=center] at (524,414) {$\scriptstyle \beta_{N}$};
\end{tikzpicture}\ \ ,
    \end{align} 
    where $A_1$ and $A_N$ are order-$3$ tensors of dimension $d \times \mu \times r$ and $A_i$ are order-$4$ tensors of dimension $d \times \mu \times r \times r$. The indices $\alpha_i$ run from 1 to $r$, the indices $\beta_i$ run from 1 to $\mu$, and both are contracted to construct $T$. Without loss of generality we can consider only $\mu \leq rd^2$.
\end{enumerate}

Note that all the representations defined above yield non-negative tensors by construction, except for MPS$_{\mathbb{R}/\mathbb{C}}$. In this work, 
we consider only the subset of MPS$_{\mathbb{R}/\mathbb{C}}$ which represent non-negative tensors.

Given a non-negative tensor $T$ we define the TT-rank$_\mathbb{F}$ (Born-rank$_\mathbb{F}$) of $T$ as the minimal $r$ such that $T$ admits an MPS$_{\mathbb{F}}$ (BM$_{\mathbb{F}}$) representation of TT-rank$_\mathbb{F}$ (Born-rank$_\mathbb{F}$) $r$. We define the puri-rank$_\mathbb{F}$ of $T$ as the minimal $r$ such that $T$ admits an LPS$_{\mathbb{F}}$ representation of puri-rank$_\mathbb{F}$ $r$, for some purification dimension $\mu$. We note that if we consider tensors $T$ with 2 $d$-dimensional indices (i.e., matrices) then the TT-rank$_{\mathbb{R}_{\geq 0}}$ is the non-negative rank, i.e., the smallest $k$ such that $T$ can be written as
$T=AB$ with $A$ being $d\times k$ and $B$ being $k\times d$ matrices with real non-negative entries. The TT-rank$_{\mathbb{R}/\mathbb{C}}$ is the conventional matrix rank, the Born-rank$_{\mathbb{R}}$ (Born-rank$_{\mathbb{C}}$) is the real (complex) Hadamard square-root rank, i.e., 
the minimal rank of a real (complex) entry-wise square root of $T$, and finally the puri-rank$_{\mathbb{R}}$ (puri-rank$_{\mathbb{C}}$) is the real (complex) positive semidefinite rank \cite{Fawzi2015}. These abbreviations, definitions and relations are summarized in Table~\ref{table:defns} below, where we use the notations of ref.~\cite{Fawzi2015} for the different matrix ranks.

\begin{table}[h!]
\centering
\caption{Summary of notations for the different tensor-network representations and their ranks.}\label{table:defns}
\begin{tabular}{ l  c  c  c  c  }
  \toprule			
    Tensor representation & MPS$_{\mathbb{R}\geq 0}$ & MPS$_{\mathbb{R}/\mathbb{C}}$ & BM$_{\mathbb{R}/\mathbb{C}}$ & LPS$_{\mathbb{R}/\mathbb{C}}$ \\
  Tensor rank & TT-rank$_{\mathbb{R}\geq 0}$ & TT-rank$_{\mathbb{R}/\mathbb{C}}$ & Born-rank$_{\mathbb{R}/\mathbb{C}}$ & puri-rank$_{\mathbb{R}/\mathbb{C}}$ \\
  Matrix rank \cite{Fawzi2015} & rank$_{+}$ & rank & rank$_{\mathbb{R}/\mathbb{C}\sqrt{}}$ & rank${_{\mathbb{R}/\mathbb{C},psd}}$ \\
  \bottomrule
\end{tabular}
\end{table}

For a given rank and a given tensor network, there is a set of non-negative tensors that can be exactly represented, and as the rank is increased, this set grows. In the limit of arbitrarily large rank, all tensor networks we consider can represent any non-negative tensor. This work is concerned with the relative expressive power of these different tensor-network representations, i.e. how do these representable sets compare for different tensor networks. This will be characterized in Section~\ref{sec:expressivity2} in terms of the different ranks needed by different tensor networks to represent a non-negative tensor.

\section{Relationship to hidden Markov models and quantum circuits}
\label{sec:hmmquantum}

In order to provide context for the factorizations introduced in Section~\ref{sec:ansatz}, we show here how they are related to other representations of probability distributions based on probabilistic graphical models and quantum circuits. In particular, we show that there is a mapping between hidden Markov models with constant number of hidden units per variable and MPS$_{\mathbb{R} \geq 0}$ with constant TT-rank$_{\mathbb{R} \geq 0}$, as well as between local quantum circuits of fixed depth and Born machines of constant Born-rank$_\mathbb{C}$. These relations imply that results on the expressive power of the former directly provide results on the expressive power of the latter.

\subsection{Hidden Markov models are non-negative matrix product states}

\begin{figure}[ht]
\centering
\includegraphics[width=0.95\linewidth]{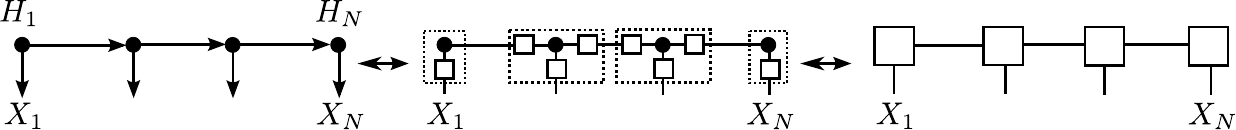}
\caption{Mapping between a HMM and a non-negative MPS.}
\label{fig:MPSHMMC}
\end{figure}

Consider a hidden Markov model (HMM) with observed variables $\{X_i\}$ taking values in $\{1,\ldots,d\}$ and hidden variables $\{H_i\}$ taking values in $\{1,\ldots,r\}$ (Fig.~\ref{fig:MPSHMMC}). The probability of the observed variables may be expressed as
\begin{align}
P(X_1,\ldots, X_N)=\sum_{H_1,\ldots , H_N} P(X_1|H_1)\prod_{i=2}^N P(H_i|H_{i-1})P(X_i|H_i).
\end{align}
Notice that $P(H_i|H_{i-1})$ and $P(X_i|H_i)$ are matrices with non-negative elements, as depicted in the factor graph in the central diagram of Fig.~\ref{fig:MPSHMMC}. Now define the tensors $A_{1,l}^{j}=P(X_i=l|H_1=j)$, and  $A_{i,l}^{jk}=P(H_i=k|H_{i-1}=j)P(X_i=l|H_i=k).$ Then the MPS with TT-rank$_{\mathbb{R}_{\geq 0}}=r$ defined with tensors $A_i$ defines the same probability distribution on the observed variables as the HMM.

Conversely, given an MPS$_{\mathbb{R} \geq 0}$ with TT-rank$_{\mathbb{R}_{\geq 0}}=r$, there exists an HMM, with hidden variables of dimension $r'\leq \min(dr,r^2)$, defining the same probability mass function, as shown in the supplementary material. We note also that by using a different graph for the HMM, it is possible to construct an equivalent HMM with hidden variables of dimension $r$ \cite{Robeva2017,Glasser2018}. As such, any results on expressivity derived for MPS$_{\mathbb{R}_{\geq 0}}$ hold also for HMM.

\subsection{Quantum circuits are Born machines or locally purified states}

In this section we provide a brief description of the connection between Born machines, locally purified states and quantum circuits, assuming some prior knowledge of the formalism of quantum computing. A more introductory presentation of this connection, assuming no knowledge of quantum mechanics and quantum computing, is contained in the supplementary material.

Consider a 2-local quantum circuit of depth $D$ acting on $N$ $d$-dimensional qudits arranged along a line, with fixed orthonormal basis $\{|X_i\rangle\}_{X = 1}^d$ for each local Hilbert space. The output state of the quantum circuit can be written as
\begin{equation}
    |\psi\rangle = \sum_{X_1=1}^d\ldots\sum_{X_N=1}^d \psi(X_1,\ldots,X_N)|X_1\rangle\otimes \ldots \otimes|X_N\rangle.
\end{equation}
Furthermore, the amplitudes of this state are given by the entries of an MPS with TT-rank$_{\mathbb{C}}$ less than $d^{D+1}$. More specifically, as shown in Equation~\eqref{eq;circuit_to_mps} below (explicitly for the case $N=4$), by starting from the circuit diagram, reshaping and splitting each gate via a singular value decomposition, and then contracting the resulting diagram as indicated, 
one finds that
\begin{align}\label{eq;circuit_to_mps}
\psi(X_1,\ldots,X_4)= \vcenter{\hbox{\includegraphics[height=0.2\linewidth]{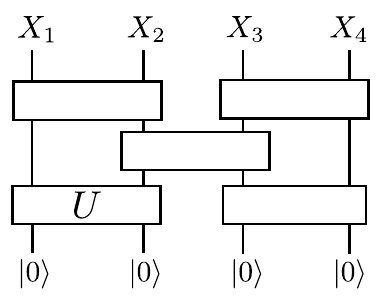}}}
=\vcenter{\hbox{\includegraphics[height=0.2\linewidth]{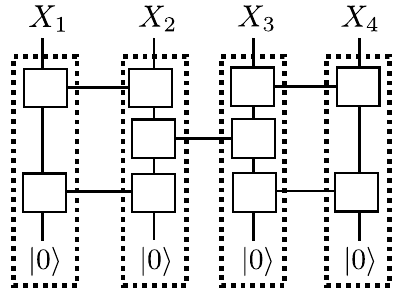}}}
=\raisebox{0.2\height}{$\vcenter{\hbox{\includegraphics[height=0.2\linewidth]{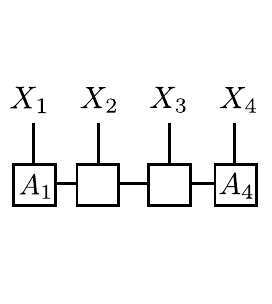}}}$}.
\end{align}
The probability of outcome $X_1,\ldots,X_4$ when performing a measurement in the specified basis is obtained from the Born rule. It is thus given by the Born machine defined from the MPS representation of the quantum circuit, i.e.,
\begin{align}\label{eq:mps_to_bm}
P(X_1,\ldots,X_4)=\left|\psi(X_1,\ldots,X_4)\right|^2=\vcenter{\hbox{\includegraphics[height=0.2\linewidth]{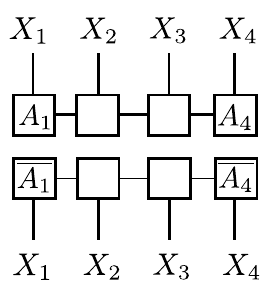}}}.
\end{align}
By performing measurements on the output state of a $2$-local quantum circuit, we are therefore effectively sampling from the probability mass function of $N$ discrete $d$-dimensional random variables $\{X_i\}$ which is given by the Born machine defined from the MPS representation of the quantum circuit, as shown in Equation \eqref{eq:mps_to_bm}. Given this correspondence, any results on the expressive power of Born machines hold also for local quantum circuits, when considered as probabilistic models via the Born rule.

In order to understand the correspondence between local quantum circuits and locally purified states, note that if we consider each second qudit as an ancilla, or ``hidden'' qudit, then the probability of outcome $X_1,X_2$ when measuring only the visible qudits is given by the marginal
\begin{align}\label{eq:mps_to_lps}
P(X_1,X_2)=\sum_{H_1,H_2}P(X_1,H_1,X_2,H_2)= \vcenter{\hbox{\includegraphics[height=0.2\linewidth]{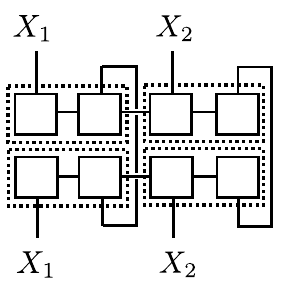}}}= \vcenter{\hbox{\includegraphics[height=0.2\linewidth]{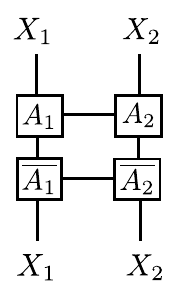}}}.
\end{align}
In particular, we see that by performing such measurements we are sampling from the probability mass function given by the locally purified state which is obtained from the MPS representation of the circuit via the contractions indicated in Equation~\eqref{eq:mps_to_lps}. Once again, this correspondence implies that any results on the expressive power of locally purified states hold also for local quantum circuits with alternating visible and hidden qudits.



\section{Expressive power of tensor-network representations}
\label{sec:expressivity2}

In this section we present various relationships between the expressive power of all representations, which constitute the primary results of this work. The proofs of the propositions in this section can be found in the supplementary material.

\begin{figure}[ht]
\centering
\includegraphics[width = 0.4\linewidth]{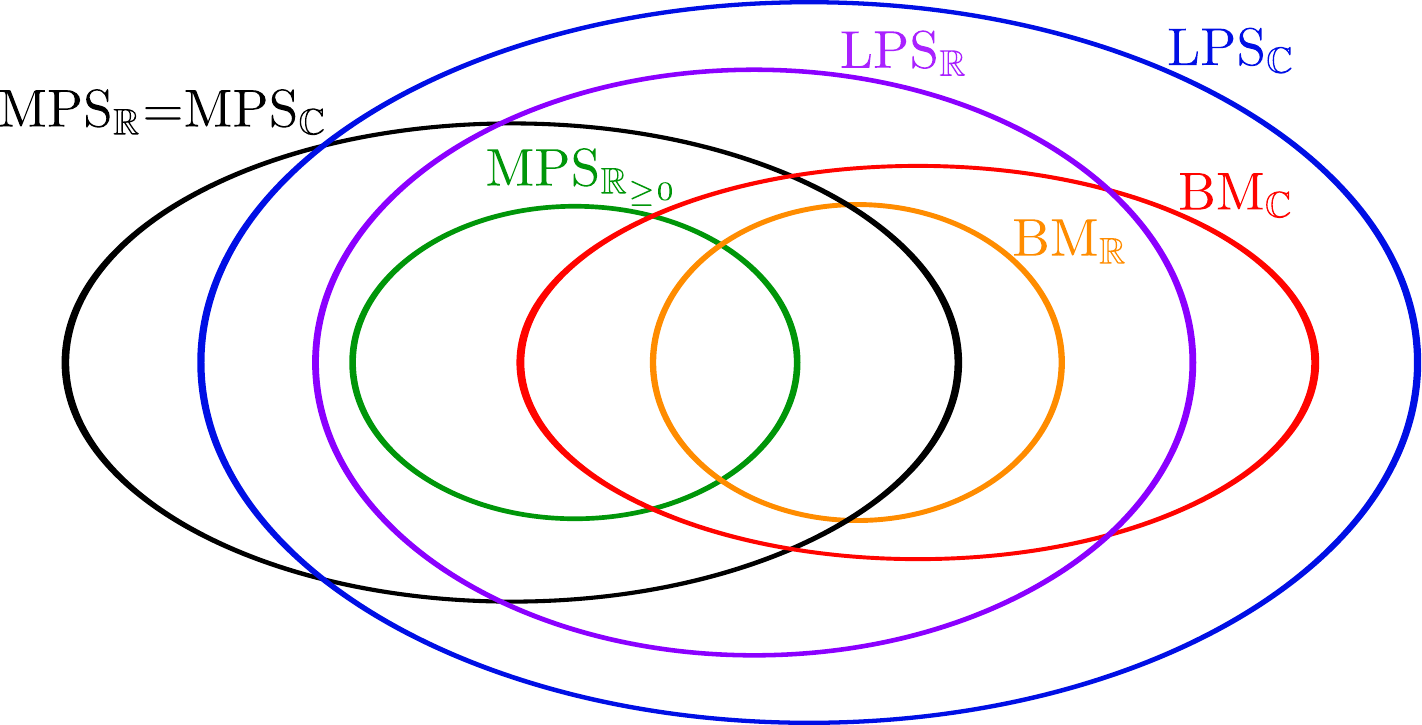} 
\caption{Representation of the sets of non-negative tensors that admit a given tensor-network factorization. In this figure we fix the different ranks of the different tensor networks to be equal.}
\label{fig:setsofmps}
\end{figure}

For a given rank, there is a set of non-negative tensors that can be exactly represented by a given tensor network. These sets are represented in Fig.~\ref{fig:setsofmps} for the case in which the ranks of the tensor networks are equal. When one set is included in another, it means that for every non-negative tensor, the rank of one of the tensor-network factorizations is always greater than or equal to the rank of the other factorization. The inclusion relationships between these sets can therefore be characterized in terms of inequalities between the ranks, as detailed in Proposition~\ref{prop:inclusions}.

\begin{proposition}
\label{prop:inclusions}
For all non-negative tensors 
$\text{TT-rank}_{\mathbb{R}_{\geq 0}} \geq \text{TT-rank}_{\mathbb{R}}$, $\text{Born-rank}_{\mathbb{R}} \geq \text{Born-rank}_{\mathbb{C}}$,
$\text{Born-rank}_{\mathbb{R}} \geq \text{puri-rank}_{\mathbb{R}}$, 
$\text{Born-rank}_{\mathbb{C}} \geq \text{puri-rank}_{\mathbb{C}}$, $\text{puri-rank}_{\mathbb{R}}\geq \text{puri-rank}_{\mathbb{C}}$,
$\text{TT-rank}_{\mathbb{R}_{\geq 0}} \geq \text{puri-rank}_{\mathbb{R}}$,
$\text{TT-rank}_{\mathbb{R}}=\text{TT-rank}_{\mathbb{C}}$.
\end{proposition}
Next, as detailed in Proposition~\ref{prop:strict_inclusions}, and summarized in Table~\ref{table:sets}, we continue by showing that all the inequalities of Proposition~\ref{prop:inclusions} can in fact be strict, and that for all other pairs of representations there exist probability distributions showing that neither rank can always be lower than the other. This shows that neither of the two corresponding sets of tensors can be included in the other. The main new result is the introduction of a matrix with non-negative rank strictly smaller than its complex Hadamard square-root rank, i.e. $\text{TT-rank}_{\mathbb{R}_{\geq 0}}<\text{Born-rank}_{\mathbb{C}}$.

\begin{proposition}
\label{prop:strict_inclusions}
The ranks of all introduced tensor-network representations satisfy the properties contained in Table~\ref{table:sets}. Specifically, denoting by $r_{\text{row}}$ ($r_{\text{column}}$) the rank appearing in the row (column), $<$ indicates that there exists a tensor satisfying $r_{\text{row}} < r_{\text{column}}$ and $<,>$ indicates that there exists both a tensor satisfying $r_{\text{row}}  < r_{\text{column}}$ and another tensor satisfying $r_{\text{column}} > r_{\text{row}}$.
\end{proposition}
\begin{table}[ht]
\caption{Results of Proposition~\ref{prop:strict_inclusions}}
\label{table:sets}
\centering
\begin{tabular}{@{} lcccccc @{}}
\toprule
& TT-rank$_{\mathbb{R}}$ & TT-rank$_{\mathbb{R}_{\geq 0}}$  & Born-rank$_{\mathbb{R}}$ & Born-rank$_{\mathbb{C}}$ & puri-rank$_{\mathbb{R}}$ & puri-rank$_{\mathbb{C}}$  \\
\midrule
TT-rank$_{\mathbb{R}}$ & = & $<$ & $<,>$ & $<,>$& $<,>$& $<,>$\\
TT-rank$_{\mathbb{R}_{\geq 0}}$ & $>$ & = &$<,>$  & $<,>$ & $>$ & $>$\\
Born-rank$_{\mathbb{R}}$  & $<,>$ & $<,>$ & = & $>$ & $>$ & $>$\\
Born-rank$_{\mathbb{C}}$  & $<,>$ & $<,>$ & $<$ & =& $<,>$ & $>$\\
puri-rank$_{\mathbb{R}}$ & $<,>$ & $<$ & $<$ &$<,>$ & = & $>$\\
puri-rank$_{\mathbb{C}}$ & $<,>$ & $<$ & $<$ & $<$ & $<$ &= \\
\bottomrule
\end{tabular}
\end{table}

We now answer the question: By how much do we need to increase the rank of a tensor network such that the set of tensors it can represent includes the set of tensors that can be represented by a different tensor network of a different rank? More specifically, consider a tensor that has rank $r$ according to one representation and rank $r'$ according to another. Can we bound the rank $r$ as a function of the rank $r'$ only? The results of Proposition \ref{prop:overheads_1}, presented via Table~\ref{table:ranks}, indicate that in many cases there is no such function - i.e. there exists a family of non-negative tensors, describing a family of probability distributions over $N$ binary variables, with the property that as $N$ goes to infinity $r'$ remains constant, while $r$ also goes to infinity.
\begin{proposition}
\label{prop:overheads_1}
The ranks of all introduced tensor-network representations satisfy the relationships without asterisk contained in Table~\ref{table:ranks}. A function $g(x)$ denotes that for all non-negative tensors $r_{\text{row}}\leq g(r_{\text{column}})$. ``No'' indicates that there exists a family of probability distributions of increasing $N$ with $d=2$ and $r_{\text{column}}$ constant, but such that $r_{\text{row}}$ goes to infinity, i.e. that no such function can exist.
\end{proposition}

\begin{table}[ht]
\caption{Results of Proposition~\ref{prop:overheads_1}.}
\label{table:ranks}
\centering
\begin{tabular}{@{} lcccccc @{}}
\toprule
& TT-rank$_{\mathbb{R}}$ & TT-rank$_{\mathbb{R}_{\geq 0}}$  & Born-rank$_{\mathbb{R}}$ & Born-rank$_{\mathbb{C}}$ & puri-rank$_{\mathbb{R}}$ & puri-rank$_{\mathbb{C}}$  \\
\midrule
TT-rank$_{\mathbb{R}}$ & = & $\leq x $ & $\leq x^2$ & $\leq x^2$& $\leq x^2$& $\leq x^2$\\
TT-rank$_{\mathbb{R}_{\geq 0}}$ & No & = & No  & No &  No  &  No \\
Born-rank$_{\mathbb{R}}$  &  No  & No  & = & No & No & No \\
Born-rank$_{\mathbb{C}}$  & No & No$^*$ & $\leq x$ & =& No$^*$ & No$^*$\\
puri-rank$_{\mathbb{R}}$ & No & $\leq x$ & $\leq x$ &$\leq 2x$ & = & $\leq 2x$\\
puri-rank$_{\mathbb{C}}$ & No & $\leq x$ & $\leq x$ & $\leq x$ & $\leq x$ & =\\
\bottomrule
\end{tabular}
\end{table}

We conjecture that the relationships with an asterisk in Table~\ref{table:ranks} also hold. The existence of a family of matrices with constant non-negative rank but unbounded complex Hadamard square-root rank, together with the techniques introduced in the supplementary material, would provide a proof of these conjectured results. 
Proposition~\ref{prop:overheads_1} indicates the existence of various families of non-negative tensors for which the rank of one representation remains constant, while the rank of another representation grows with the number of binary variables, however, the rate of this growth is not given. The following propositions provide details of the asymptotic growth of these ranks.

\begin{proposition}[\cite{Cuevas_2013}]\label{prop4} There exists a family of non-negative tensors over $2N$ binary variables and constant TT-rank$_\mathbb{R}$=3 that have puri-rank$_{\mathbb{C}}=\Omega(N)$, and hence also puri-rank$_{\mathbb{C}}$, Born-rank$_{\mathbb{R}/\mathbb{C}}$ and TT-rank$_{\mathbb{R}_{\geq 0}} \geq \Omega(N)$.
\end{proposition}

\begin{proposition}
There exists a family of non-negative tensors over $2N$ binary variables and constant TT-rank$_{\mathbb{R}_{\geq 0}}$=2 (and hence also puri-rank$_{\mathbb{R}/\mathbb{C}}=2$) that have Born-rank$_{\mathbb{R}} \geq \pi(2^{N+1})$, where $\pi(x)$ is the number of prime numbers up to $x$, which asymptotically satisfies $\pi(x)\sim x/\log(x)$.
\end{proposition}

\begin{proposition}
There exists a family of non-negative tensors over $2N$ binary variables and constant Born-rank$_{\mathbb{R}}$=2 (and hence also constant Born-rank$_{\mathbb{C}}$ and puri-rank$_{\mathbb{R}/\mathbb{C}}$) that have TT-rank$_{\mathbb{R}_{\geq 0}} \geq N$.
\end{proposition}

\begin{proposition}
There exists a family of non-negative tensors over 2$N$ binary variables and constant Born-rank$_{\mathbb{C}}$=2 that have Born-rank$_{\mathbb{R}}\geq N$.
\end{proposition}

Some comments and observations which may aid in facilitating an intuitive understanding of these results are as follows: Cancellations between negative contributions allow an MPS$_\mathbb{R}$ to represent a non-negative tensor while having lower rank than an MPS$_{\mathbb{R}_{\geq 0}}$ (this separation can also be derived from the separation between Arithmetic Circuits and Monotone Arithmetic Circuits \cite{DBLP:journals/fttcs/ShpilkaY10}). The separations between MPS$_{\mathbb{R}_{\geq 0}}$ and BM$_{\mathbb{R}/\mathbb{C}}$ are due to the difference of rank between probability distributions and their real or complex square roots. 
Finally, the difference between real and complex BM is due to the way in which real and imaginary elements are combined through the modulus squared, and this is illustrated well by the fact that real LPS of purification dimension 2 include complex BM.

As the techniques via which the results of Proposition~\ref{prop:overheads_1} have been obtained are of interest, we provide a sketch of the proof for all ``No'' entries here. Assume that for a given pair of representations there exists a family of non-negative matrices with the property that the rank $r_{\text{column}}$ of one representation remains constant as a function of matrix dimension, while the rank $r_{\text{row}}$ of the other representation grows. Now, consider such a matrix $M$ of dimension $2^N\times2^N$. The first step is to show that $M$ can be unfolded into a tensor network of constant rank $r_{\text{column}}$, for $2N$ binary variables, such that $M$ is a reshaping of the central bipartition of this 
tensor as
\begin{align}
    M=\begin{tikzpicture}[y=0.80pt, x=0.80pt, yscale=-0.700000, xscale=0.700000, inner sep=0pt, outer sep=0pt,baseline={([yshift=-12pt]current bounding box.center)}]
  \begin{scope}[rotate around={180.0:(272.71205,323.00155)},miter limit=4.00,line width=0.800pt]
    \path[draw=black,line join=miter,line cap=butt,miter limit=4.00,line
      width=0.800pt] (325.3029,282.7447) -- (350.7415,282.7447);
  \end{scope}
  \path[draw=black,line join=miter,line cap=butt,miter limit=4.00,line
    width=2.400pt] (197.0625,356.7363) -- (197.0625,344.9012);
  \path[scale=-1.000,draw=black,fill=cffffff,line join=miter,line cap=butt,miter
    limit=4.00,even odd rule,line width=0.800pt,rounded corners=0.0000cm]
    (-227.5104,-370.9657) rectangle (-212.4879,-356.3346);
  \path[scale=-1.000,draw=black,fill=cffffff,line join=miter,line cap=butt,miter
    limit=4.00,even odd rule,line width=0.800pt,rounded corners=0.0000cm]
    (-204.6419,-370.9657) rectangle (-189.6194,-356.3346);
  \path[draw=black,line join=miter,line cap=butt,miter limit=4.00,line
    width=2.400pt] (219.6839,356.6695) -- (219.6839,344.8343);
  \path[fill=black] (190.4093,339.1512) node[above right] (text1781-3) {${\scriptstyle 2^N}$};
  \path[fill=black] (214.7439,339.5204) node[above right] (text1781-8-6) {${\scriptstyle 2^N}$};
\end{tikzpicture}=\begin{tikzpicture}[y=0.80pt, x=0.80pt, yscale=-0.700000, xscale=0.700000, inner sep=0pt, outer sep=0pt,baseline={([yshift=-12pt]current bounding box.center)}]
  \begin{scope}[cm={{3.77953,0.0,0.0,3.77953,(-71.86333,163.57119)}}]
    \begin{scope}[cm={{-0.26458,0.0,0.0,-0.26458,(181.44801,127.56014)}},miter limit=4.00,line width=0.800pt]
      \path[draw=black,line join=miter,line cap=butt,miter limit=4.00,line
        width=0.800pt] (325.3029,282.7447) -- (350.7415,282.7447);
    \end{scope}
    \path[draw=black,line join=miter,line cap=butt,miter limit=4.00,line
      width=0.212pt] (89.2773,51.0250) -- (89.2773,47.8936);
    \path[scale=-1.000,draw=black,fill=cffffff,line join=miter,line cap=butt,miter
      limit=4.00,even odd rule,line width=0.212pt,rounded corners=0.0000cm]
      (-97.3334,-54.7898) rectangle (-93.3586,-50.9187);
    \path[scale=-1.000,draw=black,fill=cffffff,line join=miter,line cap=butt,miter
      limit=4.00,even odd rule,line width=0.212pt,rounded corners=0.0000cm]
      (-91.2827,-54.7898) rectangle (-87.3080,-50.9187);
    \path[draw=black,line join=miter,line cap=butt,miter limit=4.00,line
      width=0.212pt] (89.8417,51.0250) -- (89.8417,47.8936);
    \path[draw=black,line join=miter,line cap=butt,miter limit=4.00,line
      width=0.212pt] (88.7078,51.0250) -- (88.7078,47.8936);
    \path[draw=black,line join=miter,line cap=butt,miter limit=4.00,line
      width=0.212pt] (88.1644,51.0250) -- (88.1644,47.8936);
    \path[draw=black,line join=miter,line cap=butt,miter limit=4.00,line
      width=0.212pt] (90.3614,51.0250) -- (90.3614,47.8936);
    \path[draw=black,line join=miter,line cap=butt,miter limit=4.00,line
      width=0.212pt] (95.2626,51.0073) -- (95.2626,47.8759);
    \path[draw=black,line join=miter,line cap=butt,miter limit=4.00,line
      width=0.212pt] (95.8270,51.0073) -- (95.8270,47.8759);
    \path[draw=black,line join=miter,line cap=butt,miter limit=4.00,line
      width=0.212pt] (94.6930,51.0073) -- (94.6930,47.8759);
    \path[draw=black,line join=miter,line cap=butt,miter limit=4.00,line
      width=0.212pt] (94.1497,51.0073) -- (94.1497,47.8759);
    \path[draw=black,line join=miter,line cap=butt,miter limit=4.00,line
      width=0.212pt] (96.3467,51.0073) -- (96.3467,47.8759);
  \end{scope}
  \path[fill=black] (259.9811,339.0149) node[above right] (text1781) {${\scriptstyle 2^N}$};
  \path[fill=black] (283.2443,339.2056) node[above right] (text1781-8) {${\scriptstyle 2^N}$};
\end{tikzpicture}=\begin{tikzpicture}[y=0.80pt, x=0.80pt, yscale=-0.700000, xscale=0.700000, inner sep=0pt, outer sep=0pt,baseline={([yshift=-14pt]current bounding box.center)}]
    \path[draw=black,dash pattern=on 0.80pt off 1.60pt,line join=miter,line
      cap=butt,miter limit=4.00,line width=0.800pt] (420.7248,334.8708) --
      (420.7248,380.2279);
    \begin{scope}[cm={{3.77953,0.0,0.0,3.77953,(112.51054,96.63679)}}]
      \path[draw=black,line join=miter,line cap=butt,miter limit=4.00,line
        width=0.212pt] (71.8546,70.8087) -- (65.0572,70.8087);
      \path[scale=-1.000,draw=black,fill=cffffff,line join=miter,line cap=butt,miter
        limit=4.00,even odd rule,line width=0.212pt,rounded corners=0.0000cm]
        (-70.4742,-72.7854) rectangle (-66.4995,-68.9142);
      \path[draw=black,line join=miter,line cap=butt,miter limit=4.00,line
        width=0.212pt] (68.4051,69.0001) -- (68.4051,65.8687);
      \path[draw=black,line join=miter,line cap=butt,miter limit=4.00,line
        width=0.212pt] (83.0290,70.8064) -- (74.7914,70.8064);
      \path[xscale=-1.000,yscale=1.000,fill=black,miter limit=4.00,line width=0.212pt]
        (-73.8823,70.8048) circle (0.0075cm);
      \path[xscale=-1.000,yscale=1.000,fill=black,miter limit=4.00,line width=0.212pt]
        (-72.7156,70.8048) circle (0.0075cm);
      \path[scale=-1.000,draw=black,fill=cffffff,line join=miter,line cap=butt,miter
        limit=4.00,even odd rule,line width=0.212pt,rounded corners=0.0000cm]
        (-80.0694,-72.8456) rectangle (-76.0947,-68.9744);
      \path[draw=black,line join=miter,line cap=butt,miter limit=4.00,line
        width=0.212pt] (78.0003,69.0603) -- (78.0003,65.9289);
      \path[draw=black,line join=miter,line cap=butt,miter limit=4.00,line
        width=0.212pt] (62.1198,70.8044) -- (57.5524,70.8044);
      \path[scale=-1.000,draw=black,fill=cffffff,line join=miter,line cap=butt,miter
        limit=4.00,even odd rule,line width=0.212pt,rounded corners=0.0000cm]
        (-60.7393,-72.7811) rectangle (-56.7646,-68.9099);
      \path[draw=black,line join=miter,line cap=butt,miter limit=4.00,line
        width=0.212pt] (58.7120,68.9708) -- (58.7120,65.8394);
      \path[xscale=-1.000,yscale=1.000,fill=black,miter limit=4.00,line width=0.212pt]
        (-64.1892,70.7755) circle (0.0075cm);
      \path[xscale=-1.000,yscale=1.000,fill=black,miter limit=4.00,line width=0.212pt]
        (-63.0225,70.7755) circle (0.0075cm);
    \end{scope}
    \begin{scope}[cm={{-3.77953,0.0,0.0,3.77953,(728.88602,96.66401)}}]
      \path[draw=black,line join=miter,line cap=butt,miter limit=4.00,line
        width=0.212pt] (71.8546,70.8087) -- (65.0572,70.8087);
      \path[scale=-1.000,draw=black,fill=cffffff,line join=miter,line cap=butt,miter
        limit=4.00,even odd rule,line width=0.212pt,rounded corners=0.0000cm]
        (-70.4742,-72.7854) rectangle (-66.4995,-68.9142);
      \path[draw=black,line join=miter,line cap=butt,miter limit=4.00,line
        width=0.212pt] (68.4051,69.0001) -- (68.4051,65.8687);
      \path[draw=black,line join=miter,line cap=butt,miter limit=4.00,line
        width=0.212pt] (83.0290,70.8064) -- (74.7914,70.8064);
      \path[xscale=-1.000,yscale=1.000,fill=black,miter limit=4.00,line width=0.212pt]
        (-73.8823,70.8048) circle (0.0075cm);
      \path[xscale=-1.000,yscale=1.000,fill=black,miter limit=4.00,line width=0.212pt]
        (-72.7156,70.8048) circle (0.0075cm);
      \path[scale=-1.000,draw=black,fill=cffffff,line join=miter,line cap=butt,miter
        limit=4.00,even odd rule,line width=0.212pt,rounded corners=0.0000cm]
        (-80.0694,-72.8456) rectangle (-76.0947,-68.9744);
      \path[draw=black,line join=miter,line cap=butt,miter limit=4.00,line
        width=0.212pt] (78.0003,69.0603) -- (78.0003,65.9289);
      \path[draw=black,line join=miter,line cap=butt,miter limit=4.00,line
        width=0.212pt] (62.1198,70.8044) -- (57.5524,70.8044);
      \path[scale=-1.000,draw=black,fill=cffffff,line join=miter,line cap=butt,miter
        limit=4.00,even odd rule,line width=0.212pt,rounded corners=0.0000cm]
        (-60.7393,-72.7811) rectangle (-56.7646,-68.9099);
      \path[draw=black,line join=miter,line cap=butt,miter limit=4.00,line
        width=0.212pt] (58.7120,68.9708) -- (58.7120,65.8394);
      \path[xscale=-1.000,yscale=1.000,fill=black,miter limit=4.00,line width=0.212pt]
        (-64.1892,70.7755) circle (0.0075cm);
      \path[xscale=-1.000,yscale=1.000,fill=black,miter limit=4.00,line width=0.212pt]
        (-63.0225,70.7755) circle (0.0075cm);
    \end{scope}
    \path[draw=black,line join=bevel,line cap=butt,miter limit=4.00,line
      width=0.801pt] (327.7993,342.0915) .. controls (327.7993,327.6698) and
      (367.0690,341.9653) .. (371.4884,328.2019) .. controls (375.0239,341.9653) and
      (414.0411,327.8093) .. (414.0411,342.2178);
    \path[draw=black,line join=bevel,line cap=butt,miter limit=4.00,line
      width=0.801pt] (426.8574,342.1681) .. controls (426.8574,327.7463) and
      (466.1271,342.0418) .. (470.5465,328.2785) .. controls (474.0820,342.0418) and
      (513.0992,327.8859) .. (513.0992,342.2944);
    \path[fill=black,line width=3.024pt] (365.8382,323.5863) node[above right]
      (text1880) {${\scriptstyle N}$};
    \path[fill=black,line width=3.024pt] (464.8727,323.9538) node[above right]
      (text1880-6) {${\scriptstyle N}$};
\end{tikzpicture}
    .
\end{align}
If the rank $r_{\text{row}}$ of matrix $M$ is large, the rank $r_{\text{row}}$ of the corresponding tensor-network representation of the unfolded tensor will also be large. While above unfolding requires a particular matrix dimension, it is in fact possible to write any $N\times N$ matrix $M$ as a submatrix of a $2^N\times2^N$ matrix, to which the above unfolding strategy can then be used as a tool for leveraging matrix rank separations \cite{COHEN1993149,Goucha2017,Fawzi2015,Gouveia:2013} into tensor rank separations \cite{Gemma2019}.

Finally, in order to discuss the significance of these results, note firstly that the TT-rank$_\mathbb{R}$ can be arbitrarily smaller than all other ranks, however, optimizing a real MPS to represent a probability distribution presents a problem since it is not clear how to impose positivity of the contracted tensor network \cite{Kliesch2014,werner2016positive}. All other separations are relevant in practice since, as discussed in the following section, they apply to tensor networks that can be trained to represent probability distributions over many variables. Taken together, these results then show that LPS should be preferred over MPS$_{\mathbb{R}_{\geq 0}}$ or BM, since the puri-ranks will always be lower bounded compared to the other ranks. Additionally, complex BM should also be preferred to real BM as they can lead to an arbitrarily large reduction in the number of parameters of the tensor network. Note that because of the structure of the tensor networks we consider, these results also apply to more general tensor factorizations relying on a tree structure of the tensor network. How these results are affected if one considers approximate as opposed to exact representations remains an interesting open problem.

\section{Learning algorithms}
\label{sec:algo}

While the primary results of this work concern the expressive power of different tensor-network representations of probability distributions, these results are relevant in practice since MPS$_{\mathbb{R}_{\geq 0}}$, BM$_{\mathbb{R}/\mathbb{C}}$ and LPS$_{\mathbb{R}/\mathbb{C}}$ admit efficient learning algorithms, as shown in this section. 

First, given samples $\{\mathbf{x_i} = (X_1^i,\ldots,X_N^i)\}$ from a discrete multivariate distribution, they can be trained to approximate this distribution through maximum likelihood estimation. Specifically, this can be done by minimizing the negative log-likelihood,
\begin{align}
L=-\sum_i \log \frac{T_{\mathbf{x_i}}}{Z_T},
\end{align}
where $i$ indexes training samples and $T_{\mathbf{x_i}}$ is given by the contraction of one of the tensor-network models we have introduced. The derivative of the log-likelihood with respect to a parameter $w$ in the tensor network is given by
\begin{align}
\partial_w L=-\sum_i \frac{\partial_w T_\mathbf{x_i}}{T_\mathbf{x_i}}-\frac{\partial_w Z_T}{Z_T}.
\end{align}
The negative log-likelihood can be minimized using a mini-batch gradient-descent algorithm. At each step of the optimization, the sum is computed over a batch of training instances. The parameters in the tensor network are then updated by a small step in the inverse direction of the gradient. Note that when using complex tensors, the derivatives are replaced by Wirtinger derivatives with respect to the conjugated tensor elements. This algorithm requires the computation of $T_{\mathbf{x_i}}$ and $\partial_w T_{\mathbf{x_i}}$ for a training instance, as well as of $Z_T$ and $\partial_w Z_T$. 

We first focus on the computation of these quantities for LPS. Since Born machines are LPS of purification dimension $\mu=1$, they can directly use the same algorithm \cite{Han2017}. For an LPS$_\mathbb{C}$ of puri-rank $r$, the normalization $Z_T$ can be computed by contracting the 
tensor network
\begin{align}
\label{eq:normalization}
Z_T=\sum_{X_1,\ldots,X_N} T_{X_1,\ldots, X_N} = \begin{tikzpicture}[y=0.9pt, x=0.9pt,yscale=-1.0, xscale=1.0,  baseline=(current bounding box.center)]
\path[draw=black,line join=miter,line cap=butt,line width=0.800pt] (357.2487,123.8525) -- (368.6130,123.8525) -- (368.6130,191.1539) -- (357.1225,191.1539);\path[draw=black,line join=miter,line cap=butt,line width=0.800pt] (380.5454,124.0419) -- (391.9096,124.0419) -- (391.9096,191.3433) -- (380.4191,191.3433);\path[draw=black,line join=miter,line cap=butt,line width=0.800pt] (428.0226,124.0419) -- (439.3868,124.0419) -- (439.3868,191.3433) -- (427.8963,191.3433);\path[draw=black,line join=miter,line cap=butt,line width=0.800pt] (451.5086,124.0419) -- (462.8728,124.0419) -- (462.8728,191.3433) -- (451.3823,191.3433);\path[fill=cffffff,line join=miter,line cap=butt,even odd rule,line width=0.800pt] (438.1440,144.6531) -- (440.7110,144.6531) -- (440.7110,146.7513) -- (438.1217,146.7513);\path[fill=cffffff,line join=miter,line cap=butt,even odd rule,line width=0.800pt] (438.0178,167.7603) -- (440.5847,167.7603) -- (440.5847,169.8585) -- (437.9954,169.8585);\path[fill=cffffff,line join=miter,line cap=butt,even odd rule,line width=0.800pt] (367.2991,167.7206) -- (369.8661,167.7206) -- (369.8661,169.8188) -- (367.2768,169.8188);\path[fill=cffffff,line join=miter,line cap=butt,even odd rule,line width=0.800pt] (390.6027,144.6849) -- (393.1696,144.6849) -- (393.1696,146.7831) -- (390.5804,146.7831);\path[fill=cffffff,line join=miter,line cap=butt,even odd rule,line width=0.800pt] (390.6027,167.8099) -- (393.1696,167.8099) -- (393.1696,169.9081) -- (390.5804,169.9081);\begin{scope}[shift={(32.10578,-113.85825)},miter limit=4.00,line width=0.800pt]
\path[draw=black,line join=miter,line cap=butt,miter limit=4.00,line width=0.800pt] (325.3029,282.7447) -- (365.6412,282.7447);\path[cm={{0.04091,0.0,0.0,-0.04091,(365.3516,291.59635)}},fill=black,miter limit=4.00,line width=19.553pt] (137.3807,216.2084) .. controls (137.3807,229.7373) and (126.4134,240.7046) .. (112.8845,240.7046) .. controls (99.3557,240.7046) and (88.3883,229.7373) .. (88.3883,216.2084) .. controls (88.3883,202.6795) and (99.3557,191.7122) .. (112.8845,191.7122) .. controls (126.4134,191.7122) and (137.3807,202.6795) .. (137.3807,216.2084) -- cycle;\path[cm={{0.04091,0.0,0.0,-0.04091,(371.01116,291.59635)}},fill=black,miter limit=4.00,line width=19.553pt] (137.3807,216.2084) .. controls (137.3807,229.7373) and (126.4134,240.7046) .. (112.8845,240.7046) .. controls (99.3557,240.7046) and (88.3883,229.7373) .. (88.3883,216.2084) .. controls (88.3883,202.6795) and (99.3557,191.7122) .. (112.8845,191.7122) .. controls (126.4134,191.7122) and (137.3807,202.6795) .. (137.3807,216.2084) -- cycle;\path[draw=black,line join=miter,line cap=butt,miter limit=4.00,line width=0.800pt] (379.8849,282.7522) -- (420.2232,282.7522);\end{scope}
\path[draw=black,line join=miter,line cap=butt,line width=0.800pt] (452.2126,145.8686) -- (452.2126,170.0650);\path[draw=black,line join=miter,line cap=butt,line width=0.800pt] (428.3733,146.3150) -- (428.3733,170.5114);\path[draw=black,line join=miter,line cap=butt,line width=0.800pt] (380.6947,145.7793) -- (380.6947,169.9757);\path[draw=black,line join=miter,line cap=butt,line width=0.800pt] (357.1233,146.3597) -- (357.1233,170.5561);\path[draw=black,line join=miter,line cap=butt,miter limit=4.00,line width=0.800pt] (357.4036,175.7985) -- (357.4036,191.1692);\path[draw=black,line join=miter,line cap=butt,miter limit=4.00,line width=0.800pt] (380.8867,175.7985) -- (380.8867,191.1692);\path[draw=black,line join=miter,line cap=butt,miter limit=4.00,line width=0.800pt] (428.3698,175.7985) -- (428.3698,191.1692);\path[draw=black,line join=miter,line cap=butt,miter limit=4.00,line width=0.800pt] (451.8529,175.7985) -- (451.8529,191.1692);\path[draw=black,fill=cffffff,line join=miter,line cap=butt,miter limit=4.00,even odd rule,line width=0.800pt,rounded corners=0.0000cm] (350.2547,161.5852) rectangle (365.2772,176.2163);\path[draw=black,fill=cffffff,line join=miter,line cap=butt,miter limit=4.00,even odd rule,line width=0.800pt,rounded corners=0.0000cm] (373.1231,161.5852) rectangle (388.1457,176.2163);\path[draw=black,fill=cffffff,line join=miter,line cap=butt,miter limit=4.00,even odd rule,line width=0.800pt,rounded corners=0.0000cm] (420.8529,161.5852) rectangle (435.8754,176.2163);\path[draw=black,fill=cffffff,line join=miter,line cap=butt,miter limit=4.00,even odd rule,line width=0.800pt,rounded corners=0.0000cm] (444.8440,161.5852) rectangle (459.8665,176.2163);\path[draw=black,line join=miter,line cap=butt,miter limit=4.00,line width=0.800pt] (357.2083,139.2773) -- (357.2083,123.9066);\path[draw=black,line join=miter,line cap=butt,miter limit=4.00,line width=0.800pt] (380.6914,139.2773) -- (380.6914,123.9066);\path[draw=black,line join=miter,line cap=butt,miter limit=4.00,line width=0.800pt] (428.1745,139.2773) -- (428.1745,123.9066);\path[draw=black,line join=miter,line cap=butt,miter limit=4.00,line width=0.800pt] (451.6576,139.2773) -- (451.6576,123.9066);\path[fill=cffffff,line join=miter,line cap=butt,even odd rule,line width=0.800pt] (367.4107,144.7518) -- (369.9777,144.7518) -- (369.9777,146.8500) -- (367.3884,146.8500);\begin{scope}[shift={(32.10578,-136.96716)},miter limit=4.00,line width=0.800pt]
\path[draw=black,line join=miter,line cap=butt,miter limit=4.00,line width=0.800pt] (325.3029,282.7447) -- (365.6412,282.7447);\path[cm={{0.04091,0.0,0.0,-0.04091,(365.3516,291.59635)}},fill=black,miter limit=4.00,line width=19.553pt] (137.3807,216.2084) .. controls (137.3807,229.7373) and (126.4134,240.7046) .. (112.8845,240.7046) .. controls (99.3557,240.7046) and (88.3883,229.7373) .. (88.3883,216.2084) .. controls (88.3883,202.6795) and (99.3557,191.7122) .. (112.8845,191.7122) .. controls (126.4134,191.7122) and (137.3807,202.6795) .. (137.3807,216.2084) -- cycle;\path[cm={{0.04091,0.0,0.0,-0.04091,(371.01116,291.59635)}},fill=black,miter limit=4.00,line width=19.553pt] (137.3807,216.2084) .. controls (137.3807,229.7373) and (126.4134,240.7046) .. (112.8845,240.7046) .. controls (99.3557,240.7046) and (88.3883,229.7373) .. (88.3883,216.2084) .. controls (88.3883,202.6795) and (99.3557,191.7122) .. (112.8845,191.7122) .. controls (126.4134,191.7122) and (137.3807,202.6795) .. (137.3807,216.2084) -- cycle;\path[draw=black,line join=miter,line cap=butt,miter limit=4.00,line width=0.800pt] (379.8849,282.7522) -- (420.2232,282.7522);\end{scope}
\path[xscale=1.000,yscale=-1.000,draw=black,fill=cffffff,line join=miter,line cap=butt,miter limit=4.00,even odd rule,line width=0.800pt,rounded corners=0.0000cm] (350.0594,-153.4906) rectangle (365.0819,-138.8595);\path[xscale=1.000,yscale=-1.000,draw=black,fill=cffffff,line join=miter,line cap=butt,miter limit=4.00,even odd rule,line width=0.800pt,rounded corners=0.0000cm] (372.9279,-153.4906) rectangle (387.9504,-138.8595);\path[xscale=1.000,yscale=-1.000,draw=black,fill=cffffff,line join=miter,line cap=butt,miter limit=4.00,even odd rule,line width=0.800pt,rounded corners=0.0000cm] (420.6576,-153.4906) rectangle (435.6801,-138.8595);\path[xscale=1.000,yscale=-1.000,draw=black,fill=cffffff,line join=miter,line cap=butt,miter limit=4.00,even odd rule,line width=0.800pt,rounded corners=0.0000cm] (444.6487,-153.4906) rectangle (459.6713,-138.8595);
\node[] at (357.5,147) {$A_1$};
\node[] at (357.5,168.5) {$\xoverline[0.75]{A_1}$};
\end{tikzpicture}.
\end{align}
This contraction is performed in $\mathcal{O}(d\mu r^3 N)$ operations from left to right by contracting at each step the two vertical indices and then each of the two horizontal indices. During this contraction, intermediate results from the contraction of the first $i$ tensors are stored in $E_i$, and the same procedure is repeated from the right with intermediate results of the contraction of the last $N-i$ tensors stored in $F_{i+1}$. The derivatives of the normalization for each tensor are then computed as
\begin{align}
\frac{\partial Z_T}{\partial \bar{A}_{i,m}^{j,k,l}}=\begin{tikzpicture}[y=0.95pt, x=0.95pt,yscale=-1.0, xscale=1.0,  baseline=(current bounding box.center)]
\path[draw=black,line join=miter,line cap=butt,line width=0.800pt] (485.4637,207.4285) -- (496.8279,207.4285) -- (496.8279,274.7299) -- (485.3374,274.7299);\path[fill=cffffff,line join=miter,line cap=butt,even odd rule,line width=0.800pt] (495.5333,228.3063) -- (498.1003,228.3063) -- (498.1003,230.4045) -- (495.5110,230.4045);\path[fill=cffffff,line join=miter,line cap=butt,even odd rule,line width=0.800pt] (495.4702,251.3504) -- (498.0372,251.3504) -- (498.0372,253.4486) -- (495.4479,253.4486);\path[draw=black,line join=miter,line cap=butt,miter limit=4.00,line width=0.800pt] (461.7423,252.4264) -- (502.0806,252.4264);\path[draw=black,line join=miter,line cap=butt,miter limit=4.00,line width=0.800pt] (461.7423,229.3175) -- (502.0806,229.3175);\path[xscale=1.000,yscale=-1.000,draw=black,fill=cffffff,line join=miter,line cap=butt,miter limit=4.00,even odd rule,line width=0.800pt,rounded corners=0.0000cm] (447.9644,-259.8908) rectangle (469.4155,-222.9382);\path[xscale=1.000,yscale=-1.000,draw=black,fill=cffffff,line join=miter,line cap=butt,miter limit=4.00,even odd rule,line width=0.800pt,rounded corners=0.0000cm] (502.3347,-259.1765) rectangle (523.4286,-222.2239);\path[draw=black,line join=miter,line cap=butt,line width=0.800pt] (485.6130,229.1659) -- (485.6130,253.3623);\path[draw=black,line join=miter,line cap=butt,miter limit=4.00,line width=0.800pt] (485.8050,259.1851) -- (485.8050,274.5558);\path[fill=cffffff,even odd rule,rounded corners=0.0000cm] (478.7557,245.1503) rectangle (493.7783,259.7814);\path[draw=black,line join=miter,line cap=butt,miter limit=4.00,line width=0.800pt] (485.6097,222.6639) -- (485.6097,207.2932);\path[xscale=1.000,yscale=-1.000,draw=black,fill=cffffff,line join=miter,line cap=butt,miter limit=4.00,even odd rule,line width=0.800pt,rounded corners=0.0000cm] (478.5605,-237.0558) rectangle (493.5830,-222.4246);
\node[] at (485,230) {$A_i$};
\node[] at (485.5,248.5) {${\scriptscriptstyle j}$};
\node[] at (481.5,253) {${\scriptscriptstyle k}$};
\node[] at (490.5,253) {${\scriptscriptstyle l}$};
\node[] at (485.5,257.5) {${\scriptscriptstyle m}$};
\node[] at (459,240) {$E_{i-1}$};
\node[] at (513,240) {$F_{i+1}$};
\end{tikzpicture},
\end{align}
which also costs $\mathcal{O}(d\mu r^3 N)$ operations.
Computing $T_{\mathbf{x_i}}$ for a training example and its derivative is done in the same way, except that the contracted index corresponding to an observed variable is now fixed to its observed value.

Note that here the training is done by computing the gradients of the log-likelihood over all tensors for each batch of training example and then updating all tensors at once in a gradient-descent optimization scheme. A different approach would be a DMRG-like algorithm where only a few tensors are updated at a time. The computation of $Z_T$ and its derivative may be greatly simplified by using canonical forms \cite{SchollwockReview}.

The algorithm we use for training MPS$_{\mathbb{R}_{\geq 0}}$ is a variation of the one given above for LPS and is detailed in the supplementary material. MPS$_{\mathbb{R}_{\geq 0}}$ could also be trained using the expectation-maximization (EM) algorithm, but as BM and LPS use real or complex tensors, different algorithms are required. In Section~\ref{sec:numerical} we will compare the algorithms described here with the EM algorithm for HMM. Note that in all these models not only the likelihood can be evaluated efficiently: marginals and correlation functions can be computed in a time linear in the number of variables, while exact samples from the distribution can also be generated efficiently \cite{PhysRevB.85.165146,Han2017}.

Instead of approximating a distribution from samples, it might also be useful to compress a probability mass function $P$ given in the form of a non-negative tensor. Since the original probability mass function has a number of parameters that is exponential in $N$, this is only possible for a small number of variables. It can be done by minimizing the Kullback–Leibler (KL) divergence 
\begin{align}
D(P||T/Z_T)=\sum_{X_1,\ldots,X_N}P_{X_1,\ldots,X_N}\log\left(\frac{P_{X_1,\ldots,X_N}}{T_{X_1,\ldots,X_N}/Z_T}\right),
\end{align}
where $T$ is represented by a tensor-network model. The gradient of the KL-divergence can be obtained in the same way as the gradient of the log-likelihood and gradient-based optimization algorithms can then be used to solve this optimization problem. Note that for the case of matrices and MPS$_{\mathbb{R}_\geq 0}$ more specific algorithms have been developed \cite{AlgoNNMF}, and finding more efficient algorithms for factorizing a given tensor in the form of a BM or LPS represents an interesting problem that we leave for future work.

\section{Numerical experiments}
\label{sec:numerical}

Using the algorithms discussed in Section~\ref{sec:algo} we numerically investigate the extent to which the separations found in Section~\ref{sec:expressivity2} apply in both the setting of approximating a distribution from samples, and the setting of compressing given non-negative tensors. Code, data sets and choice of hyperparameters are available in the supplementary material and the provided repository \cite{githubrepo}.

\subsection{Random tensor factorizations}

We first generate random probability mass functions $P$ by generating a tensor with elements chosen uniformly in $[0,1]$ and normalizing it. We then minimize the KL-divergence $D(P||T/Z_T)$, where $T$ is the tensor defined by an MPS, BM or LPS with given rank $r$. We choose LPS to have a purification dimension of $2$. Details of the optimization are available in the supplementary material.

\begin{figure}[ht]
\centering
\includegraphics[width = 0.92\linewidth]{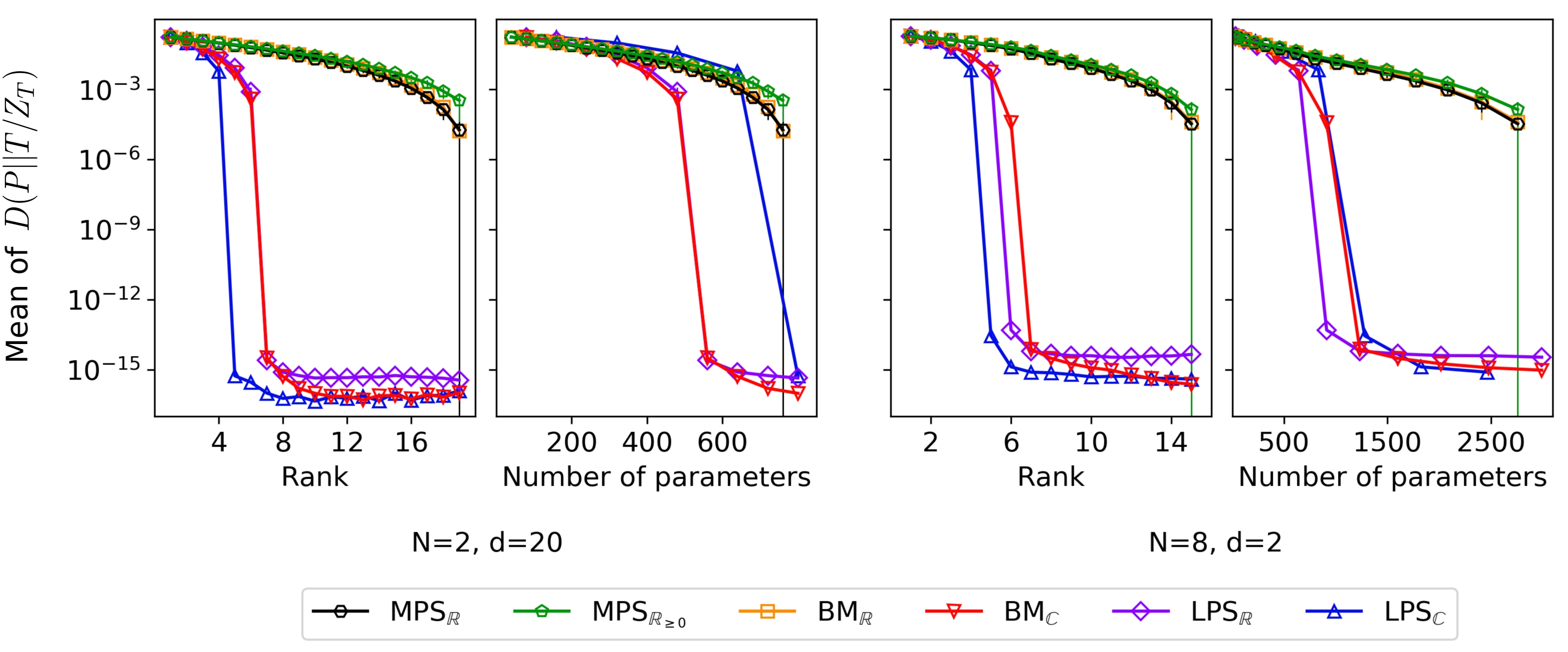}
\caption{Mean of the minimum error of the approximation of 50 random tensors $P$ with tensor networks of fixed rank, as a function of the rank or the number of (real) parameters. Left: $20\times 20$ matrix. Right: tensor over 8 binary variables. The errors bars represent one standard deviation, and are omitted below $10^{-12}$.}
\label{fig:randomtensors}
\end{figure}

Results are presented in Fig.~\ref{fig:randomtensors} for a $20\times 20$ matrix and a tensor with 8 binary variables. They show that complex BM as well as real and complex LPS generically provide a better approximation to a tensor than an MPS or real BM, for fixed rank as well as for fixed number of real parameters.

\subsection{Maximum likelihood estimation on realistic data sets}

We now investigate how well the different tensor-network representations are able to learn from realistic data sets. We train MPS$_{\mathbb{R}_{\geq 0}}$, BM$_{\mathbb{R}}$, BM$_{\mathbb{C}}$, LPS$_{\mathbb{R}}$ and LPS$_{\mathbb{C}}$ (of purification dimension $2$) using the algorithm of Section~\ref{sec:algo} on different data sets of categorical variables. Since we are interested in the expressive power of the different representations we use the complete data sets and no regularization. Additional results on generalization performance are included in the supplementary material.

\begin{figure}[ht]
\centering
\includegraphics[width = 0.94\linewidth]{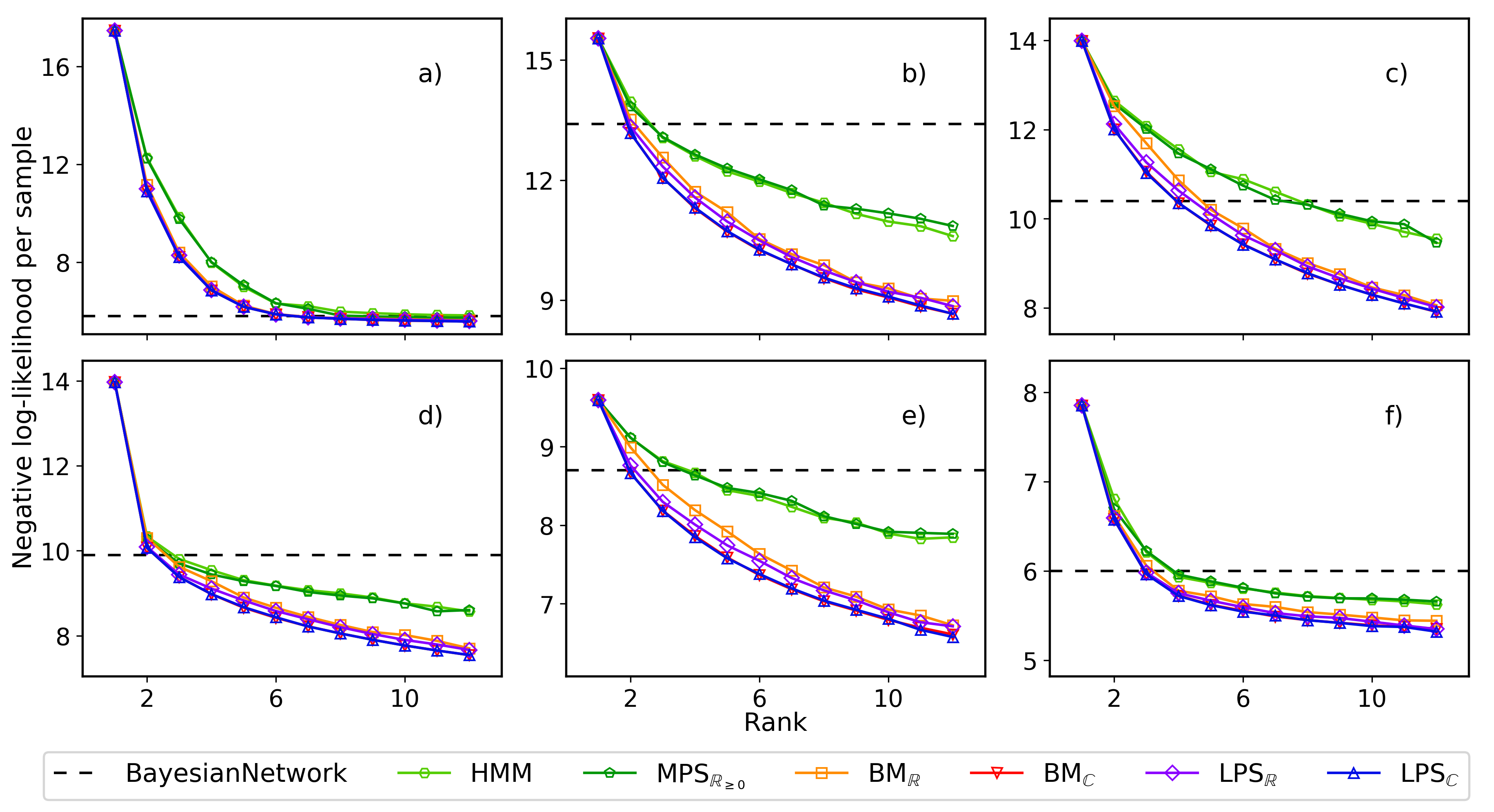}
\caption{Maximum likelihood estimation with tensor networks, HMM and a Bayesian network without hidden units with graph learned from the data on different data sets: a) biofam data set of family life states from the Swiss Household Panel biographical survey \cite{biofam}; data sets from the UCI Machine Learning Repository \cite{Dua:2019}: b) Lymphography \cite{lymphodataset}, c) SPECT Heart, d) Congressional Voting Records, e) Primary Tumor \cite{lymphodataset}, f) Solar Flare.}
\label{fig:datasets}
\end{figure}

The results in Fig.~\ref{fig:datasets} show the best negative log-likelihood per sample obtained for each tensor network of fixed rank. As a comparison we also include the best negative log-likelihood obtained from an HMM trained using the Baum-Welch algorithm \cite{baum1970,DBLP:journals/jmlr/Schreiber17}, as well as the best possible Bayesian network without hidden variables, where the network graph is learned from the data \cite{DBLP:journals/jmlr/Schreiber17}. We observe that despite the different algorithm choice, the performance of HMM and MPS$_{\mathbb{R}_{\geq 0}}$ are similar, as we could expect from their relationship. On all data sets, BM and LPS lead to significant improvements for the same rank over MPS$_{\mathbb{R}_{\geq 0}}$.

\section{Conclusion}
\label{sec:conclusion}

We have characterized the expressive power of various tensor-network models of probability distributions, in the process enhancing the scope and applicability of the tensor-network toolbox within the broader context of learning algorithms. In particular, our analysis has concrete implications for model selection, suggesting that in generic settings LPS should be preferred over both hidden Markov models and Born machines. Furthermore, our results prove that unexpectedly the use of complex tensors over real tensors can lead to an unbounded expressive advantage in particular network architectures. 
Additionally, this work contributes to the growing body of rigorous results concerning the expressive power of learning models, which have been obtained via tensor-network techniques. A formal understanding of the expressive power of state-of-the-art learning models is often elusive; it is hoped that both the techniques and spirit of this work can be used to add momentum to this program. Finally, through the formal relationship of LPS and Born machines to quantum circuits, our work provides a concrete foundation for both the development and analysis of quantum machine learning algorithms for near-term quantum devices.

\subsubsection*{Acknowledgments}

We would like to thank Vedran Dunjko for his comments on the manuscript and Jo\~{a}o Gouveia for his suggestion of the proof of Lemma 9 in the supplementary material. 
I.~G., N.~P.\ and J.~I.~C.\ are supported by an ERC Advanced Grant QENOCOBA under the EU Horizon 2020 program (grant agreement 742102) and the German Research Foundation (DFG) under Germany's Excellence Strategy through Project No.\ EXC-2111 - 390814868 (MCQST). R.~S.\ acknowledges the financial support of the Alexander von Humboldt foundation. 
N.~P.\ acknowledges financial support from ExQM.
J.~E.\ acknowledges financial support by the German Research Foundation DFG (CRC 183 project B2, EI 519/7-1, CRC 1114, GRK 2433) and MATH+. This work has also received funding from the European Union's Horizon 2020 research and innovation programme under grant agreement No 817482 (PASQuanS).

\small

\bibliographystyle{unsrt}
\bibliography{biblio}

\end{document}


\maketitle

\small

\setcounter{equation}{0}
\setcounter{figure}{0}
\setcounter{table}{0}
\setcounter{section}{0}
\setcounter{page}{1}
\setcounter{proposition}{0}
\setcounter{theorem}{0}
\renewcommand{\theequation}{S\arabic{equation}}
\renewcommand{\thefigure}{S\arabic{figure}}

\tableofcontents

\section{Relationship between tensor networks, hidden Markov models and quantum circuits}

\subsection{Non-negative MPS are HMM}

\begin{figure}[ht]
\centering
\subfloat[]{\includegraphics[width=0.27\linewidth]{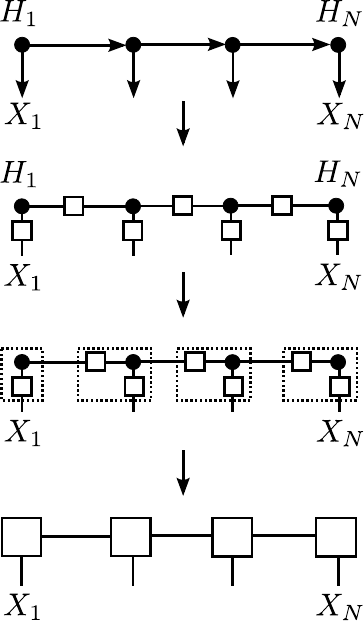}}\hspace{0.05\linewidth}
\subfloat[]{\includegraphics[width=0.28\linewidth]{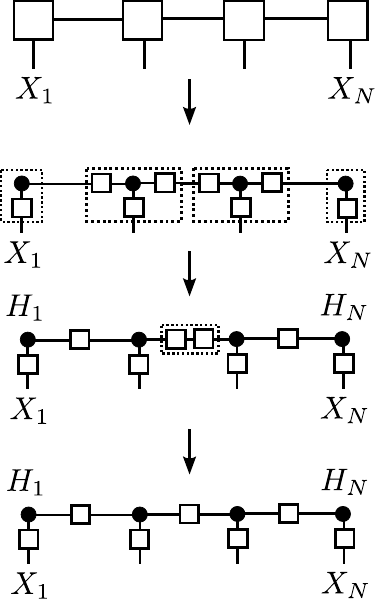}\label{fig:MPSHMMsup}}
\caption{\small (a) Mapping of a hidden Markov chain to a MPS with non-negative tensor elements. (b) Mapping of a MPS with non-negative tensor elements to a hidden Markov chain.}
\end{figure}

Consider an MPS with non-negative tensors $A_i$ and TT-rank$_{\mathbb{R}_{\geq 0}}=r$. To express the corresponding probability distribution as a HMM, we split the tensors using an (exact) non-negative canonical polyadic decomposition such that $A_{i,l}^{jk}=\sum_{s=1}^{r'} B_i^{js} C_i^{ls} D_i^{sk}$, where $r'\leq \min(dr,r^2)$ (Fig.~\ref{fig:MPSHMMsup}). We can now set
\begin{align}
P(X_i=l|H_i=s)&=C_i^{ls}\\
P(H_i=s|H_{i-1}=j)&=\sum_u D_{i-1}^{ju} B_i^{us},
\end{align}
where the probabilities must be normalized properly, which can be done by first constructing the unnormalized factor graph and then normalizing the probabilities on every edge. We have then defined a HMM with hidden variables of dimension $r'$ that defines the same probability of the observed units as the one arising from the MPS. Note that using a different graph for the hidden Markov model we could also arrive at a dimension of hidden variables of $r$ \cite{Robeva2017,Glasser2018}.

\subsection{Local quantum circuits are Born machines}

In order to clarify the relationship between Born machines and local quantum circuits we provide here a concise introduction to the formalism of circuit based quantum computing. For a more thorough description, see ref.~\cite{Nielsen:2011:QCQ:1972505}.

Consider the Hilbert space $\mathcal{H} = \mathbb{C}^d$, with ortho-normal basis $\{|X\rangle\}_{X = 1}^d$, where Dirac notation $|X\rangle$ has been used to represent a vector $\vec{X} \in \mathcal{H}$. From a mathematical perspective, a $d$-dimensional \emph{qudit} is a system whose state vector $|\psi\rangle$ can be described by a unit vector in $\mathcal{H}$. With respect to any fixed ortho-normal basis $\{|X\rangle\}_{X = 1}^d$, a qudit is therefore specified by $d$ complex amplitudes $\{\psi(X)\}_{X = 1}^d$ - i.e., $|\psi\rangle = \sum_{X = 1}^d \psi(X)|X\rangle$, with $\sum_{X = 1}^d|\psi(X)|^2 = 1$.

At a high level, a quantum circuit then consists of multiple qudits, and a sequence of quantum gates, which are unitary operations acting on (a subset of) the qudits, with unitarity required to preserve the normalization of the global state of the system. To be more precise, consider a collection of $N$ $d$-dimensional qudits, each described by unit vectors in $\mathcal{H}_i = \mathbb{C}^d$, where $i \in \{1,\ldots,N\}$ indicates a particular qudit. Given such a collection of qudits, 
the global state of the system is described by a unit vector $|\psi\rangle \in \mathcal{H} \equiv \bigotimes_{i = 1}^N\mathcal{H}_i = \mathbb{C}^{d^N}$. In particular, with respect to a fixed ortho-normal basis $\{|X_i\rangle\}_{X = 1}^d$ for each sub-system Hilbert space $\mathcal{H}_i$, the global state is specified by $d^N$ complex amplitudes $\psi(X_1,\ldots,X_N)$ - i.e.,
\begin{equation}
    |\psi\rangle = \sum_{X_1=1}^d\ldots\sum_{X_N=1}^d \psi(X_1,\ldots,X_N)|X_1\rangle\otimes \ldots \otimes|X_N\rangle,
\end{equation}
with the constraint that
\begin{equation}
    \langle \psi|\psi\rangle = \sum_{X_1=1}^d\ldots\sum_{X_N=1}^d |\psi(X_1,\ldots,X_N)|^2 = 1.
\end{equation}
Note that the set of all amplitudes, which completely defines the state vector $|\psi\rangle$ with respect to this particular basis, is naturally represented as an order-$N$ tensor with $d$-dimensional indices:
\begin{equation}
    \psi(X_1,\ldots,X_N) = \begin{tikzpicture}[y=0.80pt, x=0.80pt, yscale=-1.200000, xscale=1.200000, inner sep=0pt, outer sep=0pt, baseline={([yshift=-12pt]current bounding box.center)}]
\path[draw=black,line join=miter,line cap=butt,miter limit=4.00,line
  width=0.800pt] (288.1262,456.0122) -- (288.1262,444.4994);
\path[draw=black,line join=miter,line cap=butt,miter limit=4.00,line
  width=0.800pt] (309.1569,456.0122) -- (309.1569,444.4994);
\path[draw=black,line join=miter,line cap=butt,miter limit=4.00,line
  width=0.800pt] (267.0955,456.0122) -- (267.0955,444.4994);
\path[draw=black,line join=miter,line cap=butt,miter limit=4.00,line
  width=0.800pt] (330.1876,456.0122) -- (330.1876,444.4994);
\path[scale=-1.000,draw=black,fill=cffffff,line join=miter,line cap=butt,miter
  limit=4.00,even odd rule,line width=0.800pt,rounded corners=0.0000cm]
  (-337.7193,-471.1178) rectangle (-260.4949,-456.1295);
\node[align=center] at (268,440) {$X_1$};
\node[align=center] at (330.5,440) {$X_N$};
\end{tikzpicture}.
\end{equation}
Furthermore, in this diagrammatic notation where legs that join two tensors represent a summation over the corresponding indices of the tensors, the normalization constraint takes the particularly simple form,
\begin{equation}
\langle \psi|\psi\rangle =
    \vcenter{\hbox{\includegraphics[width=0.16\linewidth]{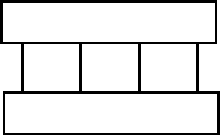}}}=1,
\end{equation}
where the upper tensor is taken to be the complex conjugate of the lower tensor.

Given some initial state of the system, a quantum circuit then consists of a sequence of unitary operations (referred to as \emph{gates}), each of which preserves the normalization of the global state of the system. We are particularly interested in \emph{local} quantum circuits, which consist of unitary operations which act only on a subset of qubits. To be more precise, given some Hilbert space $\mathcal{H}$, let us denote the set of unitary operators acting on elements of $\mathcal{H}$ as $\mathcal{U}(\mathcal{H})$. We will be concerned with $2$-local quantum circuits, which consist of unitary operations acting only on pairs of neighbouring qudits - i.e., all gates $U \in \mathcal{U}(\mathcal{H})$are of the form 
\begin{equation}
    U = \mathds{1_1}\otimes\ldots\otimes\mathds{1}_{i-1}\otimes U_{i, i+1} \otimes \mathds{1}_{i + 2} \otimes\ldots\otimes \mathds{1}_N,
\end{equation}
for some $i \in \{1,\ldots,N\}$, where $U_{j, j+1} \in \mathcal{U}(\mathcal{H}_j \otimes \mathcal{H}_{j+1})$ and $\mathds{1}_k$ is the identity operator on $\mathcal{H}_k$.

Let us now consider a quantum circuit in a one-dimensional geometry, consisting of $N$ $d$-dimensional qudits, all initialized in the $|0\rangle$ state vector, to which $D$ layers of $2$-local unitary gates are applied. As shown in Equation \eqref{supeq:circuit_to_mps} below one can write the output state of this circuit as an MPS by first splitting each unitary operator through a singular value decomposition, and then contracting all the resulting tensors as indicated by the dashed boxes. Note that as a result of $2$-locality, each unitary operator has rank less than $d^2$, and therefore, the MPS has TT-rank less than $d^{D+1}$.
\begin{align}\label{supeq:circuit_to_mps}
\psi(X_1,\ldots,X_N) = \vcenter{\hbox{\includegraphics[height=0.2\linewidth]{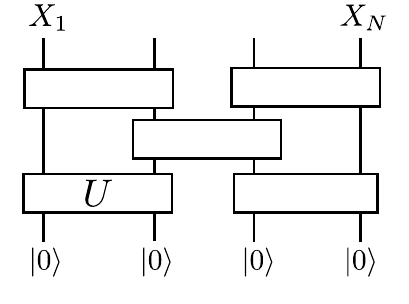}}}
=\vcenter{\hbox{\includegraphics[height=0.2\linewidth]{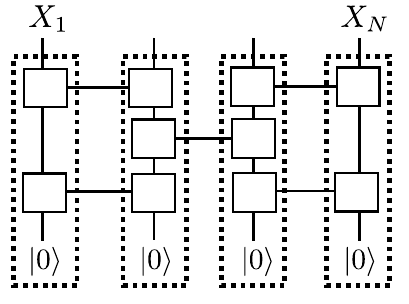}}}
=\raisebox{0.2\height}{$\vcenter{\hbox{\includegraphics[height=0.2\linewidth]{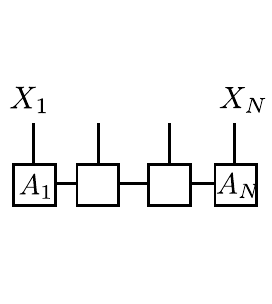}}}$}.
\end{align}

Finally, given the outcome state vector $|\psi\rangle$ of a quantum circuit, it is necessary to understand the measurement process, via which classical information can be extracted from this state. To this end, we need to understand the Born rule of quantum mechanics. More specifically, in the restricted setting of finite-dimensional Hilbert spaces which we consider here, \emph{observables} correspond to Hermitian operators, and the Born rule states that measurement of an observable $O$ will yield one of the eigenvalues $\lambda_i$ of $O$, with probability $\langle \psi |\Pi_i|\psi\rangle$, where $\Pi_i$ is the projection onto the eigenspace of $\lambda_i$. Note that the normalization of $|\psi\rangle$ is required precisely to allow for this probabilistic interpretation of measurements via the Born rule.

Let us now consider an observable $O$ which is diagonal in the fixed basis we have previously considered (often referred to as the ``computational basis''). 
In this case, we can write
\begin{align}
    O &= \sum_{X_1 = 1}^d\ldots\sum_{X_N = 1}^d  \lambda(X_1,\ldots,X_N)|X_1\rangle\langle X_1| \otimes  \ldots\otimes|X_N\rangle\langle X_N|, \\ \label{sup_eqn:observable}
    &= \sum_{X_1 = 1}^d\ldots\sum_{X_N = 1}^d   \lambda(X_1,\ldots,X_N)\Pi(X_1,\ldots,X_N),
\end{align}
and we find that $P(X_1,\ldots,X_N)$, the probability of obtaining measurement outcome $\lambda(X_1,\ldots,X_N)$, is given by
\begin{align}
 P(X_1,\ldots,X_N) &= \langle 
 \psi |\Pi(X_1,\ldots,X_N) |\psi\rangle, \\
 &= |\psi(X_1,\ldots,X_N)|^2,\\
 &= \vcenter{\hbox{\includegraphics[height=0.2\linewidth]{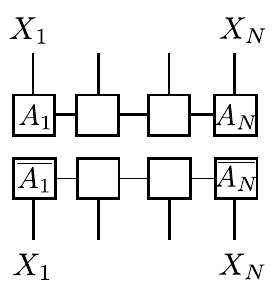}}}.
\end{align}
As such, we find that measurements of the observable $O$ allow us to sample from the probability mass function $P(X_1,\ldots,X_N) = |\psi(X_1,\ldots,X_N)|^2$, and that when $|\psi\rangle$ is the output state of a $2$-local quantum circuit of depth $D$, this probability mass function is exactly a Born machine (where the origin of the name is now clear) of Born-rank $d^{D+1}$. This shows that in this probabilistic modeling approach, local quantum circuits of fixed depth are Born machines of fixed Born-rank.

\subsection{Local quantum circuits with ancillas are locally purified states}

In order to understand the relationship between locally purified states and local quantum circuits with ancillas, it is necessary to understand the effect and formalism of measurements on subsystems.

To this end, consider a one-dimensional array of $2N$ $d$-dimensional qudits, consisting of alternating pairs of \emph{system} and \emph{ancilla} qudits respectively, where each system qudit is a unit vector in $\mathcal{H}_i^{(S)} = \mathbb{C}^d$ spanned by ortho-normal basis $\{|X_i\rangle \}_{X = 1}^d$, and each 
ancilla qudit is a unit vector in $\mathcal{H}^{(A)}_j = \mathbb{C}^{\mu}$ spanned by $\{|Y_j\rangle \}_{Y = 1}^{\mu}$. The global state vector $|\psi\rangle$ is therefore an element of the Hilbert space $\mathcal{H} =\mathcal{H}_1^{(S)}\otimes \mathcal{H}_1^{(A)}\otimes \ldots \otimes \mathcal{H}_{N}^{(S)}\otimes \mathcal{H}_N^{(A)}$ - i.e.,
\begin{align}
    |\psi\rangle &= \sum_{X_1, \ldots, X_N= 1}^d\sum_{Y_1,\ldots,Y_N = 1}^{ \mu}\psi(X_1,Y_1,\ldots,X_{N},Y_{N})|X_1\rangle\otimes|Y_1\rangle \otimes\ldots\otimes|X_{N}\rangle\otimes|Y_N\rangle.
\end{align}
As in the previous section, we can consider a $2$-local quantum circuit of depth $D$, where all qudits (system and ancilla) are 
initialized in the $|0\rangle$ state vector, and note that the output state vector $|\psi\rangle$ can again be written as a matrix product state of TT-rank less than $r^{D+1}$, where $r = \min(d,\mu)$.
\begin{equation}
|\psi \rangle = \vcenter{\hbox{\includegraphics[height=0.2\linewidth]{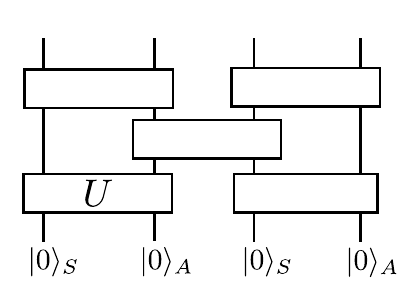}}}
=\vcenter{\hbox{\includegraphics[height=0.2\linewidth]{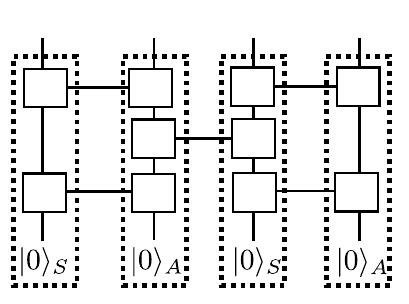}}}
=\raisebox{0.2\height}{$\vcenter{\hbox{\includegraphics[height=0.2\linewidth]{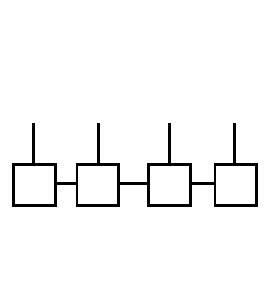}}}$}.
\end{equation}

Now, consider an observable $O$, as in Equation~\eqref{sup_eqn:observable}, which is defined only on the system qudits, and is diagonal in the computational basis for this subsystem - i.e.,
\begin{equation}
    O= \sum_{X_1 = 1}^d\ldots\sum_{X_{N} = 1}^d   \lambda(X_1,\ldots,X_{N})\Pi(X_1\ldots,X_{N}).
\end{equation}

The postulates of quantum mechanics state that the probability $P(X_1,\ldots,X_N)$ of measurement outcome $\lambda(X_1,\ldots,X_{N})$, when performing a measurement of observable $O$ on the system qudits, is given by
\begin{equation}\label{sup_eqn:prob}
    P(X_1,\ldots,X_N) = \mathrm{Tr}\big(\rho_S \Pi(X_1,\ldots,X_N)\big),
\end{equation}
where $\rho_S$ is the system \emph{density matrix}, given by 
\begin{equation}\label{sup_eqn:density_matrix}
    \rho_S = \mathrm{Tr}_{A}(|\psi\rangle\langle \psi|),
\end{equation}
and where $\mathrm{Tr}_{A}$ indicates the partial trace over the Hilbert space of all ancilla qudits. Luckily, equations \eqref{sup_eqn:prob} and \eqref{sup_eqn:density_matrix} are both easily and concisely expressed in tensor network notation (which is in fact a particularly strong motivation for such a notation). Specifically,
\begin{equation}
    \rho_S = \vcenter{\hbox{\includegraphics[height=0.14\linewidth]{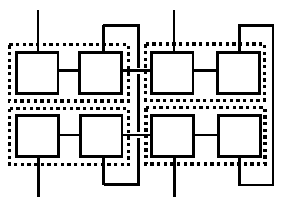}}} = \vcenter{\hbox{\includegraphics[height=0.14\linewidth]{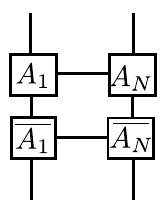}}},
\end{equation}
and therefore
\begin{align}
    P(X_1,\ldots,X_N)&= \sum_{Y_1, 
    \ldots Y_N = 1}^{\mu}|\psi(X_1,Y_1,\ldots,X_N,Y_N)|^2\\
    &=\vcenter{\hbox{\includegraphics[height=0.2\linewidth]{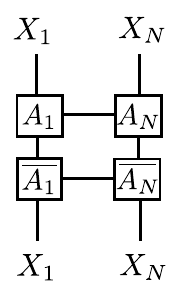}}}. \label{sup_eqn:pmf_lps}
\end{align}
As such, we find that measurements of the observable $O$ on the system qudits allow us to sample from the probability mass function \eqref{sup_eqn:pmf_lps}, which is precisely a locally purified state of puri-rank $r^{D+1}$, where $r = \min(d,\mu)$, if $|\psi\rangle$ is the output state vector of a $2$-local quantum circuit of depth $D$, consisting of $2N$ alternating $d$-dimensional system and $\mu$-dimensional local ancilla qudits.

\section{Proofs on the expressive power of tensor networks}

We provide here proofs for all propositions in Section 5 of the main text. To facilitate ease of presentation and understanding, we restate the propositions here. We begin with the following proposition, concerning inclusions between the sets of probability distributions which can be exactly represented by different tensor-network representations of the 
same rank.

\setcounter{proposition}{0}
\begin{proposition}
\label{sup_prop:inclusions}
For all non-negative tensors 
$\text{TT-rank}_{\mathbb{R}_{\geq 0}} \geq \text{TT-rank}_{\mathbb{R}}$, $\text{Born-rank}_{\mathbb{R}} \geq \text{Born-rank}_{\mathbb{C}}$,
$\text{Born-rank}_{\mathbb{R}} \geq \text{puri-rank}_{\mathbb{R}}$, 
$\text{Born-rank}_{\mathbb{C}} \geq \text{puri-rank}_{\mathbb{C}}$, $\text{puri-rank}_{\mathbb{R}}\geq \text{puri-rank}_{\mathbb{C}}$,
$\text{TT-rank}_{\mathbb{R}_{\geq 0}} \geq \text{puri-rank}_{\mathbb{R}}$,
$\text{TT-rank}_{\mathbb{R}}=\text{TT-rank}_{\mathbb{C}}$.
\end{proposition}
Proposition \ref{sup_prop:inclusions} is proven via Lemmas~\ref{sup_lem:prop11}-~\ref{sup_lem:prop13} below:

\begin{lemma} \label{sup_lem:prop11} For all non-negative tensors,
$\text{TT-rank}_{\mathbb{R}_{\geq 0}} \geq \text{TT-rank}_{\mathbb{R}}$, $\text{Born-rank}_{\mathbb{R}} \geq \text{Born-rank}_{\mathbb{C}}$, $\text{Born-rank}_{\mathbb{R}} \geq \text{puri-rank}_{\mathbb{R}}$, $\text{Born-rank}_{\mathbb{C}} \geq \text{puri-rank}_{\mathbb{C}}$, $\text{puri-rank}_{\mathbb{R}}\geq \text{puri-rank}_{\mathbb{C}}$.
\end{lemma}
\begin{proof}
It is clear that enlarging the set of tensor elements can only reduce the corresponding rank. Moreover a BM is a LPS with purification dimension $\mu=1$.
\end{proof}

\begin{lemma}\label{sup_lem:prop12}
For all non-negative tensors, $\text{TT-rank}_{\mathbb{R}}=\text{TT-rank}_{\mathbb{C}}$.
\end{lemma}
\begin{proof}
The canonical MPS decomposition of a non-negative tensor can be obtained by successive singular value decompositions \cite{SchollwockReview}, and has the same TT-rank as the highest rank across a bipartition. Because the rank of a non-negative matrix is the same over $\mathbb{R}$ or $\mathbb{C}$, $\text{TT-rank}_{\mathbb{R}}=\text{TT-rank}_{\mathbb{C}}$.
\end{proof}

\begin{lemma}\label{sup_lem:prop13}For all non-negative tensors,
$\text{TT-rank}_{\mathbb{R}_{\geq 0}} \geq \text{puri-rank}_{\mathbb{R}}$ 
\end{lemma}
\begin{proof} 
Let us denote the non-negative tensors of an MPS of TT-rank$_{\mathbb{R}_{\geq 0}}=r$ as $A_i$. We define a LPS$_{\mathbb{R}}$ of purification index of size $\mu=r^2$ with the tensors 

\begin{align}
    B_{1,X_1}^{\beta_1,\alpha_1}&=\delta_{\alpha_1,\beta_1} \sqrt{A_{1,X_1}^{\alpha_1}},\\
    B_{N,X_N}^{\beta_N,\alpha_{N-1}}&=\delta_{\alpha_{N-1},\beta_N} \sqrt{A_{N,X_N}^{\alpha_{N-1}}},  \\
    B_{i,X_i}^{\beta_i,\alpha_{i-1},\alpha{i}}&=\delta_{\alpha_{i-1}r+\alpha_{i},\beta_i} \sqrt{A_{i,X_i}^{\beta_i,\alpha_{i-1},\alpha{i}}}.
\end{align}
We now observe that 
\begin{align}
     \sum_{\beta_i} B_{i,X_i}^{\beta_i,\alpha_{i-1},\alpha{i}}\overline{B}_{i,X_i}^{\beta_i,\alpha'_{i-1},\alpha'{i}} & =\sum_{\beta_i} \delta_{\alpha_{i-1}r+\alpha_{i},\beta_i}\delta_{\alpha'_{i-1}r+\alpha'_{i},\beta_i}\sqrt{A_{i,X_i}^{\beta_i,\alpha_{i-1},\alpha{i}}}\sqrt{A_{i,X_i}^{\beta_i,\alpha'_{i-1},\alpha'{i}}} \\ &=\delta_{\alpha_{i-1},\alpha'_{i-1}}\delta_{\alpha_{i},\alpha'_{i}} A_{i,X_i}^{\alpha_{i-1},\alpha{i}},
\end{align}
or equivalently, in graphical notation, that
\begin{align}\label{sup_eqn:lem3proof}
\input{figuressup/proof2.tex}
\end{align}
Therefore, this LPS defines the same tensor as the original MPS and has $\text{puri-rank}_{\mathbb{R}}=r$.
\end{proof}

We now turn to Proposition~\ref{sup_prop:strict_inclusions}, showing that all inequalities given in Proposition~\ref{sup_prop:inclusions} can in fact be strict, and that for all other pairs of representations there exist probability distributions (non-negative tensors) showing that neither rank can always be lower than the other.

\begin{proposition}
\label{sup_prop:strict_inclusions}
The ranks of all introduced tensor-network representations satisfy the properties contained in Table~\ref{sup_table:sets}. Specifically, denoting by $r_{\text{row}}$ ($r_{\text{column}}$) the rank appearing in the row (column), $<$ indicates that there exists a tensor satisfying $r_{\text{row}} < r_{\text{column}}$ and $<,>$ indicates that there exists both a tensor satisfying $r_{\text{row}}  < r_{\text{column}}$ and another tensor satisfying $r_{\text{column}} > r_{\text{row}}$.
\end{proposition}
\setcounter{table}{0}
\begin{table}[ht]
\caption{Results of Proposition~\ref{sup_prop:strict_inclusions}}
\label{sup_table:sets}
\centering
\begin{tabular}{@{} lcccccc @{}}
\toprule
& TT-rank$_{\mathbb{R}}$ & TT-rank$_{\mathbb{R}_{\geq 0}}$  & Born-rank$_{\mathbb{R}}$ & Born-rank$_{\mathbb{C}}$ & puri-rank$_{\mathbb{R}}$ & puri-rank$_{\mathbb{C}}$  \\
\midrule
TT-rank$_{\mathbb{R}}$ & = & $<$ & $<,>$ & $<,>$& $<,>$& $<,>$\\
TT-rank$_{\mathbb{R}_{\geq 0}}$ & $>$ & = &$<,>$  & $<,>$ & $>$ & $>$\\
Born-rank$_{\mathbb{R}}$  & $<,>$ & $<,>$ & = & $>$ & $>$ & $>$\\
Born-rank$_{\mathbb{C}}$  & $<,>$ & $<,>$ & $<$ & =& $<,>$ & $>$\\
puri-rank$_{\mathbb{R}}$ & $<,>$ & $<$ & $<$ &$<,>$ & = & $>$\\
puri-rank$_{\mathbb{C}}$ & $<,>$ & $<$ & $<$ & $<$ & $<$ &= \\
\bottomrule
\end{tabular}
\end{table}

Again, we prove Proposition~\ref{sup_prop:strict_inclusions} via Lemmas~\ref{sup_lem:matA}-\ref{sup_lem:matG}, each of which addresses a subset of the entries in Table~\ref{sup_table:sets}. The particular entry addressed by a specific lemma is indicated by $[k,l]_>$ or $[k,l]_<$, where $k$ denotes the row and $l$ the column of Table~\ref{sup_table:sets}, and the subscript $<$ ($>$) is used to indicate a specific case. Note also that a tensor providing a proof for entry $[k,l]_>$ ($[k,l]_<$) provides also a proof for entry $[l,k]_<$ ($[l,k]_>$).

\begin{lemma}[${[1,2]}$]\label{sup_lem:matA}
There exists a non-negative matrix $A$ with $\text{TT-rank}_{\mathbb{R}} <\text{TT-rank}_{\mathbb{R}_{\geq 0}}$.
\end{lemma}

\begin{proof}
Consider the matrix $A=\left(\begin{matrix}
0& 1& 1& 0 \\
0& 0& 1& 1\\
1& 0& 0& 1 \\
1& 1& 0& 0
\end{matrix}\right)$. $A$ has $\text{TT-rank}_\mathbb{R}=3$ and $\text{TT-rank}_{\mathbb{R}\geq 0}=4$ \cite{COHEN1993149}.
\end{proof}

\begin{lemma}[${[1,3]}_<$, ${[2,3]}_<$, ${[3,4]}$, ${[3,5]}$, ${[3,6]}$]\label{sup_lem:mat_B}
There exists a non-negative matrix $B$ with $\text{TT-rank}_{\mathbb{R}} <\text{Born-rank}_{\mathbb{R}}$, $\text{TT-rank}_{\mathbb{R}_{\geq 0}} < \text{Born-rank}_{\mathbb{R}}$ and $\text{Born-rank}_{\mathbb{R}} > \text{puri-rank}_{\mathbb{R}}$.
\end{lemma}

\begin{proof}
Consider the matrix $B=\left(\begin{matrix}
2& 1& 1 \\
1& 0& 1\\
1& 1& 0
\end{matrix}\right)$. $B$ has $\text{TT-rank}_\mathbb{R}=2$ and $\text{TT-rank}_{\mathbb{R}\geq 0}=2$. Moreover the square root element-wise of $B$ is $\left(\begin{matrix}
\sqrt{2}& 1& 1 \\
1& 0& 1\\
1& 1& 0
\end{matrix}\right)$, which has rank 3, as well as all square roots obtained by changing signs of each element, so $\text{Born-rank}_\mathbb{R}=3$. On the other hand it is also possible to write $B$ as the absolute value squared of $\left(\begin{matrix}
1+i& 1& 1 \\
1& 0& 1\\
i& 1& 0
\end{matrix}\right)$, which has rank 2, so $\text{Born-rank}_\mathbb{C}=2$. Furthermore, from Proposition \ref{sup_prop:inclusions} and the fact that $\text{TT-rank}_{\mathbb{R}\geq 0}=2$,  we have that $\text{puri-rank}_{\mathbb{R}/\mathbb{C}} \leq 2$.
\end{proof}

\begin{lemma}[${[2,5]}$, ${[2,6]}$, ${[2,3]}_>$, ${[2,4]}_>$, ${[1,3]}_>$, ${[1,4]}_>$, ${[1,5]}_>$, ${[1,6]}_>$]
There exists a non-negative matrix $C$ with $\text{TT-rank}_{\mathbb{R}_{\geq 0}} > \text{puri-rank}_{\mathbb{R}}$, $\text{TT-rank}_{\mathbb{R}_{\geq 0}} > \text{puri-rank}_{\mathbb{C}}$, $\text{TT-rank}_{\mathbb{R}_{\geq 0}} > \text{Born-rank}_{\mathbb{R}}$,
$\text{TT-rank}_{\mathbb{R}_{\geq 0}} > \text{Born-rank}_{\mathbb{C}}$,
$\text{TT-rank}_{\mathbb{R}} > \text{Born-rank}_{\mathbb{R}}$,
$\text{TT-rank}_{\mathbb{R}} > \text{Born-rank}_{\mathbb{C}}$,
$\text{TT-rank}_{\mathbb{R}} > \text{puri-rank}_{\mathbb{R}}$ and $\text{TT-rank}_{\mathbb{R}} > \text{puri-rank}_{\mathbb{C}}$.
\end{lemma}

\begin{proof} Consider the matrix
$C=\left(\begin{matrix}
4& 1& 1 \\
1& 0& 1 \\
1& 1& 0
\end{matrix}\right)$. $C$ has $\text{TT-rank}_{\mathbb{R}\geq 0}=\text{TT-rank}_\mathbb{R}=3$, but the square root is matrix $B$ from Lemma \ref{sup_lem:mat_B}, of rank $2$, so $\text{Born-rank}_\mathbb{R}=\text{Born-rank}_\mathbb{C}=2$. Again, from Proposition \ref{sup_prop:inclusions} and the fact that $\text{Born-rank}_\mathbb{R}=2$ we have that $\text{puri-rank}_{\mathbb{R}/\mathbb{C}} \leq 2$.
\end{proof}

\begin{lemma}[${[1,4]_<}$, ${[1,5]_<}$, ${[1,6]_<}$]
There exists a non-negative matrix $D$ with
$\text{TT-rank}_{\mathbb{R}} < \text{Born-rank}_{\mathbb{C}}$, $\text{TT-rank}_{\mathbb{R}} < \text{puri-rank}_{\mathbb{R}}$ and $\text{TT-rank}_{\mathbb{R}} < \text{puri-rank}_{\mathbb{C}}$.
\end{lemma}

\begin{proof}
Consider $a=(1+\sqrt{5})/2$ and define $D=\left(\begin{matrix}
0& 1& a& 1& 0\\
0& 0& 1& a& 1\\
1& 0& 0& 1& a\\
a& 1& 0& 0& 1\\
1& a& 1& 0& 0
\end{matrix}\right)$. $D$ is the slack matrix of a regular pentagon, and has $\text{TT-rank}_{\mathbb{R}\geq 0}=5$ while $\text{TT-rank}_\mathbb{R}=3$. $D$ has $\text{puri-rank}_\mathbb{R}=4$ and $\text{puri-rank}_\mathbb{C}=4$, as proven in ref.~\cite{Goucha2017}. By Proposition \ref{sup_prop:inclusions} and the fact that $\text{puri-rank}_\mathbb{C}=4$ we have that $\text{Born-rank}_{\mathbb{C}} \geq 4$.
\end{proof}

\begin{lemma}[${[5,6]}$, ${[4,5]_<}$]
There exists a non-negative matrix $E$ with
$\text{puri-rank}_{\mathbb{R}} > \text{puri-rank}_{\mathbb{C}}$ and
$\text{Born-rank}_{\mathbb{C}} < \text{puri-rank}_{\mathbb{R}}$.
\end{lemma}

\begin{proof}
Consider the matrix $E=\left(\begin{matrix}
0& 1& 1& 1\\
1& 0& 1& 1\\
1& 1& 0& 1\\
1& 1& 1& 0
\end{matrix}\right)$. $E$ can be written as the product of $\left(\begin{matrix}
1& 0\\
0& 1\\
1& -1\\
1& e^{2i\pi/3}
\end{matrix}\right)$ and $\left(\begin{matrix}
0& 1& 1& 1\\
1& 0& 1& -e^{-2i\pi/3}
\end{matrix}\right)$, which shows that
$\text{Born-rank}_\mathbb{C}\leq2$, and therefore, $\text{puri-rank}_\mathbb{C}\leq2$ by Proposition \ref{sup_prop:inclusions}. Bounds on the real positive semidefinite rank imply that it is equal to 3 \cite{Fawzi2015}. 
\end{proof}

\begin{lemma}[${[4,6]}$, ${[2,4]_<}$, ${[4,5]_>}$]
\label{sup_lem:matG}
There exists a non-negative matrix $F$ with 
$\text{puri-rank}_{\mathbb{C}} < \text{Born-rank}_{\mathbb{C}}$, 
$\text{TT-rank}_{\mathbb{R} \geq 0} < \text{Born-rank}_{\mathbb{C}}$ and
$\text{Born-rank}_{\mathbb{C}} > \text{puri-rank}_{\mathbb{R}}$.
\end{lemma}

\begin{proof}
Consider the matrix $F=\left(\begin{matrix}
1& 0& 0& 1& 1& 0& 1\\
0& 1& 0& 0& 1& 1& 1\\
0& 0& 1& 1& 0& 1& 1\\
1& 0& 1& 2& 1& 1& 2\\
1& 1& 0& 1& 2& 1& 2\\
0& 1& 1& 1& 1& 2& 2\\
1& 1& 1& 2& 2& 2& 3
\end{matrix}\right)$. $F$ has $\text{TT-rank}_\mathbb{R}=3$ and is equal to the product of $\left(\begin{matrix}
0& 0& 1\\
1& 0& 0\\
0& 1& 0\\
0& 1& 1\\
1& 0& 1\\
1& 1& 0\\
1& 1& 1
\end{matrix}\right)$ and its transpose, so $F$ has $\text{TT-rank}_{\mathbb{R}\geq 0}=3$. In addition, we now prove that $F$ has $\text{Born-rank}_\mathbb{C}\geq 4$. To this end, consider a complex Hadamard square root of matrix $F$ given by
\begin{align}
\sqrt{F}=\left(\begin{matrix}
\Ph{1}& 0& 0& \Ph{2}& \Ph{3}& 0& \Ph{4}\\
0& \Ph{5}& 0& 0& \Ph{6}& \Ph{7}& \Ph{8}\\
0& 0& \Ph{9}& \Ph{10}& 0& \Ph{11}& \Ph{12}\\
\Ph{13}& 0& \Ph{14}& \sqrt{2}\Ph{15}& \Ph{16}& \Ph{17}& \sqrt{2}\Ph{18}\\
\Ph{19}& \Ph{20}& 0& \Ph{21}& \sqrt{2}\Ph{22}& \Ph{23}& \sqrt{2}\Ph{24}\\
0& \Ph{25}& \Ph{26}& \Ph{27}& \Ph{28}& \sqrt{2}\Ph{29}& \sqrt{2}\Ph{30}\\
\Ph{31}& \Ph{32}& \Ph{33}& \sqrt{2}\Ph{34}& \sqrt{2}\Ph{35}& \sqrt{2}\Ph{36}& \sqrt{3}\Ph{37}
\end{matrix}\right),
\end{align}
where the $\phi_i$ are real parameters. We will prove that the rank of $\sqrt{F}$ is at least 4. First observe that the rank is invariant under multiplication of a row or a column by a phase. By performing such operations in the right order, we obtain that the rank of $\sqrt{F}$ is the same as the rank of a matrix
\begin{align}
M=\left(\begin{matrix}
1& 0& 0& 1& \Ph{1}& 0& 1\\
0& 1& 0& 0& 1& \Ph{2}& 1\\
0& 0& 1& \Ph{3}& 0& 1& 1\\
\Ph{4}& 0& \Ph{5}& \sqrt{2}& \Ph{6}& \Ph{7}& \sqrt{2}\Ph{8}\\
\Ph{9}& \Ph{10}& 0& \Ph{11}& \sqrt{2}& \Ph{12}& \sqrt{2}\Ph{13}\\
0& \Ph{14}& \Ph{15}& \Ph{16}& \Ph{17}& \sqrt{2}& \sqrt{2}\Ph{18}\\
\Ph{19}& \Ph{20}& \Ph{21}& \sqrt{2}\Ph{22}& \sqrt{2}\Ph{23}& \sqrt{2}\Ph{24}& \sqrt{3}
\end{matrix}\right),
\end{align}
with new real parameters $\phi_i$ (defined modulo $2\pi$). We will prove that such a matrix has always rank at least 4. It is clear that the first three rows are independent, so the rank is at least 3. Now suppose that the rank is 3, the rows 4 to 7 are therefore complex linear combinations of the first 3 rows. Let us write such a linear combination for row 4:
\begin{align}
(4)=\alpha(1)+\beta(2)+\gamma(3).
\end{align}
The first columns imply that $\alpha=\Ph{4}$, $\beta=0$ and $\gamma=\Ph{5}$.
Moreover we have
\begin{align}
    \Ph{4}+\Ph{5}\Ph{3}&=\sqrt{2},\\
    \Ph{4}+\Ph{5}&=\sqrt{2}\Ph{8}.
\end{align}
Let us take the absolute value squared of these equations, we obtain
\begin{align}
    2+2\cos(\phi_4-\phi5-\phi3)&=2,\\
    2+2\cos(\phi_4-\phi5)&=2.
\end{align}
Therefore, $\cos(\phi_4-\phi_5-\phi_3)=\cos(\phi_4-\phi_5)=0$, which implies that $\phi_4-\phi_5=\pm \pi/2$ and $\phi_3=0$ or $\pi$, so that $\Ph{3}=\pm 1$. By similarly writing that row 5 and 6 are linear combinations of the first three rows, we obtain by symmetry that $\Ph{1}=\pm 1$ and 
$\Ph{2}=\pm 1$.
Let us now show that the last row cannot be written as a linear combination of the first three rows. Suppose this is the case, so that
\begin{align}
(7)=\alpha(1)+\beta(2)+\gamma(3).
\end{align}
Then the first columns imply that $\alpha=\Ph{19}$, $\beta=\Ph{20}$ and $\gamma=\Ph{21}$. We know that $\Ph{1}=\pm 1$, $\Ph{2}=\pm 1$ and $\Ph{3}=\pm 1$.
We then have 
    \begin{align}
        \Ph{19}\pm\Ph{21}&=\sqrt{2}\Ph{22},\\
        \pm\Ph{19}+\Ph{20}&=\sqrt{2}\Ph{23},\\
        \pm\Ph{20}+\Ph{21}&=\sqrt{2}\Ph{24}.
    \end{align}
    From this we obtain, by taking the absolute value squared,
    \begin{align}
    \cos(\phi_{19}-(\phi_{21}\pm\pi))&=0,\\
    \cos((\phi_{19}\pm\pi)-\phi_{20})&=0,\\
    \cos((\phi_{20}\pm\pi)-\phi_{21})&=0,
    \end{align}
    which implies 
    \begin{align}
    \phi_{19}-\phi_{21}&=\pm\pi/2,\\
    \phi_{19}-\phi_{20}&=\pm\pi/2,\\
    \phi_{20}-\phi_{21}&=\pm\pi/2,
    \end{align}
    which is impossible.
We therefore conclude that $M$, and thus also $\sqrt{F}$, has rank at least 4.
\end{proof}

Before continuing to Proposition \ref{sup_prop:overheads_1}, it is interesting to note that the proofs of Lemma's \ref{sup_lem:matA} - \ref{sup_lem:matG} all involve lower-bounding a given rank, and that this problem may be cast into the form of a polynomial optimization problem, for which hierarchies of semi-definite
relaxations are available~\cite{Lasserre}. For example, the non-negative
rank TT-rank$_{\mathbb{R}\geq 0}$ 
can for a given $d\times d$-matrix $T$
be computed via the minimization problem
\begin{equation}
\min \|T- C\|_2   
\end{equation}  
subject to $C= A B$, where $A$ and $B$ are $d\times k$ and $k\times d$
matrices with non-negative entries, respectively. A hierarchy of convex relaxations can then be used to provide increasingly better approximations to the optimal solution, and Kuhn-Tucker conditions
can be made use of to check for global optimality of a solution. In practice the required relaxations can soon become infeasibly large, however this strategy is worth noting as a potentially interesting tool, particularly for the complex Hadamard square root rank.

Finally, we move onto the proof of Proposition \ref{sup_prop:overheads_1}, addressing the question of the overheads required to exactly represent a tensor network representation of a given rank with an alternative representation.

\begin{proposition}
\label{sup_prop:overheads_1}
The ranks of all introduced tensor-network representations satisfy the relationships without asterisk contained in Table~\ref{sup_table:ranks}. A function $g(x)$ denotes that for all non-negative tensors $r_{\text{row}}\leq g(r_{\text{column}})$. ``No'' indicates that there exists a family of probability distributions of increasing $N$ with $d=2$ and $r_{\text{column}}$ constant, but such that $r_{\text{row}}$ goes to infinity, i.e., that no such function can exist.
\end{proposition}

\begin{table}[ht]
\caption{Results of Proposition~\ref{sup_prop:overheads_1}.}
\label{sup_table:ranks}
\centering
\begin{tabular}{@{} lcccccc @{}}
\toprule
& TT-rank$_{\mathbb{R}}$ & TT-rank$_{\mathbb{R}_{\geq 0}}$  & Born-rank$_{\mathbb{R}}$ & Born-rank$_{\mathbb{C}}$ & puri-rank$_{\mathbb{R}}$ & puri-rank$_{\mathbb{C}}$  \\
\midrule
TT-rank$_{\mathbb{R}}$ & = & $\leq x $ & $\leq x^2$ & $\leq x^2$& $\leq x^2$& $\leq x^2$\\
TT-rank$_{\mathbb{R}_{\geq 0}}$ & No & = & No  & No &  No  &  No \\
Born-rank$_{\mathbb{R}}$  &  No  & No  & = & No & No & No \\
Born-rank$_{\mathbb{C}}$  & No & No$^*$ & $\leq x$ & =& No$^*$ & No$^*$\\
puri-rank$_{\mathbb{R}}$ & No & $\leq x$ & $\leq x$ &$\leq 2x$ & = & $\leq 2x$\\
puri-rank$_{\mathbb{C}}$ & No & $\leq x$ & $\leq x$ & $\leq x$ & $\leq x$ &=
\end{tabular}
\end{table}

Once again, it is convenient to prove Proposition~\ref{sup_prop:overheads_1} via a series of lemmas. However, note first that all entries of Table~\ref{sup_table:ranks} containing the function $g(x) = x$ follow straightforwardly from Proposition~\ref{sup_prop:inclusions}. Given this, we begin with Lemmas \ref{sup_lemma:overheads_explicit_1} and \ref{sup_lemma:overheads_explicit_2} addressing the remaining entries of Table~\ref{sup_table:ranks} for which explicit functions can be found:

\begin{lemma}
\label{sup_lemma:overheads_explicit_1}
For all non-negative tensors $\text{TT-rank}_{\mathbb{R}}\leq (\text{puri-rank}_{\mathbb{C}})^2$, therefore, also $\text{TT-rank}_{\mathbb{R}}\leq (\text{puri-rank}_{\mathbb{R}})^2$, $\text{TT-rank}_{\mathbb{R}}\leq (\text{Born-rank}_{\mathbb{R}})^2$ and $\text{TT-rank}_{\mathbb{R}}\leq (\text{Born-rank}_{\mathbb{C}})^2$.
\end{lemma}
\begin{proof}
Consider an LPS$_\mathbb{C}$ of puri-rank$_\mathbb{C}=r$. Let us denote the tensors defining this LPS as $A_{i,X_i}^{\beta_i,\alpha_{i-1},\alpha_i}$. Define new tensors $B_{i,X_i}^{\alpha_{i-1},r+\alpha'_{i-1},\alpha_{i},r+\alpha'_{i}}=\sum_{\beta_i} A_{i,X_i}^{\beta_i,\alpha_{i-1},\alpha_i}\overline{A}_{i,X_i}^{\beta_i,\alpha'_{i-1},\alpha'_i}$. As shown in Equation~\eqref{sup_eqn:lem11proof}, these tensors define an MPS$_{\mathbb{R}}$ of TT-rank$_{\mathbb{R}}=r^2$ corresponding to the same probability mass function as the original LPS.
\begin{align}\label{sup_eqn:lem11proof}
\begin{tikzpicture}[y=0.80pt, x=0.80pt, yscale=-1.000000, xscale=1.000000, inner sep=0pt, outer sep=0pt,baseline=(current bounding box.center)]
\path[draw=black,line join=miter,line cap=butt,miter limit=4.00,line
  width=1.600pt] (237.6880,277.1836) -- (287.1334,277.1836);
\path[draw=black,line join=miter,line cap=butt,line width=0.800pt]
  (178.6383,252.3005) -- (190.3813,252.4267) -- (190.3813,302.6187) --
  (179.2065,302.6187);
\path[fill=cffffff,even odd rule] (188.9481,265.4935) -- (191.5151,265.4935) --
  (191.5151,267.5917) -- (188.9258,267.5917);
\path[fill=cffffff,even odd rule] (188.9481,288.6185) -- (191.5151,288.6185) --
  (191.5151,290.7167) -- (188.9258,290.7167);
\path[draw=black,line join=miter,line cap=butt,miter limit=4.00,line
  width=0.800pt] (155.7542,289.6950) -- (194.1071,289.6950) --
  (202.3424,289.6950);
\path[draw=black,line join=miter,line cap=butt,line width=0.800pt]
  (179.0402,266.5879) -- (179.0402,290.7844);
\path[draw=black,line join=miter,line cap=butt,miter limit=4.00,line
  width=0.800pt] (179.2322,296.6071) -- (179.2322,303.0127);
\path[draw=black,fill=cffffff,line join=miter,line cap=butt,miter
  limit=4.00,even odd rule,line width=0.800pt,rounded corners=0.0000cm]
  (171.4686,282.3938) rectangle (186.4911,297.0249);
\path[draw=black,line join=miter,line cap=butt,miter limit=4.00,line
  width=0.800pt] (179.0369,260.0859) -- (179.0369,241.3113);
\path[draw=black,line join=miter,line cap=butt,miter limit=4.00,line
  width=0.800pt] (155.7542,266.5861) -- (203.2353,266.5861);
\path[xscale=1.000,yscale=-1.000,draw=black,fill=cffffff,line join=miter,line
  cap=butt,miter limit=4.00,even odd rule,line width=0.800pt,rounded
  corners=0.0000cm] (171.2733,-274.2992) rectangle (186.2959,-259.6681);
\path[draw=black,dash pattern=on 0.82pt off 0.82pt,miter limit=4.00,line
  width=0.823pt,rounded corners=0.0000cm] (162.6539,249.4434) rectangle
  (196.6909,304.6603);
\path[cm={{1.9625,0.0,0.0,1.9625,(-171.7447,-242.51634)}},color=black,fill=black,line
  width=1.600pt]
  (179.6652,252.1390)arc(-0.125:180.125:0.893)arc(-180.125:0.125:0.893) --
  cycle;
\path[draw=black,line join=miter,line cap=butt,miter limit=4.00,line
  width=0.800pt] (262.5738,267.8571) -- (262.5738,242.8325);
\path[xscale=1.000,yscale=-1.000,draw=black,fill=cffffff,line join=miter,line
  cap=butt,miter limit=4.00,even odd rule,line width=0.800pt,rounded
  corners=0.0000cm] (249.6871,-289.9607) rectangle (275.0113,-265.2966);
\begin{scope}[cm={{0.36159,0.0,0.0,0.36159,(100.90983,152.9081)}}]
  \begin{scope}[cm={{1.06299,0.0,0.0,-1.06299,(-186.02362,789.27165)}},draw=black,line join=miter,line cap=butt,miter limit=10.43]
    \path[draw,fill=black,line width=0.000pt] (489.7500,422.5000) --
      (489.8900,422.5000) -- (490.0400,422.5000) -- (490.1800,422.5000) --
      (490.2500,422.5100) -- (490.3300,422.5100) -- (490.4000,422.5200) --
      (490.4700,422.5300) -- (490.5400,422.5400) -- (490.6100,422.5500) --
      (490.6800,422.5600) -- (490.7400,422.5800) -- (490.8100,422.6000) --
      (490.8700,422.6200) -- (490.9300,422.6400) -- (490.9900,422.6700) --
      (491.0500,422.7000) -- (491.1000,422.7400) -- (491.1500,422.7700) --
      (491.2000,422.8200) -- (491.2400,422.8600) -- (491.2800,422.9100) --
      (491.3000,422.9400) -- (491.3200,422.9700) -- (491.3300,422.9900) --
      (491.3500,423.0200) -- (491.3600,423.0500) -- (491.3800,423.0900) --
      (491.3900,423.1200) -- (491.4000,423.1600) -- (491.4100,423.1900) --
      (491.4200,423.2300) -- (491.4200,423.2700) -- (491.4300,423.3100) --
      (491.4300,423.3500) -- (491.4400,423.3900) -- (491.4400,423.4300) --
      (491.4400,423.4800) .. controls (491.4400,424.4800) and (490.4900,424.4800) ..
      (489.8000,424.4800) -- (459.9600,424.4800) .. controls (459.2500,424.4800) and
      (458.3200,424.4800) .. (458.3200,423.4800) .. controls (458.3200,422.5000) and
      (459.2500,422.5000) .. (460.0000,422.5000) -- cycle;
    \path[draw,fill=black,line width=0.000pt] (489.8000,412.8200) --
      (489.8700,412.8200) -- (489.9300,412.8200) -- (490.0700,412.8300) --
      (490.2100,412.8300) -- (490.2800,412.8300) -- (490.3500,412.8400) --
      (490.4200,412.8500) -- (490.4900,412.8500) -- (490.5500,412.8600) --
      (490.6200,412.8800) -- (490.6900,412.8900) -- (490.7500,412.9100) --
      (490.8200,412.9300) -- (490.8800,412.9500) -- (490.9400,412.9700) --
      (491.0000,413.0000) -- (491.0500,413.0300) -- (491.1000,413.0700) --
      (491.1500,413.1100) -- (491.2000,413.1500) -- (491.2400,413.2000) --
      (491.2600,413.2200) -- (491.2800,413.2500) -- (491.3000,413.2700) --
      (491.3200,413.3000) -- (491.3300,413.3300) -- (491.3500,413.3600) --
      (491.3600,413.3900) -- (491.3800,413.4200) -- (491.3900,413.4600) --
      (491.4000,413.4900) -- (491.4100,413.5300) -- (491.4200,413.5700) --
      (491.4200,413.6100) -- (491.4300,413.6500) -- (491.4300,413.6900) --
      (491.4400,413.7300) -- (491.4400,413.7800) -- (491.4400,413.8200) .. controls
      (491.4400,414.8200) and (490.4900,414.8200) .. (489.7500,414.8200) --
      (460.0000,414.8200) .. controls (459.2500,414.8200) and (458.3200,414.8200) ..
      (458.3200,413.8200) .. controls (458.3200,412.8200) and (459.2500,412.8200) ..
      (459.9600,412.8200) -- cycle;
  \end{scope}
\end{scope}
\path[fill=black] (169.1071,235.3979) node[above right] (text4969) {$X_i$};
\path[fill=black] (252.5931,235.8008) node[above right] (text4969-1) {$X_i$};
\path[fill=black,line join=miter,line cap=butt,line width=0.800pt]
  (155.1786,263.5911) node[above right] (text983) {$r$};
\path[fill=black,line join=miter,line cap=butt,line width=0.800pt]
  (155.1392,287.3193) node[above right] (text983-0) {$r$};
\path[fill=black,line join=miter,line cap=butt,line width=0.800pt]
  (199.6035,263.5693) node[above right] (text983-0-5) {$r$};
\path[fill=black,line join=miter,line cap=butt,line width=0.800pt]
  (199.6035,287.1408) node[above right] (text983-0-5-3) {$r$};
\path[fill=black,line join=miter,line cap=butt,line width=0.800pt]
  (237.4607,273.3908) node[above right] (text983-0-5-6) {$r^2$};
\path[fill=black,line join=miter,line cap=butt,line width=0.800pt]
  (278.9445,273.2518) node[above right] (text983-0-5-6-5) {$r^2$};
\path[fill=black,line join=miter,line cap=butt,line width=0.800pt]
  (174.9330,270.8737) node[above right] (text1077) {$A$};
\path[fill=black,line join=miter,line cap=butt,line width=0.800pt]
  (258.3195,281.4575) node[above right] (text1077-5) {$B$};
\end{tikzpicture}
\end{align}
\end{proof}

\begin{lemma}
\label{sup_lemma:overheads_explicit_2}
For all non-negative tensors $\text{puri-rank}_{\mathbb{R}}\leq 2\text{puri-rank}_{\mathbb{C}}$ and $\text{puri-rank}_{\mathbb{R}}\leq 2\text{Born-rank}_{\mathbb{C}}$.
\end{lemma}
\begin{proof}
Consider an LPS$_\mathbb{C}$ with purification dimension equal to $\mu$ and puri-rank$_\mathbb{C}$ equal to $r$, constructed from tensors $A_{i}$. If we fix $i$ and $X_i$, $A_{i}$ is an $r\times \mu$ matrix for $i=1$ or $i=N$ and an order-3 tensor of size $r\times r \times \mu$ otherwise. Now define new tensors by blocks as 
\begin{align}
B_1=\left(\begin{matrix} \text{Re}(A_1) & -\text{Im}(A_1) \\ \text{Im}(A_1) & \text{Re}(A_1)
\end{matrix}\right), \ \ \ B_N=\left(\begin{matrix} \text{Re}(A_N) & \text{Im}(A_1) \\ -\text{Im}(A_N) & \text{Re}(A_N)
\end{matrix}\right),\\
B_i=\definecolor{c050505}{RGB}{5,5,5}
\begin{tikzpicture}[y=0.8pt, x=0.8pt,yscale=-0.7, xscale=0.7,  baseline={([yshift=-18pt]current bounding box.center)}]
\begin{scope}[cm={{0.44864,0.0,0.0,0.44864,(1035.2664,-127.35645)}}]
  \path[color=black,draw=c050505,line join=round,line cap=butt,miter
    limit=4.00,line width=0.800pt] (-1374.9136,810.3803) -- (-1374.9136,1087.9972)
    -- (-1019.7770,920.5721) -- (-1019.7770,642.9552) -- cycle;
  \path[color=black,draw=c050505,dash pattern=on 4.80pt off 4.80pt,line
    join=round,line cap=butt,miter limit=4.00,line width=0.800pt]
    (-1292.0864,920.5721) -- (-1019.7770,920.5721);
  \path[color=black,draw=c050505,line join=round,line cap=butt,miter
    limit=4.00,line width=0.800pt] (-1019.7770,920.5721) -- (-1019.7770,642.9552)
    -- (-1292.0864,642.9552);
  \path[color=black,draw=c050505,dash pattern=on 4.80pt off 4.80pt,line
    join=round,line cap=butt,miter limit=4.00,line width=0.800pt]
    (-1292.0864,920.5721) -- (-1292.0864,642.9552);
  \path[color=black,draw=c050505,line join=round,line cap=butt,miter
    limit=4.00,line width=0.800pt] (-1292.0864,642.9552) -- (-1647.2230,810.3803)
    -- (-1647.2230,1087.9972);
  \path[color=black,draw=c050505,dash pattern=on 4.80pt off 4.80pt,line
    join=round,line cap=butt,miter limit=4.00,line width=0.800pt]
    (-1647.2230,1087.9972) -- (-1292.0864,920.5721);
  \path[color=black,draw=c050505,line join=round,line cap=butt,miter
    limit=4.00,line width=0.800pt] (-1374.9136,810.3803) -- (-1647.2230,810.3803)
    -- (-1647.2230,1087.9972) -- (-1374.9136,1087.9972) -- cycle;
\end{scope}
\path[fill=black] (293,272) node[above right] (text3261)
  {$\text{Re}(A_i)$};
\path[fill=black] (361.6409,343.3218) node[above right] (text3261-8)
  {$\text{Re}(A_i)$};
\path[fill=black] (292.3552,343.3504) node[above right] (text3261-8-5)
  {$-\text{Im}(A_i)$};
\path[fill=black] (452.7915,195) node[above right] (text3261-8-5-4)
  {$\text{Im}(A_i)$};
\path[fill=black] (515,275) node[above right] (text3261-8-5-8)
  {$\text{Im}(A_i)$};
\path[fill=black] (452.6409,275) node[above right] (text3261-8-2)
  {$\text{Re}(A_i)$};
\path[fill=black] (510,195) node[above right] (text3261-8-2-1)
  {$-\text{Re}(A_i)$};
\path[fill=black] (360,272) node[above right] (text3261-8-5-0)
  {$\text{Im}(A_i)$};
\begin{scope}[cm={{3.77953,0.0,0.0,3.77953,(3.55381,-449.41764)}}]
  \path[color=black,fill=black,even odd rule] (68.5469,175.8574) --
    (68.2812,175.8594) -- (68.3281,183.9629) -- (68.5937,183.9609) -- cycle;
  \path[draw=black,fill=black,line join=round,even odd rule,line width=0.146pt]
    (69.6184,181.1262) -- (68.4676,184.3128) -- (67.2798,181.1398) .. controls
    (67.9731,181.6438) and (68.9176,181.6354) .. (69.6184,181.1262) -- cycle;
\end{scope}
\begin{scope}[cm={{3.77953,0.0,0.0,3.77953,(3.55381,-449.41764)}}]
  \path[color=black,fill=black,even odd rule] (68.4141,175.7266) --
    (68.4141,175.9902) -- (77.3438,175.9902) -- (77.3438,175.7266) -- cycle;
  \path[draw=black,fill=black,line join=round,even odd rule,line width=0.146pt]
    (74.5149,174.6847) -- (77.6947,175.8540) -- (74.5149,177.0233) .. controls
    (75.0229,176.3329) and (75.0199,175.3884) .. (74.5149,174.6847) -- cycle;
\end{scope}
\begin{scope}[cm={{3.77953,0.0,0.0,3.77953,(3.55381,-449.41764)}}]
  \path[color=black,fill=black,even odd rule] (76.7793,171.6953) --
    (68.3555,175.7402) -- (68.4707,175.9785) -- (76.8945,171.9336) -- cycle;
  \path[draw=black,fill=black,line join=round,even odd rule,line width=0.146pt]
    (73.7794,171.9807) -- (77.1522,171.6587) -- (74.7915,174.0890) .. controls
    (74.9507,173.2468) and (74.5393,172.3965) .. (73.7794,171.9807) -- cycle;
\end{scope}
\path[fill=black] (244.4752,265.2923) node[above right] (text5241) {$2r$};
\path[fill=black] (284.8157,234.9169) node[above right] (text5241-2) {$2r$};
\path[fill=black] (278.2701,196.0565) node[above right] (text5241-2-8) {$2\mu$};
\end{tikzpicture}, \ \ \forall i\in\{2,\ldots,N-1\}.
\end{align}
Here $B_i$ is an order-3 tensor defined by blocks, where each block has dimension $r\times r\times \mu$. These tensors define a real LPS with purification dimension equal to $2\mu$ and puri-rank$_\mathbb{R}$ equal to $2r$ which represents the same probability mass function as the original complex LPS. Applying this result when $\mu=1$ shows that $\text{puri-rank}_{\mathbb{R}}\leq 2\text{Born-rank}_{\mathbb{C}}$.
\end{proof}

We now move on to the proofs for the ``No'' entries of Table~\ref{sup_table:ranks}. As discussed in the main text, for each ``No'' entry the strategy is to first prove the existence of a family of non-negative matrices (probability distributions over two discrete random variables) with the property that $r_{\text{column}}$ remains constant with respect to the dimension of the matrix, while $r_{\text{row}}$ grows. To this end, consider lemmas \ref{sup_lemma:matrix_families_ngons}-\ref{sup_lemma:matrix_families_born_ranks}, each of which addresses a specific entry of Table~\ref{sup_table:ranks}, for the restricted case of only two random variables:

\begin{lemma}[${[6,1]}$]
\label{sup_lemma:matrix_families_ngons}
There exists a family of non-negative matrices, of increasing dimension $d$, with rank equal to 3 and puri-rank$_{\mathbb{C}}\geq\Omega(\log d)$.
\end{lemma}
\begin{proof}
Slack matrices of regular $n$-gons in the plane have and rank 3 but puri-rank$_{\mathbb{C}}\geq\Omega(\log n)$ \cite{Gouveia:2013}.
\end{proof}

\begin{lemma}[${[3,2]}$]
\label{sup_lemma:matrix_families_prime_matrices}
There exists a family of non-negative matrices, of increasing dimension $d$, with TT-rank$_{\mathbb{R}_{\geq 0}}=2$ and Born-rank$_\mathbb{R}=d$.
\end{lemma}
\begin{proof}
Consider a sequence of integers $n_i$ such that $2n_i-1$ is the $i$-th prime. Define the primes matrices $K_{i,j} = n_i+n_j-1$. $K$ has rank 2 and non-negative rank 2. It was shown by induction in ref.~\cite{Fawzi2015} that the real square root rank of $K$ is full.
\end{proof}

\begin{lemma}[${[2,3]}$]
\label{sup_lemma:euclidian_matrices}
There exists a family of non-negative matrices, of increasing dimension $d$, with Born-rank$_\mathbb{R}=2$ and TT-rank$_{\mathbb{R}_{\geq 0}}\geq \log_2 d$.
\end{lemma}
\begin{proof}
Consider the linear Euclidean distance matrices defined as $M_{i,j} = (j-i)^2$. $M$ is the element-wise square of a matrix with elements equal to $i-j$, so has real square root rank equal to $2$. Moreover, 
it was shown in ref.~\cite{Fawzi2015} that $M$ has non-negative rank at least $\log_2 d$.
\end{proof}

\begin{lemma}[${[3,4]}$]
\label{sup_lemma:matrix_families_born_ranks}
There exists a family of non-negative matrices, with increasing dimension $d$, with Born-rank$_\mathbb{C}=2$ and Born-rank$_\mathbb{R}=d$.
\end{lemma}
\begin{proof}
The prime matrices $K$ introduced in Lemma~\ref{sup_lemma:matrix_families_prime_matrices} have full real square root rank. They can be written as the absolute value squared element-wise of a matrix $M_{i,j}=\sqrt{n_i}+i\sqrt{n_j-1}$. $M$ has rank 2, so the complex square root rank of $K$ is 2.
\end{proof}

Before continuing, note that all the remaining ``No'' entries not explicitly covered 
by Lemmas \ref{sup_lemma:matrix_families_ngons}-\ref{sup_lemma:matrix_families_born_ranks} in fact follow directly from these lemmas when combined with Proposition \ref{sup_prop:inclusions} (this is made explicit shortly, in Propositions \ref{sup_prop:asymptotics_1}-\ref{sup_prop:asymptotics_4}).

Given these families of probability distributions over two random variables, we now extend these results to the case of probability mass functions over many variables of small dimension. As discussed in the main text, for a particular $[\textit{row},\textit{column}]$ entry, the strategy is to start with a matrix $M$ of size $2^N\times 2^N$ such that $r_\text{column}$ is constant with respect to $N$, while $r_\text{row}$ grows. Via an ``unfolding'' technique, applied to the \textit{column} decomposition of $M$, we then show that there exists a non-negative tensor with $2N$ two-dimensional indices such that (a) the tensor rank corresponding to the column is equal to $r_{\text{column}}$ and (b) the matrix M is a reshaping of the central bipartition of the tensor. As a result of (b) it then follows that the tensor rank corresponding to the row  is lower bounded by $r_{\text{row}}$, therefore extending the separation from the case of two-variables to the case of many variables of small dimension.

\begin{align}
\label{eq:unfolding}
M=\begin{tikzpicture}[y=0.80pt, x=0.80pt, yscale=-1.000000, xscale=1.000000, inner sep=0pt, outer sep=0pt,baseline={([yshift=-14pt]current bounding box.center)}]
  \begin{scope}[rotate around={180.0:(272.71205,323.00155)},miter limit=4.00,line width=0.800pt]
    \path[draw=black,line join=miter,line cap=butt,miter limit=4.00,line
      width=0.800pt] (325.3029,282.7447) -- (350.7415,282.7447);
  \end{scope}
  \path[draw=black,line join=miter,line cap=butt,miter limit=4.00,line
    width=2.400pt] (197.0625,356.7363) -- (197.0625,344.9012);
  \path[scale=-1.000,draw=black,fill=cffffff,line join=miter,line cap=butt,miter
    limit=4.00,even odd rule,line width=0.800pt,rounded corners=0.0000cm]
    (-227.5104,-370.9657) rectangle (-212.4879,-356.3346);
  \path[scale=-1.000,draw=black,fill=cffffff,line join=miter,line cap=butt,miter
    limit=4.00,even odd rule,line width=0.800pt,rounded corners=0.0000cm]
    (-204.6419,-370.9657) rectangle (-189.6194,-356.3346);
  \path[draw=black,line join=miter,line cap=butt,miter limit=4.00,line
    width=2.400pt] (219.6839,356.6695) -- (219.6839,344.8343);
  \path[fill=black] (190.4093,339.1512) node[above right] (text1781-3) {$2^N$};
  \path[fill=black] (214.7439,339.5204) node[above right] (text1781-8-6) {$2^N$};
\end{tikzpicture}=\begin{tikzpicture}[y=0.80pt, x=0.80pt, yscale=-1.000000, xscale=1.000000, inner sep=0pt, outer sep=0pt,baseline={([yshift=-14pt]current bounding box.center)}]
  \begin{scope}[cm={{3.77953,0.0,0.0,3.77953,(-71.86333,163.57119)}}]
    \begin{scope}[cm={{-0.26458,0.0,0.0,-0.26458,(181.44801,127.56014)}},miter limit=4.00,line width=0.800pt]
      \path[draw=black,line join=miter,line cap=butt,miter limit=4.00,line
        width=0.800pt] (325.3029,282.7447) -- (350.7415,282.7447);
    \end{scope}
    \path[draw=black,line join=miter,line cap=butt,miter limit=4.00,line
      width=0.212pt] (89.2773,51.0250) -- (89.2773,47.8936);
    \path[scale=-1.000,draw=black,fill=cffffff,line join=miter,line cap=butt,miter
      limit=4.00,even odd rule,line width=0.212pt,rounded corners=0.0000cm]
      (-97.3334,-54.7898) rectangle (-93.3586,-50.9187);
    \path[scale=-1.000,draw=black,fill=cffffff,line join=miter,line cap=butt,miter
      limit=4.00,even odd rule,line width=0.212pt,rounded corners=0.0000cm]
      (-91.2827,-54.7898) rectangle (-87.3080,-50.9187);
    \path[draw=black,line join=miter,line cap=butt,miter limit=4.00,line
      width=0.212pt] (89.8417,51.0250) -- (89.8417,47.8936);
    \path[draw=black,line join=miter,line cap=butt,miter limit=4.00,line
      width=0.212pt] (88.7078,51.0250) -- (88.7078,47.8936);
    \path[draw=black,line join=miter,line cap=butt,miter limit=4.00,line
      width=0.212pt] (88.1644,51.0250) -- (88.1644,47.8936);
    \path[draw=black,line join=miter,line cap=butt,miter limit=4.00,line
      width=0.212pt] (90.3614,51.0250) -- (90.3614,47.8936);
    \path[draw=black,line join=miter,line cap=butt,miter limit=4.00,line
      width=0.212pt] (95.2626,51.0073) -- (95.2626,47.8759);
    \path[draw=black,line join=miter,line cap=butt,miter limit=4.00,line
      width=0.212pt] (95.8270,51.0073) -- (95.8270,47.8759);
    \path[draw=black,line join=miter,line cap=butt,miter limit=4.00,line
      width=0.212pt] (94.6930,51.0073) -- (94.6930,47.8759);
    \path[draw=black,line join=miter,line cap=butt,miter limit=4.00,line
      width=0.212pt] (94.1497,51.0073) -- (94.1497,47.8759);
    \path[draw=black,line join=miter,line cap=butt,miter limit=4.00,line
      width=0.212pt] (96.3467,51.0073) -- (96.3467,47.8759);
  \end{scope}
  \path[fill=black] (259.9811,339.0149) node[above right] (text1781) {$2^N$};
  \path[fill=black] (283.2443,339.2056) node[above right] (text1781-8) {$2^N$};
\end{tikzpicture}=\begin{tikzpicture}[y=0.80pt, x=0.80pt, yscale=-1.000000, xscale=1.000000, inner sep=0pt, outer sep=0pt,baseline={([yshift=-16pt]current bounding box.center)}]
    \path[draw=black,dash pattern=on 0.80pt off 1.60pt,line join=miter,line
      cap=butt,miter limit=4.00,line width=0.800pt] (420.7248,334.8708) --
      (420.7248,380.2279);
    \begin{scope}[cm={{3.77953,0.0,0.0,3.77953,(112.51054,96.63679)}}]
      \path[draw=black,line join=miter,line cap=butt,miter limit=4.00,line
        width=0.212pt] (71.8546,70.8087) -- (65.0572,70.8087);
      \path[scale=-1.000,draw=black,fill=cffffff,line join=miter,line cap=butt,miter
        limit=4.00,even odd rule,line width=0.212pt,rounded corners=0.0000cm]
        (-70.4742,-72.7854) rectangle (-66.4995,-68.9142);
      \path[draw=black,line join=miter,line cap=butt,miter limit=4.00,line
        width=0.212pt] (68.4051,69.0001) -- (68.4051,65.8687);
      \path[draw=black,line join=miter,line cap=butt,miter limit=4.00,line
        width=0.212pt] (83.0290,70.8064) -- (74.7914,70.8064);
      \path[xscale=-1.000,yscale=1.000,fill=black,miter limit=4.00,line width=0.212pt]
        (-73.8823,70.8048) circle (0.0075cm);
      \path[xscale=-1.000,yscale=1.000,fill=black,miter limit=4.00,line width=0.212pt]
        (-72.7156,70.8048) circle (0.0075cm);
      \path[scale=-1.000,draw=black,fill=cffffff,line join=miter,line cap=butt,miter
        limit=4.00,even odd rule,line width=0.212pt,rounded corners=0.0000cm]
        (-80.0694,-72.8456) rectangle (-76.0947,-68.9744);
      \path[draw=black,line join=miter,line cap=butt,miter limit=4.00,line
        width=0.212pt] (78.0003,69.0603) -- (78.0003,65.9289);
      \path[draw=black,line join=miter,line cap=butt,miter limit=4.00,line
        width=0.212pt] (62.1198,70.8044) -- (57.5524,70.8044);
      \path[scale=-1.000,draw=black,fill=cffffff,line join=miter,line cap=butt,miter
        limit=4.00,even odd rule,line width=0.212pt,rounded corners=0.0000cm]
        (-60.7393,-72.7811) rectangle (-56.7646,-68.9099);
      \path[draw=black,line join=miter,line cap=butt,miter limit=4.00,line
        width=0.212pt] (58.7120,68.9708) -- (58.7120,65.8394);
      \path[xscale=-1.000,yscale=1.000,fill=black,miter limit=4.00,line width=0.212pt]
        (-64.1892,70.7755) circle (0.0075cm);
      \path[xscale=-1.000,yscale=1.000,fill=black,miter limit=4.00,line width=0.212pt]
        (-63.0225,70.7755) circle (0.0075cm);
    \end{scope}
    \begin{scope}[cm={{-3.77953,0.0,0.0,3.77953,(728.88602,96.66401)}}]
      \path[draw=black,line join=miter,line cap=butt,miter limit=4.00,line
        width=0.212pt] (71.8546,70.8087) -- (65.0572,70.8087);
      \path[scale=-1.000,draw=black,fill=cffffff,line join=miter,line cap=butt,miter
        limit=4.00,even odd rule,line width=0.212pt,rounded corners=0.0000cm]
        (-70.4742,-72.7854) rectangle (-66.4995,-68.9142);
      \path[draw=black,line join=miter,line cap=butt,miter limit=4.00,line
        width=0.212pt] (68.4051,69.0001) -- (68.4051,65.8687);
      \path[draw=black,line join=miter,line cap=butt,miter limit=4.00,line
        width=0.212pt] (83.0290,70.8064) -- (74.7914,70.8064);
      \path[xscale=-1.000,yscale=1.000,fill=black,miter limit=4.00,line width=0.212pt]
        (-73.8823,70.8048) circle (0.0075cm);
      \path[xscale=-1.000,yscale=1.000,fill=black,miter limit=4.00,line width=0.212pt]
        (-72.7156,70.8048) circle (0.0075cm);
      \path[scale=-1.000,draw=black,fill=cffffff,line join=miter,line cap=butt,miter
        limit=4.00,even odd rule,line width=0.212pt,rounded corners=0.0000cm]
        (-80.0694,-72.8456) rectangle (-76.0947,-68.9744);
      \path[draw=black,line join=miter,line cap=butt,miter limit=4.00,line
        width=0.212pt] (78.0003,69.0603) -- (78.0003,65.9289);
      \path[draw=black,line join=miter,line cap=butt,miter limit=4.00,line
        width=0.212pt] (62.1198,70.8044) -- (57.5524,70.8044);
      \path[scale=-1.000,draw=black,fill=cffffff,line join=miter,line cap=butt,miter
        limit=4.00,even odd rule,line width=0.212pt,rounded corners=0.0000cm]
        (-60.7393,-72.7811) rectangle (-56.7646,-68.9099);
      \path[draw=black,line join=miter,line cap=butt,miter limit=4.00,line
        width=0.212pt] (58.7120,68.9708) -- (58.7120,65.8394);
      \path[xscale=-1.000,yscale=1.000,fill=black,miter limit=4.00,line width=0.212pt]
        (-64.1892,70.7755) circle (0.0075cm);
      \path[xscale=-1.000,yscale=1.000,fill=black,miter limit=4.00,line width=0.212pt]
        (-63.0225,70.7755) circle (0.0075cm);
    \end{scope}
    \path[draw=black,line join=bevel,line cap=butt,miter limit=4.00,line
      width=0.801pt] (327.7993,342.0915) .. controls (327.7993,327.6698) and
      (367.0690,341.9653) .. (371.4884,328.2019) .. controls (375.0239,341.9653) and
      (414.0411,327.8093) .. (414.0411,342.2178);
    \path[draw=black,line join=bevel,line cap=butt,miter limit=4.00,line
      width=0.801pt] (426.8574,342.1681) .. controls (426.8574,327.7463) and
      (466.1271,342.0418) .. (470.5465,328.2785) .. controls (474.0820,342.0418) and
      (513.0992,327.8859) .. (513.0992,342.2944);
    \path[fill=black,line width=3.024pt] (367.8382,322.5863) node[above right]
      (text1880) {$N$};
    \path[fill=black,line width=3.024pt] (466.8727,323.9538) node[above right]
      (text1880-6) {$N$};
\end{tikzpicture}
\end{align}
More generally, we can write any $N\times N$ matrix $M$ as a submatrix of a $2^N\times 2^N$ matrix for which we can apply the previous idea. In this case $M$ is a submatrix of the central bipartition of the obtained tensor over $2N$ binary variables,
\begin{align}
M=\begin{tikzpicture}[y=0.80pt, x=0.80pt, yscale=-1.000000, xscale=1.000000, inner sep=0pt, outer sep=0pt,baseline={([yshift=-14pt]current bounding box.center)}]
  \begin{scope}[rotate around={180.0:(272.71205,323.00155)},miter limit=4.00,line width=0.800pt]
    \path[draw=black,line join=miter,line cap=butt,miter limit=4.00,line
      width=0.800pt] (325.3029,282.7447) -- (350.7415,282.7447);
  \end{scope}
  \path[draw=black,line join=miter,line cap=butt,miter limit=4.00,line
    width=2.400pt] (197.0625,356.7363) -- (197.0625,344.9012);
  \path[scale=-1.000,draw=black,fill=cffffff,line join=miter,line cap=butt,miter
    limit=4.00,even odd rule,line width=0.800pt,rounded corners=0.0000cm]
    (-227.5104,-370.9657) rectangle (-212.4879,-356.3346);
  \path[scale=-1.000,draw=black,fill=cffffff,line join=miter,line cap=butt,miter
    limit=4.00,even odd rule,line width=0.800pt,rounded corners=0.0000cm]
    (-204.6419,-370.9657) rectangle (-189.6194,-356.3346);
  \path[draw=black,line join=miter,line cap=butt,miter limit=4.00,line
    width=2.400pt] (219.6839,356.6695) -- (219.6839,344.8343);
  \path[fill=black] (190.4093,339.1512) node[above right] (text1781-3) {$N$};
  \path[fill=black] (214.7439,339.5204) node[above right] (text1781-8-6) {$N$};
\end{tikzpicture}\subset R=\begin{tikzpicture}[y=0.80pt, x=0.80pt, yscale=-1.000000, xscale=1.000000, inner sep=0pt, outer sep=0pt,baseline={([yshift=-14pt]current bounding box.center)}]
  \begin{scope}[rotate around={180.0:(272.71205,323.00155)},miter limit=4.00,line width=0.800pt]
    \path[draw=black,line join=miter,line cap=butt,miter limit=4.00,line
      width=0.800pt] (325.3029,282.7447) -- (350.7415,282.7447);
  \end{scope}
  \path[draw=black,line join=miter,line cap=butt,miter limit=4.00,line
    width=2.400pt] (197.0625,356.7363) -- (197.0625,344.9012);
  \path[scale=-1.000,draw=black,fill=cffffff,line join=miter,line cap=butt,miter
    limit=4.00,even odd rule,line width=0.800pt,rounded corners=0.0000cm]
    (-227.5104,-370.9657) rectangle (-212.4879,-356.3346);
  \path[scale=-1.000,draw=black,fill=cffffff,line join=miter,line cap=butt,miter
    limit=4.00,even odd rule,line width=0.800pt,rounded corners=0.0000cm]
    (-204.6419,-370.9657) rectangle (-189.6194,-356.3346);
  \path[draw=black,line join=miter,line cap=butt,miter limit=4.00,line
    width=2.400pt] (219.6839,356.6695) -- (219.6839,344.8343);
  \path[fill=black] (190.4093,339.1512) node[above right] (text1781-3) {$2^N$};
  \path[fill=black] (214.7439,339.5204) node[above right] (text1781-8-6) {$2^N$};
\end{tikzpicture}=\begin{tikzpicture}[y=0.80pt, x=0.80pt, yscale=-1.000000, xscale=1.000000, inner sep=0pt, outer sep=0pt,baseline={([yshift=-14pt]current bounding box.center)}]
  \begin{scope}[cm={{3.77953,0.0,0.0,3.77953,(-71.86333,163.57119)}}]
    \begin{scope}[cm={{-0.26458,0.0,0.0,-0.26458,(181.44801,127.56014)}},miter limit=4.00,line width=0.800pt]
      \path[draw=black,line join=miter,line cap=butt,miter limit=4.00,line
        width=0.800pt] (325.3029,282.7447) -- (350.7415,282.7447);
    \end{scope}
    \path[draw=black,line join=miter,line cap=butt,miter limit=4.00,line
      width=0.212pt] (89.2773,51.0250) -- (89.2773,47.8936);
    \path[scale=-1.000,draw=black,fill=cffffff,line join=miter,line cap=butt,miter
      limit=4.00,even odd rule,line width=0.212pt,rounded corners=0.0000cm]
      (-97.3334,-54.7898) rectangle (-93.3586,-50.9187);
    \path[scale=-1.000,draw=black,fill=cffffff,line join=miter,line cap=butt,miter
      limit=4.00,even odd rule,line width=0.212pt,rounded corners=0.0000cm]
      (-91.2827,-54.7898) rectangle (-87.3080,-50.9187);
    \path[draw=black,line join=miter,line cap=butt,miter limit=4.00,line
      width=0.212pt] (89.8417,51.0250) -- (89.8417,47.8936);
    \path[draw=black,line join=miter,line cap=butt,miter limit=4.00,line
      width=0.212pt] (88.7078,51.0250) -- (88.7078,47.8936);
    \path[draw=black,line join=miter,line cap=butt,miter limit=4.00,line
      width=0.212pt] (88.1644,51.0250) -- (88.1644,47.8936);
    \path[draw=black,line join=miter,line cap=butt,miter limit=4.00,line
      width=0.212pt] (90.3614,51.0250) -- (90.3614,47.8936);
    \path[draw=black,line join=miter,line cap=butt,miter limit=4.00,line
      width=0.212pt] (95.2626,51.0073) -- (95.2626,47.8759);
    \path[draw=black,line join=miter,line cap=butt,miter limit=4.00,line
      width=0.212pt] (95.8270,51.0073) -- (95.8270,47.8759);
    \path[draw=black,line join=miter,line cap=butt,miter limit=4.00,line
      width=0.212pt] (94.6930,51.0073) -- (94.6930,47.8759);
    \path[draw=black,line join=miter,line cap=butt,miter limit=4.00,line
      width=0.212pt] (94.1497,51.0073) -- (94.1497,47.8759);
    \path[draw=black,line join=miter,line cap=butt,miter limit=4.00,line
      width=0.212pt] (96.3467,51.0073) -- (96.3467,47.8759);
  \end{scope}
  \path[fill=black] (259.9811,339.0149) node[above right] (text1781) {$2^N$};
  \path[fill=black] (283.2443,339.2056) node[above right] (text1781-8) {$2^N$};
\end{tikzpicture}=\begin{tikzpicture}[y=0.80pt, x=0.80pt, yscale=-1.000000, xscale=1.000000, inner sep=0pt, outer sep=0pt,baseline={([yshift=-16pt]current bounding box.center)}]
    \path[draw=black,dash pattern=on 0.80pt off 1.60pt,line join=miter,line
      cap=butt,miter limit=4.00,line width=0.800pt] (420.7248,334.8708) --
      (420.7248,380.2279);
    \begin{scope}[cm={{3.77953,0.0,0.0,3.77953,(112.51054,96.63679)}}]
      \path[draw=black,line join=miter,line cap=butt,miter limit=4.00,line
        width=0.212pt] (71.8546,70.8087) -- (65.0572,70.8087);
      \path[scale=-1.000,draw=black,fill=cffffff,line join=miter,line cap=butt,miter
        limit=4.00,even odd rule,line width=0.212pt,rounded corners=0.0000cm]
        (-70.4742,-72.7854) rectangle (-66.4995,-68.9142);
      \path[draw=black,line join=miter,line cap=butt,miter limit=4.00,line
        width=0.212pt] (68.4051,69.0001) -- (68.4051,65.8687);
      \path[draw=black,line join=miter,line cap=butt,miter limit=4.00,line
        width=0.212pt] (83.0290,70.8064) -- (74.7914,70.8064);
      \path[xscale=-1.000,yscale=1.000,fill=black,miter limit=4.00,line width=0.212pt]
        (-73.8823,70.8048) circle (0.0075cm);
      \path[xscale=-1.000,yscale=1.000,fill=black,miter limit=4.00,line width=0.212pt]
        (-72.7156,70.8048) circle (0.0075cm);
      \path[scale=-1.000,draw=black,fill=cffffff,line join=miter,line cap=butt,miter
        limit=4.00,even odd rule,line width=0.212pt,rounded corners=0.0000cm]
        (-80.0694,-72.8456) rectangle (-76.0947,-68.9744);
      \path[draw=black,line join=miter,line cap=butt,miter limit=4.00,line
        width=0.212pt] (78.0003,69.0603) -- (78.0003,65.9289);
      \path[draw=black,line join=miter,line cap=butt,miter limit=4.00,line
        width=0.212pt] (62.1198,70.8044) -- (57.5524,70.8044);
      \path[scale=-1.000,draw=black,fill=cffffff,line join=miter,line cap=butt,miter
        limit=4.00,even odd rule,line width=0.212pt,rounded corners=0.0000cm]
        (-60.7393,-72.7811) rectangle (-56.7646,-68.9099);
      \path[draw=black,line join=miter,line cap=butt,miter limit=4.00,line
        width=0.212pt] (58.7120,68.9708) -- (58.7120,65.8394);
      \path[xscale=-1.000,yscale=1.000,fill=black,miter limit=4.00,line width=0.212pt]
        (-64.1892,70.7755) circle (0.0075cm);
      \path[xscale=-1.000,yscale=1.000,fill=black,miter limit=4.00,line width=0.212pt]
        (-63.0225,70.7755) circle (0.0075cm);
    \end{scope}
    \begin{scope}[cm={{-3.77953,0.0,0.0,3.77953,(728.88602,96.66401)}}]
      \path[draw=black,line join=miter,line cap=butt,miter limit=4.00,line
        width=0.212pt] (71.8546,70.8087) -- (65.0572,70.8087);
      \path[scale=-1.000,draw=black,fill=cffffff,line join=miter,line cap=butt,miter
        limit=4.00,even odd rule,line width=0.212pt,rounded corners=0.0000cm]
        (-70.4742,-72.7854) rectangle (-66.4995,-68.9142);
      \path[draw=black,line join=miter,line cap=butt,miter limit=4.00,line
        width=0.212pt] (68.4051,69.0001) -- (68.4051,65.8687);
      \path[draw=black,line join=miter,line cap=butt,miter limit=4.00,line
        width=0.212pt] (83.0290,70.8064) -- (74.7914,70.8064);
      \path[xscale=-1.000,yscale=1.000,fill=black,miter limit=4.00,line width=0.212pt]
        (-73.8823,70.8048) circle (0.0075cm);
      \path[xscale=-1.000,yscale=1.000,fill=black,miter limit=4.00,line width=0.212pt]
        (-72.7156,70.8048) circle (0.0075cm);
      \path[scale=-1.000,draw=black,fill=cffffff,line join=miter,line cap=butt,miter
        limit=4.00,even odd rule,line width=0.212pt,rounded corners=0.0000cm]
        (-80.0694,-72.8456) rectangle (-76.0947,-68.9744);
      \path[draw=black,line join=miter,line cap=butt,miter limit=4.00,line
        width=0.212pt] (78.0003,69.0603) -- (78.0003,65.9289);
      \path[draw=black,line join=miter,line cap=butt,miter limit=4.00,line
        width=0.212pt] (62.1198,70.8044) -- (57.5524,70.8044);
      \path[scale=-1.000,draw=black,fill=cffffff,line join=miter,line cap=butt,miter
        limit=4.00,even odd rule,line width=0.212pt,rounded corners=0.0000cm]
        (-60.7393,-72.7811) rectangle (-56.7646,-68.9099);
      \path[draw=black,line join=miter,line cap=butt,miter limit=4.00,line
        width=0.212pt] (58.7120,68.9708) -- (58.7120,65.8394);
      \path[xscale=-1.000,yscale=1.000,fill=black,miter limit=4.00,line width=0.212pt]
        (-64.1892,70.7755) circle (0.0075cm);
      \path[xscale=-1.000,yscale=1.000,fill=black,miter limit=4.00,line width=0.212pt]
        (-63.0225,70.7755) circle (0.0075cm);
    \end{scope}
    \path[draw=black,line join=bevel,line cap=butt,miter limit=4.00,line
      width=0.801pt] (327.7993,342.0915) .. controls (327.7993,327.6698) and
      (367.0690,341.9653) .. (371.4884,328.2019) .. controls (375.0239,341.9653) and
      (414.0411,327.8093) .. (414.0411,342.2178);
    \path[draw=black,line join=bevel,line cap=butt,miter limit=4.00,line
      width=0.801pt] (426.8574,342.1681) .. controls (426.8574,327.7463) and
      (466.1271,342.0418) .. (470.5465,328.2785) .. controls (474.0820,342.0418) and
      (513.0992,327.8859) .. (513.0992,342.2944);
    \path[fill=black,line width=3.024pt] (367.8382,322.5863) node[above right]
      (text1880) {$N$};
    \path[fill=black,line width=3.024pt] (466.8727,323.9538) node[above right]
      (text1880-6) {$N$};
\end{tikzpicture} .
\end{align}
The ``unfolding'' technique upon which this proof strategy relies is formalized by Lemma~\ref{sup_lemma:embedding_and_unfolding}:

\begin{lemma}\label{sup_lemma:embedding_and_unfolding}
\label{lemma:matrixtomps}
Consider a non-negative matrix $M$ of TT-rank$_{\mathbb{R}_{\geq 0}}$ (resp. TT-rank$_{\mathbb{R}}$, Born-rank$_{\mathbb{R}}$, Born-rank$_{\mathbb{C}}$) equal to $r$ and size $N\times N$. Then there exists an MPS$_{\mathbb{R}_{\geq 0}}$ (resp. MPS$_{\mathbb{R}}$, BM$_{\mathbb{R}}$, BM$_{\mathbb{C}}$) over $2N$ binary variables with TT-rank$_{\mathbb{R}_{\geq 0}}$ (resp. TT-rank$_{\mathbb{R}}$, Born-rank$_{\mathbb{R}}$, Born-rank$_{\mathbb{C}}$) equal to $r$ such that $M$ is a submatrix of the central bipartition of the resulting tensor. 
\end{lemma}
\begin{proof}
Let us first prove the case where $M$ has TT-rank$_{\mathbb{R}}$ $r$. In this case we can write 
\begin{align}
    M_{i,j}=\sum_{\alpha=1}^r E_{i,\alpha} F_{\alpha, j},
\end{align}
where $E$ and $F$ are real matrices.
 
We now define the appropriate MPS of TT-rank $r$ by direct specification of its tensors. Specifically, define the boundary tensors for site one ($2N$) such that the row (column) vector in the zero-index is a vector of ones and the row (column) vector in the one-index is the first row (last column) of $E$ ($F$), i.e.,
\begin{align}
   & A_{1,0}^{\alpha}=1,\ \  A_{1,1}^{\alpha}=E_{1 ,\alpha} ,\\
   &A_{2N,0}^{\alpha}=1,\ \ A_{2N,1}^{\alpha}=F_{\alpha, N}.
\end{align}
We then define the bulk tensors such that left (right) of the central bipartition the matrix in the zero-index is the identity matrix, while the matrix in the one-index is diagonal with a row (column) of $E$ ($F$) on the diagonal, i.e., for all $n$ in $\{2,\ldots,N\}$,
\begin{align}
    &A_{n,0}^{\alpha,\beta}=\delta_{\alpha,\beta},\ \ A_{n,1}^{\alpha,\beta}=\delta_{\alpha,\beta} E_{n ,\alpha},\\
    &A_{n-1 + N,0}^{\alpha,\beta}=\delta_{\alpha,\beta},\ \ A_{n-1 + N,1}^{\alpha,\beta}=\delta_{\alpha,\beta} F_{\alpha ,n} .
\end{align}
This MPS defines a tensor over $2N$ variables. 
\begin{align}
    \begin{tikzpicture}[y=0.8pt, x=0.8pt,yscale=-1.0, xscale=1.0,  baseline=(current bounding box.center)]
\path[draw=black,dash pattern=on 0.80pt off 1.60pt,line join=miter,line
  cap=butt,miter limit=4.00,line width=0.800pt] (441.4580,367.3041) --
  (441.4580,412.6613);
\begin{scope}[cm={{3.77953,0.0,0.0,3.77953,(-47.91589,197.07599)}}]
  \begin{scope}[cm={{-0.26458,0.0,0.0,-0.26458,(181.44801,127.56014)}},miter limit=4.00,line width=0.800pt]
    \path[draw=black,line join=miter,line cap=butt,miter limit=4.00,line
      width=0.800pt] (325.3029,282.7447) -- (350.7415,282.7447);
  \end{scope}
  \path[draw=black,line join=miter,line cap=butt,miter limit=4.00,line
    width=0.212pt] (89.2773,51.0250) -- (89.2773,47.8936);
  \path[scale=-1.000,draw=black,fill=cffffff,line join=miter,line cap=butt,miter
    limit=4.00,even odd rule,line width=0.212pt,rounded corners=0.0000cm]
    (-97.3334,-54.7898) rectangle (-93.3586,-50.9187);
  \path[scale=-1.000,draw=black,fill=cffffff,line join=miter,line cap=butt,miter
    limit=4.00,even odd rule,line width=0.212pt,rounded corners=0.0000cm]
    (-91.2827,-54.7898) rectangle (-87.3080,-50.9187);
  \path[draw=black,line join=miter,line cap=butt,miter limit=4.00,line
    width=0.212pt] (89.8417,51.0250) -- (89.8417,47.8936);
  \path[draw=black,line join=miter,line cap=butt,miter limit=4.00,line
    width=0.212pt] (88.7078,51.0250) -- (88.7078,47.8936);
  \path[draw=black,line join=miter,line cap=butt,miter limit=4.00,line
    width=0.212pt] (88.1644,51.0250) -- (88.1644,47.8936);
  \path[draw=black,line join=miter,line cap=butt,miter limit=4.00,line
    width=0.212pt] (90.3614,51.0250) -- (90.3614,47.8936);
  \path[draw=black,line join=miter,line cap=butt,miter limit=4.00,line
    width=0.212pt] (95.2626,51.0073) -- (95.2626,47.8759);
  \path[draw=black,line join=miter,line cap=butt,miter limit=4.00,line
    width=0.212pt] (95.8270,51.0073) -- (95.8270,47.8759);
  \path[draw=black,line join=miter,line cap=butt,miter limit=4.00,line
    width=0.212pt] (94.6930,51.0073) -- (94.6930,47.8759);
  \path[draw=black,line join=miter,line cap=butt,miter limit=4.00,line
    width=0.212pt] (94.1497,51.0073) -- (94.1497,47.8759);
  \path[draw=black,line join=miter,line cap=butt,miter limit=4.00,line
    width=0.212pt] (96.3467,51.0073) -- (96.3467,47.8759);
\end{scope}
\begin{scope}[cm={{3.77953,0.0,0.0,3.77953,(133.24369,129.07016)}}]
  \path[draw=black,line join=miter,line cap=butt,miter limit=4.00,line
    width=0.212pt] (71.8546,70.8087) -- (65.0572,70.8087);
  \path[scale=-1.000,draw=black,fill=cffffff,line join=miter,line cap=butt,miter
    limit=4.00,even odd rule,line width=0.212pt,rounded corners=0.0000cm]
    (-70.4742,-72.7854) rectangle (-66.4995,-68.9142);
  \path[draw=black,line join=miter,line cap=butt,miter limit=4.00,line
    width=0.212pt] (68.4051,69.0001) -- (68.4051,65.8687);
  \path[draw=black,line join=miter,line cap=butt,miter limit=4.00,line
    width=0.212pt] (83.0290,70.8064) -- (74.7914,70.8064);
  \path[xscale=-1.000,yscale=1.000,fill=black,miter limit=4.00,line width=0.212pt]
    (-73.8823,70.8048) circle (0.0075cm);
  \path[xscale=-1.000,yscale=1.000,fill=black,miter limit=4.00,line width=0.212pt]
    (-72.7156,70.8048) circle (0.0075cm);
  \path[scale=-1.000,draw=black,fill=cffffff,line join=miter,line cap=butt,miter
    limit=4.00,even odd rule,line width=0.212pt,rounded corners=0.0000cm]
    (-80.0694,-72.8456) rectangle (-76.0947,-68.9744);
  \path[draw=black,line join=miter,line cap=butt,miter limit=4.00,line
    width=0.212pt] (78.0003,69.0603) -- (78.0003,65.9289);
  \path[draw=black,line join=miter,line cap=butt,miter limit=4.00,line
    width=0.212pt] (62.1198,70.8044) -- (57.5524,70.8044);
  \path[scale=-1.000,draw=black,fill=cffffff,line join=miter,line cap=butt,miter
    limit=4.00,even odd rule,line width=0.212pt,rounded corners=0.0000cm]
    (-60.7393,-72.7811) rectangle (-56.7646,-68.9099);
  \path[draw=black,line join=miter,line cap=butt,miter limit=4.00,line
    width=0.212pt] (58.7120,68.9708) -- (58.7120,65.8394);
  \path[xscale=-1.000,yscale=1.000,fill=black,miter limit=4.00,line width=0.212pt]
    (-64.1892,70.7755) circle (0.0075cm);
  \path[xscale=-1.000,yscale=1.000,fill=black,miter limit=4.00,line width=0.212pt]
    (-63.0225,70.7755) circle (0.0075cm);
\end{scope}
\begin{scope}[cm={{-3.77953,0.0,0.0,3.77953,(749.61917,129.09738)}}]
  \path[draw=black,line join=miter,line cap=butt,miter limit=4.00,line
    width=0.212pt] (71.8546,70.8087) -- (65.0572,70.8087);
  \path[scale=-1.000,draw=black,fill=cffffff,line join=miter,line cap=butt,miter
    limit=4.00,even odd rule,line width=0.212pt,rounded corners=0.0000cm]
    (-70.4742,-72.7854) rectangle (-66.4995,-68.9142);
  \path[draw=black,line join=miter,line cap=butt,miter limit=4.00,line
    width=0.212pt] (68.4051,69.0001) -- (68.4051,65.8687);
  \path[draw=black,line join=miter,line cap=butt,miter limit=4.00,line
    width=0.212pt] (83.0290,70.8064) -- (74.7914,70.8064);
  \path[xscale=-1.000,yscale=1.000,fill=black,miter limit=4.00,line width=0.212pt]
    (-73.8823,70.8048) circle (0.0075cm);
  \path[xscale=-1.000,yscale=1.000,fill=black,miter limit=4.00,line width=0.212pt]
    (-72.7156,70.8048) circle (0.0075cm);
  \path[scale=-1.000,draw=black,fill=cffffff,line join=miter,line cap=butt,miter
    limit=4.00,even odd rule,line width=0.212pt,rounded corners=0.0000cm]
    (-80.0694,-72.8456) rectangle (-76.0947,-68.9744);
  \path[draw=black,line join=miter,line cap=butt,miter limit=4.00,line
    width=0.212pt] (78.0003,69.0603) -- (78.0003,65.9289);
  \path[draw=black,line join=miter,line cap=butt,miter limit=4.00,line
    width=0.212pt] (62.1198,70.8044) -- (57.5524,70.8044);
  \path[scale=-1.000,draw=black,fill=cffffff,line join=miter,line cap=butt,miter
    limit=4.00,even odd rule,line width=0.212pt,rounded corners=0.0000cm]
    (-60.7393,-72.7811) rectangle (-56.7646,-68.9099);
  \path[draw=black,line join=miter,line cap=butt,miter limit=4.00,line
    width=0.212pt] (58.7120,68.9708) -- (58.7120,65.8394);
  \path[xscale=-1.000,yscale=1.000,fill=black,miter limit=4.00,line width=0.212pt]
    (-64.1892,70.7755) circle (0.0075cm);
  \path[xscale=-1.000,yscale=1.000,fill=black,miter limit=4.00,line width=0.212pt]
    (-63.0225,70.7755) circle (0.0075cm);
\end{scope}
\path[draw=black,line join=bevel,line cap=butt,miter limit=4.00,line
  width=0.801pt] (348.5324,374.5249) .. controls (348.5324,360.1031) and
  (387.8021,374.3986) .. (392.2215,360.6353) .. controls (395.7571,374.3986) and
  (434.7742,360.2427) .. (434.7742,374.6512);
\path[draw=black,line join=bevel,line cap=butt,miter limit=4.00,line
  width=0.801pt] (447.5905,374.6015) .. controls (447.5905,360.1797) and
  (486.8602,374.4752) .. (491.2796,360.7119) .. controls (494.8152,374.4752) and
  (533.8323,360.3192) .. (533.8323,374.7277);
\node[] at (391,354) {$N$};
\node[] at (491,354) {$N$};
\node[] at (391,414) {$i$};
\node[] at (491,414) {$j$};
\node[] at (300,384) {$r$};
\node[] at (333,398) {$=$};
\node[] at (290,372) {${\scriptstyle i}$};
\node[] at (310,372) {${\scriptstyle j}$};
\node[] at (354.8,374) {${\scriptstyle 0}$};
\node[] at (391,374) {${\scriptstyle 1}$};
\node[] at (428,374) {${\scriptstyle 0}$};
\node[] at (454.5,374) {${\scriptstyle 0}$};
\node[] at (490.8,374) {${\scriptstyle 1}$};
\node[] at (527,374) {${\scriptstyle 0}$};
\end{tikzpicture}
\end{align}
Define $R$ the $2^N \times 2^N$ matrix corresponding to a reshaping as a matrix of this tensor across the central bipartition. Consider $R_{0\cdots 010 \cdots 0,0\cdots 010 \cdots 0}$, where the variables are all 0 except a 1 in position $i\leq d$ and a 1 in position $j\geq d+1$, then $R_{0\cdots 010 \cdots 0,0\cdots 010 \cdots 0}=\sum_{\alpha=1}^r E_{i,\alpha}F_{\alpha, j}=M_{i,j}$. Therefore, $M$ is a submatrix of $R$, up to a reshaping of $R$ as a matrix. The exact same proof can be done if $M$ has TT-rank$_{\mathbb{R}_{\geq 0}}=r$.

If $M$ has Born-rank$_{\mathbb{R}}$ (resp.~Born-rank$_{\mathbb{C}}$) $r$, apply the previous result to a real (resp.~complex) element-wise square root of M of rank $r$. This leads to an MPS for which the square root of $M$ is a submatrix of the central bipartition of the MPS. Therefore, $M$ is a submatrix of the central bipartition of the tensor obtained from the corresponding BM, which is the square of this MPS.
\end{proof}

Via the strategy discussed above - i.e., applying the unfolding to the matrix examples in Lemmas \ref{sup_lemma:matrix_families_ngons} -\ref{sup_lemma:matrix_families_born_ranks} (with respect to the decomposition for which the corresponding rank remains constant) - we are able to use Lemma~\ref{sup_lemma:embedding_and_unfolding}  to leverage the matrix families from Lemmas~\ref{sup_lemma:matrix_families_ngons}-\ref{sup_lemma:matrix_families_born_ranks} into families of probability distributions over $N$ random variables of dimension 2, which prove all the ``No'' entries in Table~\ref{sup_table:ranks} (when combined with Proposition 1). Note that the entries ``No$^*$'' remain conjectures. The existence of a family of matrices of constant non-negative rank but unbounded complex Hadamard square root rank, together with Lemma~\ref{sup_lemma:embedding_and_unfolding}, would prove these entries. 

While Lemma~\ref{sup_lemma:embedding_and_unfolding} provides an explicit construction for the ``No'' entries in Table~\ref{sup_table:ranks}, the separations it provides are not optimal, since it is sometimes possible to unfold a $2^N \times 2^N$ matrix into a tensor network of only $2N$ variables, as in Equation~\eqref{eq:unfolding}, rather than into a tensor network of $2(2^N)$ variables as is done in Lemma~\ref{sup_lemma:embedding_and_unfolding}. For this reason we provide more detailed proofs for the explicit asymptotics of the relevant separations, all of which use a similar strategy, but some of which use alternative unfolding techniques. These results are stated here as Propositions \ref{sup_prop:asymptotics_1}-\ref{sup_prop:asymptotics_4} to reflect their discussion in the main text. In particular, in order to obtain the asymptotic separations given in Propositions \ref{sup_prop:asymptotics_1}-\ref{sup_prop:asymptotics_3}, it is necessary to use alternative unfolding techniques to the one presented in Lemma~\ref{sup_lemma:embedding_and_unfolding}. Once again, the propositions are labelled by the the specific cases of Table \ref{sup_table:ranks} which they address.

\begin{proposition}[${[6,1]}$, ${[5,1]}$, ${[4,1]}$, ${[3,1]}$, ${[2,1]}$]\label{sup_prop:asymptotics_1}
There exists a family of non-negative tensors over $2N$ binary variables with constant TT-rank$_\mathbb{R}$=3 but with puri-rank$_{\mathbb{C}}=\Omega(N)$, and hence also puri-rank$_{\mathbb{C}}$, Born-rank$_{\mathbb{R}/\mathbb{C}}$ and TT-rank$_{\mathbb{R}_{\geq 0}} \geq \Omega(N)$.
\end{proposition}
\begin{proof}
The result for the case $[6,1]$ has already been proven in
ref.~\cite{Cuevas_2013}. Note that the remaining cases then follow from Proposition~\ref{sup_prop:inclusions}.
\end{proof}

\begin{proposition}[${[3,2]}$, ${[3,5]}$, ${[3,6]}$] \label{sup_prop:asymptotics_2}
There exists a family of non-negative tensors over $2N$ binary variables with constant TT-rank$_{\mathbb{R}_{\geq 0}}=2$ (and hence also puri-rank$_{\mathbb{R}/\mathbb{C}}=2$) but with Born-rank$_{\mathbb{R}} \geq \pi(2^{N+1})$, where $\pi(x)$ is the number of prime numbers up to $x$, which asymptotically satisfies $\pi(x)\sim x/\log(x)$.
\end{proposition}
\begin{proof}
Consider the $2^N\times 2^N$ matrix with entries 
$M_{i,j}=i+j$. A submatrix of $M$ is the prime matrix $K$ defined in Lemma~\ref{sup_lemma:matrix_families_prime_matrices} of size $\pi(2^{N+1})$, where $\pi(x)$ is the number of prime numbers lower than $x$. Let us show that we can define an MPS$_{\mathbb{R}_{\geq 0}}$ of TT-rank$_{\mathbb{R}_{\geq 0}}=2$ such that $M$ is the central bipartition of the resulting tensor. Let us first define the matrices 
\begin{align}
P_N=\left(\begin{matrix}
1& 1\\
1& 2\\
\vdots& \vdots \\
1& 2^N
\end{matrix}\right),
Q_N=\left(\begin{matrix}
1& 2& \cdots & 2^N\\
1& 1& \cdots & 1
\end{matrix}\right),
\end{align}
and observe that $M=P_N Q_N$,
\begin{align}
M=\begin{tikzpicture}[y=0.80pt, x=0.80pt, yscale=-1.3000000, xscale=1.3000000, inner sep=0pt, outer sep=0pt,baseline={([yshift=-12pt]current bounding box.center)}]
\path[draw=black,line join=miter,line cap=butt,miter limit=4.00,line
  width=0.800pt] (311.0620,324.3042) -- (283.2237,324.3042);
\path[draw=black,line join=miter,line cap=butt,miter limit=4.00,line
  width=1.600pt] (290.8707,317.5526) -- (290.8707,302.1819);
\path[scale=-1.000,draw=black,fill=cffffff,line join=miter,line cap=butt,miter
  limit=4.00,even odd rule,line width=0.800pt,rounded corners=0.0000cm]
  (-298.3453,-331.6139) rectangle (-283.3228,-316.9828);
\path[fill=black] (285.3148,326.8904) node[above right] (text5755-9-1-0)
  {${\scriptstyle P_N}$};
\path[draw=black,line join=miter,line cap=butt,miter limit=4.00,line
  width=0.800pt] (301.2829,324.2943) -- (329.1212,324.2943);
\path[draw=black,line join=miter,line cap=butt,miter limit=4.00,line
  width=1.600pt] (321.4742,317.5427) -- (321.4742,302.1720);
\path[xscale=1.000,yscale=-1.000,draw=black,fill=cffffff,line join=miter,line
  cap=butt,miter limit=4.00,even odd rule,line width=0.800pt,rounded
  corners=0.0000cm] (313.9996,-331.6040) rectangle (329.0221,-316.9729);
\path[fill=black] (315.8605,326.8904) node[above right] (text5755-9-1-5-0)
  {${\scriptstyle Q_N}$};
\end{tikzpicture} .
\end{align}
Now consider the  tensors
\begin{align}
A_{1,0}&=\left(\begin{matrix}
1& 1
\end{matrix}\right),\ \
A_{1,1}=\left(\begin{matrix}
1& 2
\end{matrix}\right),\\
\forall n\in\{2,\ldots,N\},\ \ 
A_{n,0}&=\left(\begin{matrix}
1& 0\\
0& 1
\end{matrix}\right),\ \ 
A_{n,1}=\left(\begin{matrix}
1& 2^n\\
0& 1
\end{matrix}\right),
\end{align}
and build an MPS by contracting tensors $A_1$ to $A_N$, as
\begin{align}\label{eq:openMPS}
\begin{tikzpicture}[y=0.80pt, x=0.80pt, yscale=-1.3000000, xscale=1.3000000, inner sep=0pt, outer sep=0pt,baseline=(current bounding box.center)]
\path[draw=black,line join=miter,line cap=butt,miter limit=4.00,line
  width=0.800pt] (420.0494,218.6181) -- (392.2111,218.6181);
\path[draw=black,line join=miter,line cap=butt,miter limit=4.00,line
  width=0.800pt] (399.6468,211.8664) -- (399.6468,196.4957);
\path[draw=black,line join=miter,line cap=butt,miter limit=4.00,line
  width=0.800pt] (397.7961,211.8664) -- (397.7961,196.4957);
\path[draw=black,line join=miter,line cap=butt,miter limit=4.00,line
  width=0.800pt] (395.9455,211.8664) -- (395.9455,196.4957);
\path[draw=black,line join=miter,line cap=butt,miter limit=4.00,line
  width=0.800pt] (401.4975,211.8664) -- (401.4975,196.4957);
\path[draw=black,line join=miter,line cap=butt,miter limit=4.00,line
  width=0.800pt] (403.3482,211.8664) -- (403.3482,196.4957);
\begin{scope}[cm={{-1.0,0.0,0.0,-1.0,(697.92114,501.33238)}},miter limit=4.00,line width=0.800pt]
  \path[draw=black,line join=miter,line cap=butt,miter limit=4.00,line
    width=0.800pt] (329.7672,282.7447) -- (365.6412,282.7447);
  \path[cm={{0.04091,0.0,0.0,-0.04091,(365.3516,291.59635)}},fill=black,miter
    limit=4.00,line width=19.553pt] (137.3807,216.2084) .. controls
    (137.3807,229.7373) and (126.4134,240.7046) .. (112.8845,240.7046) .. controls
    (99.3557,240.7046) and (88.3883,229.7373) .. (88.3883,216.2084) .. controls
    (88.3883,202.6795) and (99.3557,191.7122) .. (112.8845,191.7122) .. controls
    (126.4134,191.7122) and (137.3807,202.6795) .. (137.3807,216.2084) -- cycle;
  \path[cm={{0.04091,0.0,0.0,-0.04091,(371.01116,291.59635)}},fill=black,miter
    limit=4.00,line width=19.553pt] (137.3807,216.2084) .. controls
    (137.3807,229.7373) and (126.4134,240.7046) .. (112.8845,240.7046) .. controls
    (99.3557,240.7046) and (88.3883,229.7373) .. (88.3883,216.2084) .. controls
    (88.3883,202.6795) and (99.3557,191.7122) .. (112.8845,191.7122) .. controls
    (126.4134,191.7122) and (137.3807,202.6795) .. (137.3807,216.2084) -- cycle;
  \path[draw=black,line join=miter,line cap=butt,miter limit=4.00,line
    width=0.800pt] (379.8849,282.7522) -- (420.2232,282.7522);
\end{scope}
\path[draw=black,line join=miter,line cap=butt,miter limit=4.00,line
  width=0.800pt] (349.3754,212.0816) -- (349.3754,196.7109);
\path[draw=black,line join=miter,line cap=butt,miter limit=4.00,line
  width=0.800pt] (301.8923,212.0816) -- (301.8923,196.7109);
\path[draw=black,line join=miter,line cap=butt,miter limit=4.00,line
  width=0.800pt] (278.4092,212.0816) -- (278.4092,196.7109);
\path[scale=-1.000,draw=black,fill=cffffff,line join=miter,line cap=butt,miter
  limit=4.00,even odd rule,line width=0.800pt,rounded corners=0.0000cm]
  (-357.1390,-226.2949) rectangle (-342.1164,-211.6638);
\path[scale=-1.000,draw=black,fill=cffffff,line join=miter,line cap=butt,miter
  limit=4.00,even odd rule,line width=0.800pt,rounded corners=0.0000cm]
  (-309.4093,-226.2949) rectangle (-294.3867,-211.6638);
\path[scale=-1.000,draw=black,fill=cffffff,line join=miter,line cap=butt,miter
  limit=4.00,even odd rule,line width=0.800pt,rounded corners=0.0000cm]
  (-285.4182,-226.2949) rectangle (-270.3956,-211.6638);
\path[scale=-1.000,draw=black,fill=cffffff,line join=miter,line cap=butt,miter
  limit=4.00,even odd rule,line width=0.800pt,rounded corners=0.0000cm]
  (-407.3327,-225.9277) rectangle (-392.3101,-211.2966);
\path[fill=black] (375.9604,223.0300) node[above right] (text1906-6) {$=$};
\path[draw=black,line join=miter,line cap=butt,miter limit=4.00,line
  width=0.800pt] (469.8120,218.5185) -- (441.9737,218.5185);
\path[draw=black,line join=miter,line cap=butt,miter limit=4.00,line
  width=1.600pt] (449.6207,211.7669) -- (449.6207,196.3962);
\path[scale=-1.000,draw=black,fill=cffffff,line join=miter,line cap=butt,miter
  limit=4.00,even odd rule,line width=0.800pt,rounded corners=0.0000cm]
  (-457.0953,-225.8282) rectangle (-442.0728,-211.1971);
\path[fill=black] (427.0318,222.8515) node[above right] (text1906-6-3) {$=$};
\path[fill=black] (272.9734,221.0066) node[above right] (text5755) {${\scriptstyle A_1}$};
\path[fill=black] (344.1198,221.0664) node[above right] (text5755-9) {${\scriptstyle A_N}$};
\path[fill=black] (444.0648,220.2971) node[above right] (text5755-9-1) {${\scriptstyle P_N}$};
\end{tikzpicture}\\
\begin{tikzpicture}[y=0.80pt, x=0.80pt, yscale=-1.3000000, xscale=1.3000000, inner sep=0pt, outer sep=0pt,baseline=(current bounding box.center)]
\path[draw=black,line join=miter,line cap=butt,miter limit=4.00,line
  width=0.800pt] (317.4645,268.4561) -- (345.3027,268.4561);
\path[draw=black,line join=miter,line cap=butt,miter limit=4.00,line
  width=0.800pt] (337.8670,261.7045) -- (337.8670,246.3338);
\path[draw=black,line join=miter,line cap=butt,miter limit=4.00,line
  width=0.800pt] (339.7177,261.7045) -- (339.7177,246.3338);
\path[draw=black,line join=miter,line cap=butt,miter limit=4.00,line
  width=0.800pt] (341.5684,261.7045) -- (341.5684,246.3338);
\path[draw=black,line join=miter,line cap=butt,miter limit=4.00,line
  width=0.800pt] (336.0163,261.7045) -- (336.0163,246.3338);
\path[draw=black,line join=miter,line cap=butt,miter limit=4.00,line
  width=0.800pt] (334.1657,261.7045) -- (334.1657,246.3338);
\path[xscale=1.000,yscale=-1.000,draw=black,fill=cffffff,line join=miter,line
  cap=butt,miter limit=4.00,even odd rule,line width=0.800pt,rounded
  corners=0.0000cm] (330.1812,-275.7658) rectangle (345.2037,-261.1347);
\path[fill=black] (301.2138,272.5109) node[above right] (text1906-6-6) {$=$};
\path[draw=black,line join=miter,line cap=butt,miter limit=4.00,line
  width=0.800pt] (264.0128,268.7137) -- (291.8510,268.7137);
\path[draw=black,line join=miter,line cap=butt,miter limit=4.00,line
  width=1.600pt] (284.2041,261.9621) -- (284.2041,246.5914);
\path[xscale=1.000,yscale=-1.000,draw=black,fill=cffffff,line join=miter,line
  cap=butt,miter limit=4.00,even odd rule,line width=0.800pt,rounded
  corners=0.0000cm] (276.7295,-276.0234) rectangle (291.7520,-261.3923);
\path[fill=black] (352.2852,272.6895) node[above right] (text1906-6-3-8) {$=$};
\begin{scope}[cm={{1.0,0.0,0.0,-1.0,(39.3587,551.26813)}},miter limit=4.00,line width=0.800pt]
  \path[draw=black,line join=miter,line cap=butt,miter limit=4.00,line
    width=0.800pt] (329.7672,282.7447) -- (365.6412,282.7447);
  \path[cm={{0.04091,0.0,0.0,-0.04091,(365.3516,291.59635)}},fill=black,miter
    limit=4.00,line width=19.553pt] (137.3807,216.2084) .. controls
    (137.3807,229.7373) and (126.4134,240.7046) .. (112.8845,240.7046) .. controls
    (99.3557,240.7046) and (88.3883,229.7373) .. (88.3883,216.2084) .. controls
    (88.3883,202.6795) and (99.3557,191.7122) .. (112.8845,191.7122) .. controls
    (126.4134,191.7122) and (137.3807,202.6795) .. (137.3807,216.2084) -- cycle;
  \path[cm={{0.04091,0.0,0.0,-0.04091,(371.01116,291.59635)}},fill=black,miter
    limit=4.00,line width=19.553pt] (137.3807,216.2084) .. controls
    (137.3807,229.7373) and (126.4134,240.7046) .. (112.8845,240.7046) .. controls
    (99.3557,240.7046) and (88.3883,229.7373) .. (88.3883,216.2084) .. controls
    (88.3883,202.6795) and (99.3557,191.7122) .. (112.8845,191.7122) .. controls
    (126.4134,191.7122) and (137.3807,202.6795) .. (137.3807,216.2084) -- cycle;
  \path[draw=black,line join=miter,line cap=butt,miter limit=4.00,line
    width=0.800pt] (379.8849,282.7522) -- (420.2232,282.7522);
\end{scope}
\path[draw=black,line join=miter,line cap=butt,miter limit=4.00,line
  width=0.800pt] (387.9045,262.0174) -- (387.9045,246.6467);
\path[draw=black,line join=miter,line cap=butt,miter limit=4.00,line
  width=0.800pt] (435.3876,262.0174) -- (435.3876,246.6467);
\path[draw=black,line join=miter,line cap=butt,miter limit=4.00,line
  width=0.800pt] (458.8706,262.0174) -- (458.8706,246.6467);
\path[xscale=1.000,yscale=-1.000,draw=black,fill=cffffff,line join=miter,line
  cap=butt,miter limit=4.00,even odd rule,line width=0.800pt,rounded
  corners=0.0000cm] (378.1408,-276.2307) rectangle (399.1634,-261.5996);
\path[xscale=1.000,yscale=-1.000,draw=black,fill=cffffff,line join=miter,line
  cap=butt,miter limit=4.00,even odd rule,line width=0.800pt,rounded
  corners=0.0000cm] (427.8705,-276.2307) rectangle (442.8931,-261.5996);
\path[xscale=1.000,yscale=-1.000,draw=black,fill=cffffff,line join=miter,line
  cap=butt,miter limit=4.00,even odd rule,line width=0.800pt,rounded
  corners=0.0000cm] (451.8617,-276.2307) rectangle (466.8842,-261.5996);
\path[fill=black] (379.0107,271.3098) node[above right] (text5755-3)
  {${\scriptscriptstyle A_{N{+}1}}$};
\path[fill=black] (451.5438,271.3098) node[above right] (text5755-3-6)
  {${\scriptstyle A_{2N}}$};
\path[fill=black] (278.5904,271.3098) node[above right] (text5755-9-1-5)
  {${\scriptstyle Q_{N}}$};
\end{tikzpicture}
\end{align}
We obtain a tensor $T_N$ with $N$ open indices corresponding to $N$ binary variables and an extra virtual index of dimension 2. If we reshape this tensor as a $2^N \times 2$ matrix we obtain matrix $P_N$. In the same way we can obtain an MPS of non-negative tensors such that contracting $N$ sites gives a tensor that can be reshaped as $Q_N$. Contracting the two extra virtual indices between the two MPS, we finally obtain an MPS$_{\mathbb{R}_{\geq 0}}$ over $2N$ variables such that $M$ is the central bipartition of the resulting tensor. Suppose that there is a BM$_\mathbb{R}$ defining the same probability mass function over $2N$ variables, then it has Born-rank$_\mathbb{R}$ larger or equal to the square root rank of $M$, which is larger than the square root rank of $K$, which is $\pi(2^{N+1})$. This proves the case $[3,2]$, the remaining cases follow directly from Proposition~\ref{sup_prop:inclusions}.
\end{proof}

\begin{proposition}[${[2,3]}$, ${[2,4]}$, ${[2,5]}$, ${[2,6]}$ ] \label{sup_prop:asymptotics_3}
There exists a family of non-negative tensors over $2N$ binary variables with constant Born-rank$_{\mathbb{R}}=2$ (and hence also constant Born-rank$_{\mathbb{C}}$ and puri-rank$_{\mathbb{R}/\mathbb{C}}$) that have TT-rank$_{\mathbb{R}_{\geq 0}} \geq N$.
\end{proposition}
\begin{proof}
Consider the linear Euclidean matrices (see Lemma \ref{sup_lemma:euclidian_matrices}) defined as $M_{i,j} = (j-i)^2$ and observe that $M$ is the element-wise square of a matrix $H_{i,j}=j-i$. We have $H=P_N Q_N$, where 
\begin{align}
P_N=\left(\begin{matrix}
1& 0\\
1& -1\\
1& -2\\
\vdots& \vdots \\
1& -2^N+1
\end{matrix}\right),
Q_N=\left(\begin{matrix}
0& 1& 2& \cdots& 2^N-1\\
1& 1& 1& \cdots& 1
\end{matrix}\right).
\end{align}
Now consider the tensors
\begin{align}
A_{1,0}&=\left(\begin{matrix}
1& 0
\end{matrix}\right),\ \
A_{1,1}=\left(\begin{matrix}
1& -1
\end{matrix}\right),\\
\forall n>1,\ \ 
A_{n,0}&=\left(\begin{matrix}
1& 0\\
0& 1
\end{matrix}\right),\ \ 
A_{n,1}=\left(\begin{matrix}
1& -2^n\\
0& 1
\end{matrix}\right),
\end{align}
and build an MPS by contracting tensors $A_1$ to $A_n$. We obtain a tensor $T_N$ with $N$ open indices corresponding to $N$ binary variables and an extra virtual index of dimension 2, as in Equation~\eqref{eq:openMPS}. If we reshape this tensor as a $2^N \times 2$ matrix we obtain matrix $P_N$. In the same way we can obtain an MPS of non-negative tensors such that contracting $N$ sites gives a tensor that can be reshaped as $Q_N$. Contracting the two extra virtual indices between the two MPS, we finally obtain an MPS$_\mathbb{R}$ over $2N$ binary variables such that $H$ is the central bipartition of the resulting tensor. By squaring this MPS we obtain a BM$_\mathbb{R}$ over $2N$ binary variables such that $M$ is the central bipartition of the resulting tensor. Suppose that there is an MPS$_{\mathbb{R}_{\geq 0}}$ defining the same probability mass function over $2N$ variables, then it has TT-rank$_{\mathbb{R}_{\geq 0}}$ larger or equal to the non-negative rank of $M$, which is larger than $\log_2 2^N=N$. This proves the case $[2,3]$, and again the remaining cases follow from Proposition \ref{sup_prop:inclusions}.
\end{proof}

\begin{proposition}[${[3,4]}$]
\label{sup_prop:asymptotics_4}
There exists a family of non-negative tensors over 2$N$ binary variables with constant Born-rank$_{\mathbb{C}}=2$, but with Born-rank$_{\mathbb{R}}\geq N$.
\end{proposition}
\begin{proof}
Consider the $N\times N$ matrices $M_N$ from Lemma~\ref{sup_lemma:matrix_families_born_ranks}. These matrices have complex square root rank 2 but real square root rank $N$. Using Lemma~\ref{sup_lemma:embedding_and_unfolding}, this means that there is a BM$_\mathbb{C}$ over $2N$ 
variables of Born-rank$_\mathbb{C}$ equal to 2 such that $M_N$ is a submatrix of the central bipartition of the resulting tensor. The Born-rank$_{\mathbb{R}}$ of this tensor is at least the Born-rank$_{\mathbb{R}}$ of $M_N$, which is $N$.
\end{proof}

\section{Learning algorithms and numerical experiments}

\subsection{Learning algorithms for MPS$_{\mathbb{R}_{\geq 0}}$}

As in the case of LPS, MPS$_{\mathbb{R}_{\geq 0}}$ can be trained using gradient descent to minimize the log-likelihood. Consider an MPS$_{\mathbb{R}_{\geq 0}}$, and assume that the tensors $A_i$ in the MPS are given by the element-wise square of real tensors $B_i$. The normalization can be computed by contracting the following tensor network from left to right, where the circles 
represent a vector of ones of dimension $d$, as
\begin{align}
Z_T=\begin{tikzpicture}[y=0.9pt, x=0.9pt,yscale=-1.0, xscale=1.0,  baseline=(current bounding box.center)]
\begin{scope}[shift={(5.9747,-163.24549)},miter limit=4.00,line width=0.800pt]
  \path[draw=black,line join=miter,line cap=butt,miter limit=4.00,line
    width=0.800pt] (325.3029,282.7447) -- (365.6412,282.7447);
  \path[cm={{0.04091,0.0,0.0,-0.04091,(365.3516,291.59635)}},fill=black,miter
    limit=4.00,line width=19.553pt] (137.3807,216.2084) .. controls
    (137.3807,229.7373) and (126.4134,240.7046) .. (112.8845,240.7046) .. controls
    (99.3557,240.7046) and (88.3883,229.7373) .. (88.3883,216.2084) .. controls
    (88.3883,202.6795) and (99.3557,191.7122) .. (112.8845,191.7122) .. controls
    (126.4134,191.7122) and (137.3807,202.6795) .. (137.3807,216.2084) -- cycle;
  \path[cm={{0.04091,0.0,0.0,-0.04091,(371.01116,291.59635)}},fill=black,miter
    limit=4.00,line width=19.553pt] (137.3807,216.2084) .. controls
    (137.3807,229.7373) and (126.4134,240.7046) .. (112.8845,240.7046) .. controls
    (99.3557,240.7046) and (88.3883,229.7373) .. (88.3883,216.2084) .. controls
    (88.3883,202.6795) and (99.3557,191.7122) .. (112.8845,191.7122) .. controls
    (126.4134,191.7122) and (137.3807,202.6795) .. (137.3807,216.2084) -- cycle;
  \path[draw=black,line join=miter,line cap=butt,miter limit=4.00,line
    width=0.800pt] (379.8849,282.7522) -- (420.2232,282.7522);
\end{scope}
\path[draw=black,line join=miter,line cap=butt,miter limit=4.00,line
  width=0.800pt] (331.0374,126.0053) -- (331.0374,141.3759);
\path[draw=black,line join=miter,line cap=butt,miter limit=4.00,line
  width=0.800pt] (354.5205,126.0053) -- (354.5205,141.3759);
\path[draw=black,line join=miter,line cap=butt,miter limit=4.00,line
  width=0.800pt] (402.0036,126.0053) -- (402.0036,141.3759);
\path[draw=black,line join=miter,line cap=butt,miter limit=4.00,line
  width=0.800pt] (425.4866,126.0052) -- (425.4866,141.3759);
\path[draw=black,fill=cffffff,line join=miter,line cap=butt,miter
  limit=4.00,even odd rule,line width=0.800pt,rounded corners=0.0000cm]
  (323.8884,111.7919) rectangle (338.9110,126.4231);
\path[draw=black,fill=cffffff,line join=miter,line cap=butt,miter
  limit=4.00,even odd rule,line width=0.800pt,rounded corners=0.0000cm]
  (346.7569,111.7919) rectangle (361.7794,126.4231);
\path[draw=black,fill=cffffff,line join=miter,line cap=butt,miter
  limit=4.00,even odd rule,line width=0.800pt,rounded corners=0.0000cm]
  (394.4866,111.7919) rectangle (409.5091,126.4231);
\path[draw=black,fill=cffffff,line join=miter,line cap=butt,miter
  limit=4.00,even odd rule,line width=0.800pt,rounded corners=0.0000cm]
  (418.4777,111.7919) rectangle (433.5002,126.4231);
\path[shift={(-0.43111,-1.11691)},color=black,draw=black,opacity=0.850,line
  join=miter,line cap=butt,miter limit=4.00,line width=0.800pt]
  (337.0122,147.8337)arc(0.000:180.000:5.745)arc(-180.000:0.000:5.745) -- cycle;
\path[shift={(23.11806,-1.11691)},color=black,draw=black,opacity=0.850,line
  join=miter,line cap=butt,miter limit=4.00,line width=0.800pt]
  (337.0122,147.8337)arc(0.000:180.000:5.745)arc(-180.000:0.000:5.745) -- cycle;
\path[shift={(70.84777,-1.11691)},color=black,draw=black,opacity=0.850,line
  join=miter,line cap=butt,miter limit=4.00,line width=0.800pt]
  (337.0122,147.8337)arc(0.000:180.000:5.745)arc(-180.000:0.000:5.745) -- cycle;
\path[shift={(94.33381,-1.11691)},color=black,draw=black,opacity=0.850,line
  join=miter,line cap=butt,miter limit=4.00,line width=0.800pt]
  (337.0122,147.8337)arc(0.000:180.000:5.745)arc(-180.000:0.000:5.745) -- cycle;
\end{tikzpicture} .
\end{align}
This contraction is performed in $\mathcal{O}(d r^2 N)$ operations. During this contraction, intermediate results from the contraction of the first $i$ tensors are stored in $E_i$, and the same procedure is repeated from the right with intermediate results of the contraction of the last $N-i$ tensors stored in $F_{i+1}$. The derivatives of the normalization for each tensor are then computed as
\begin{align}
\frac{\partial Z_T}{\partial A_{i,m}^{j,k}}=\begin{tikzpicture}[y=0.9pt, x=0.9pt,yscale=-1.0, xscale=1.0,  baseline=(current bounding box.center)]
\path[draw=black,line join=miter,line cap=butt,line width=0.800pt]
  (334.6130,196.7630) -- (357.0889,196.7630);
\path[draw=black,fill=cffffff,line join=miter,line cap=butt,miter
  limit=4.00,even odd rule,line width=0.800pt,rounded corners=0.0000cm]
  (316.0902,189.5558) rectangle (350.4876,204.1869);
\path[fill=black] (320.9359,205.5713) node[above right] (text4737-8)
  {$E_{i-1}$};
\path[draw=black,line join=miter,line cap=butt,line width=0.800pt]
  (375.1211,196.7783) -- (400.6275,196.7783);
\path[draw=black,fill=cffffff,line join=miter,line cap=butt,miter
  limit=4.00,even odd rule,line width=0.800pt,rounded corners=0.0000cm]
  (382.2552,189.5558) rectangle (416.6526,204.1869);
\path[fill=black] (388.3728,205.7726) node[above right] (text4737-0)
  {$F_{i+1}$};
\path[fill=black] (354.8979,203.9769) node[above right] (text5815) {$j$};
\path[fill=black] (363.1071,202.9769) node[above right] (text5819) {$k$};
\path[draw=black,line join=miter,line cap=butt,miter limit=4.00,line
  width=0.800pt] (365.4711,215.1811) -- (365.4711,225.0161);
\path[shift={(34.00259,82.5232)},color=black,draw=black,opacity=0.850,line
  join=miter,line cap=butt,miter limit=4.00,line width=0.800pt]
  (337.0122,147.8337)arc(0.000:180.000:5.745)arc(-180.000:0.000:5.745) -- cycle;
\path[fill=black] (358.5357,213.6122) node[above right] (text5950) {$m$};
\end{tikzpicture},
\end{align}
and the derivatives with respect to the original parameters are obtained as
\begin{align}
    \frac{\partial Z_T}{\partial B_{i,m}^{j,k}}=
    2\frac{\partial Z_T}{\partial A_{i,m}^{j,k}} B_{i,m}^{j,k}.
\end{align}
Applying the same procedure by replacing the circle tensors with indices 
corresponding to the observed variables at a training example leads to the computation of $T_{\mathbf{x_i}}$ and its derivative.

\subsection{Tensor factorizations}

We minimize $D(P||T/Z_T)$ through a non-linear limited-memory BFGS optimization algorithm. The gradient of the KL-divergence depends on the log-derivatives of $T$ and of $Z_T$, which have already been obtained while computing the gradient of the log-likelihood. 

Optimizing an MPS$_\mathbb{R}$ requires to impose the non-negativity of the tensor network. In order to still provide a comparison with MPS$_\mathbb{R}$, we optimize them by adding a penalty term constraining all elements of the contracted tensor to be non-negative. This constraint is difficult to satisfy, so the optimization may not converge to the global minimum.

Note that we could have chosen any distance instead of the KL-divergence. In particular if instead we minimize the 2-norm for vectors $||P-T||_2$ the optimization for an MPS$_\mathbb{R}$ can be done by keeping only the largest singular values of the tensor $P$ (in the case of matrices), and a good starting point for tensors can be obtained through successive truncated singular value decompositions. We find that the results we obtained are not significantly modified if the 2-norm was considered instead of the KL-divergence.

\subsection{Generalization performance}

Our results focus on the expressive power of different tensor-network representations, and for this reason the numerical experiments focused on the accuracy obtained while training on the full datasets. It is nevertheless also interesting to see how much these results would be modified if we considered generalization performance instead. 

\begin{figure}[h]
\centering
\includegraphics[width =0.97\linewidth]{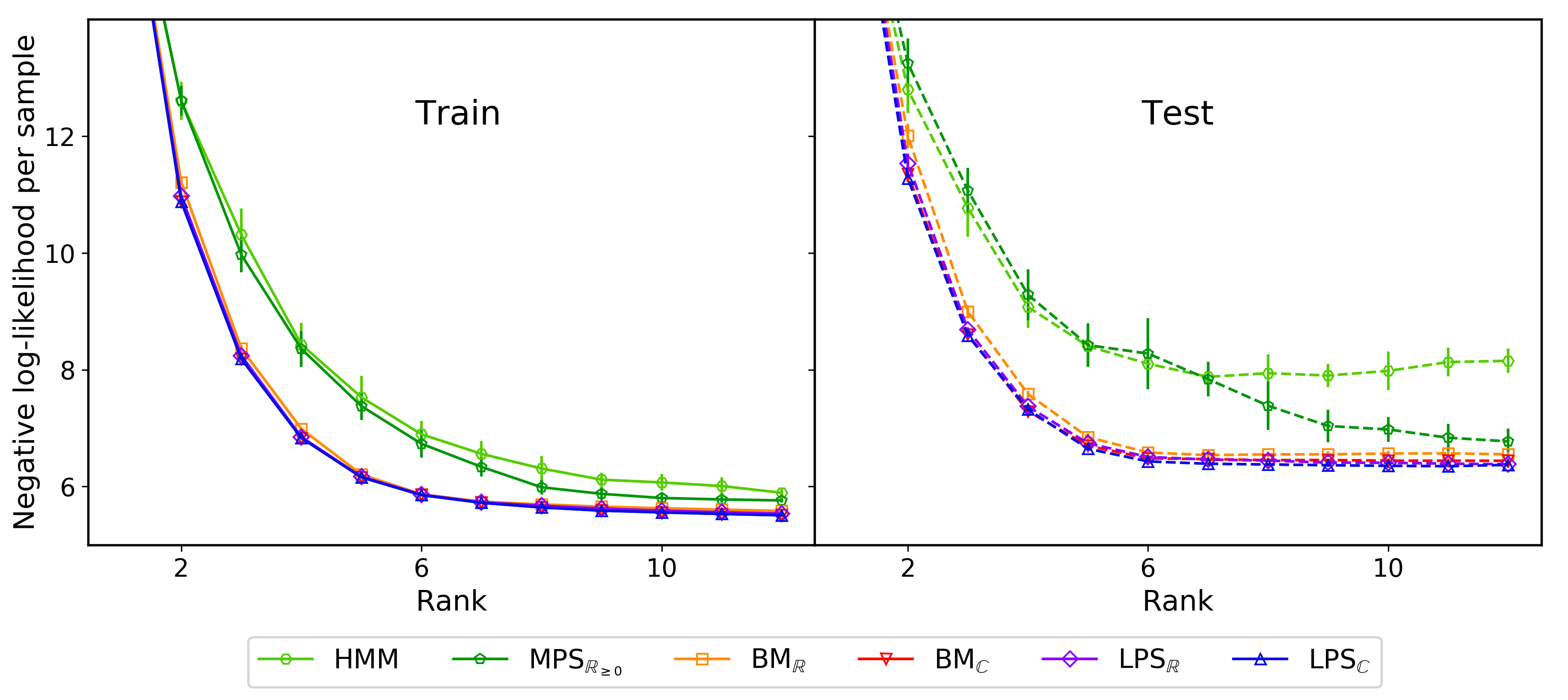}
\caption{\small Negative log-likelihood per sample obtained on the training set (left) and on the test set (right) for different tensor-network representations on the biofam data set. The error bars represent one standard deviation with different initial conditions.}
\label{fig:biofamgeneralization}
\end{figure}

In order to investigate the differences obtained on training sets or test sets, we focus on the biofam data set of family life states from the Swiss Household Panel biographical survey \cite{biofam}, since small bond dimensions are sufficient to get already converged results on training sets and since this data set includes variables that have a natural sequential order : they represent the family life states from age 15 to 30 (sequence length is 16) and the variables can take 8 possible values. The probability distribution is therefore a tensor of size $8^{16}=2^{48}$, which requires to use models with smaller number of parameters. We use 1000 examples in the training set, 500 in the validation set and 500 in the test set. All models are trained for $20000$ epochs with a batch size of 20 on the training set, and the best training accuracy with respect to different learning rates (using a grid search on powers of $10$ going from $10^{-5}$ to $10^5$) is recorded. The procedure is repeated 20 times with a different random initialization of the tensors, and the mean and standard deviation are reported in Fig.~\ref{fig:biofamgeneralization} (left). For each initial condition, the best model on the validation set is chosen to be evaluated on the test set. The mean and standard deviation of the accuracy on the test set with respect to different initial conditions are reported Fig.~\ref{fig:biofamgeneralization} (right).

The results indicate that despite rather small differences between HMM and non-negative MPS on the training set, the differences on the test set are important and non-negative MPS are able to reach a better accuracy at larger ranks, when HMM do not improve anymore.
Born machines and LPS all reach better performance than HMM and non-negative MPS already for small ranks, and with smaller variance with respect to the initial conditions.

\bibliographystyle{unsrt}
\bibliography{biblio}